%% file: root.tex
\documentclass[smallcondensed]{svjour3}     

\smartqed  

\input{packages.tex}
\input{math_shortcut.tex}

\usepackage{hyperref}

\contourlength{2.3pt}

\makeatletter
\let\orgdescriptionlabel\descriptionlabel
\renewcommand*{\descriptionlabel}[1]{%
  \let\orglabel\label
  \let\label\@gobble
  \phantomsection
  \edef\@currentlabel{#1}%
  \let\label\orglabel
  \orgdescriptionlabel{#1}%
}
\makeatother

\begin{document}

\title{Dynamic Movement Primitives: Volumetric Obstacle Avoidance Using Dynamic Potential Functions
\thanks{
    *This research has received funding from the European Research Council (ERC) under the European Union’s Horizon 2020 research and innovation programme, ARS (Autonomous Robotic Surgery) project, grant agreement No. 742671.
}
}

\titlerunning{DMPs: Volumetric Obstacle Avoidance Using Dynamic Potential Functions}        

\author{
    Michele Ginesi$^1$ \and
    Daniele Meli$^1$ \and
    Andrea Roberti$^1$ \and
    Nicola Sansonetto$^1$ \and
    Paolo Fiorini$^1$
}

\authorrunning{M. Ginesi \and D. Meli \and N.Sansonetto \and P. Fiorini} 

\institute{M. Ginesi \at
    \email{michele.ginesi@univr.it}
    \\
    \and{}
    $\,^1$ Department of Computer Science, University of Verona, Strada le Grazie 15, 37134 Verona, Italy
}

\date{Received: date / Accepted: date}

\maketitle

\begin{abstract}
    Obstacle avoidance for DMPs is still a challenging problem.
    In our previous work, we proposed a framework for obstacle avoidance based on superquadric potential functions to represent volumes.
    In this work, we extend our previous work to include the velocity of the trajectory in the definition of the potential.
    Our formulations guarantee smoother behavior with respect to state-of-the-art point-like methods.
    Moreover, our new formulation allows to obtain a smoother behavior in proximity of the obstacle than when using a static (i.e. velocity independent) potential.
    We validate our framework for obstacle avoidance in a simulated multi-robot scenario and with different real robots: a pick-and-place task for an industrial manipulator and a surgical robot to show scalability; and navigation with a mobile robot in dynamic environment.
\keywords{Obstacle Avoidance \and Dynamic Movement Primitives \and Learning from Demonstration}
\end{abstract}



\section{Introduction}

Robots are now used in complex scenarios, ranging from industrial and manufacturing processes to aerospace and health care.
As their involvement in common human tasks increases, adaptability and reliability at the motion planning level is often required, and imitation of human behavior often helps in this direction.\\
Standard motion planning techniques, such as splines, potentials and others \cite{LCL83,RK92,MKY06,RZBS09}, work well when an objective function has to be optimized (e.g. minimize the time of execution of the trajectory, or the energy consumption).
A \emph{Learning from Demonstration} (LfD) approach, is usually preferable if one needs to learn human gestures.
In LfD, a human operator shows an example trajectory or task execution, and parameters are learned for replication in different situations and environment.
In last fifteen years, various LfD approaches (as \emph{Gaussian Mixture Models} \cite{KZB11}, \emph{Extreme Learning Machines} \cite{huang2006extreme,DOHWJX17}, and others \cite{ARRWLUB11}) have been developed in order to replicate human gestures.
These LfD techniques may require a huge amount of demonstrations to be properly trained, which can represent a bottleneck when many different motion primitives have to be learned (e.g., for productive and cost reasons in industry).\\
In this paper we focus on the obstacle avoidance problem within the Dynamic Movement Primitives (DMPs) framework\cite{INS02,Sch06,PHPS08,HPPS09}.
DMPs permit to learn a trajectory from just one demonstration.
They encode the trajectory in a system of second-order linear Ordinary Differential Equation (ODE), where a forcing term is learned as a linear combination of predefined time-dependent functions.
They are successfully used in many robotic scenarios, such as cloth manufacturing \cite{joshi2017robotic}, reproduction of human walk for exoskeletons \cite{HCGC16}, and collaborative bimanual tasks \cite{GNIU14}.
Obstacle avoidance for DMPs has been successfully treated for point-like obstacles (e.g. \cite{PHPS08} and \cite{HPPS09}.
On the other hand, volumetric obstacle avoidance has been treated in our previous work \cite{GMCDSF19} using potential functions.
Other approaches (e.g. \cite{MHM10,RMIS14,RSSM17,SSSM18}) require multiple demonstrations with different types and sizes of the obstacles.\\
In this work we improve our previous framework \cite{GMCDSF19}.
In particular, we introduce a new potential function.
This new potential is velocity-dependent, and this allows to achieve smoother obstacle avoidance behaviors compared to static (i.e. dependent only on position) potentials.
Moreover, we will show that our approach results in trajectories that deviate less from the desired behavior than other frameworks.
We validate our approach in a simulated multi-robot coordination scenario, where three mobile robots have to reach pre-defined targets while avoiding each other and obstacles in the scene.
We also show the generality of our frameworks as applied to different real robotic scenarios.
In detail, we test a pick-and-place task with an encumbrant industrial manipulator, combining DMP-level obstacle avoidance with collision-free inverse kinematic computation.
We then show that the scalability of DMPs is preserved with our framework, replicating the pick-and-place task on a smaller setup with a bi-manual surgical robot.
Finally, we show the reactivity of our approach with a mobile robot in a dynamic scene with moving obstacles to be detected by a RGB-D camera.
In Section~\ref{sec:dmp_general} we recall the theory of DMPs, focusing, in Section~\ref{subsec:sota_obstacle}, on the existing methods to treat obstacle avoidance.
Then, in Section~\ref{sec:contribution} we present our new dynamic potential function.
In Section~\ref{sec:results} we show our results: in Section~\ref{subsec:result_synth} we compare our new method to the state of the art, showing that our novel method results in a trajectory that is both smoother and it remains to the learned one;
in Section \ref{subsec:sim_robot} we compare our previous static potential for volumes with the new dynamic one, in a scenario with multiple mobile robots and prior scene awareness;
in Section \ref{subsec:result_real} we compare our frameworks (static and dynamic) with the aforementioned robots.

Our code, freely available at \texttt{https://github.com/mginesi/dmp\_vol\_obst} includes a Python 3.5 implementation of DMPs and our proposed approach to volumetric obstacle avoidance.


\section{Dynamic Movement Primitives}
\label{sec:dmp_general}

\emph{Dynamic Movement Primitives} is a framework for trajectory learning.
It is based upon an Ordinary Differential Equation (ODE) of spring-mass-damper type with a forcing term.
This framework has numerous advantages that make it well suited for robotic applications.
First, any trajectory can be learned and subsequently executed while changing starting and goal positions.
Second, the executed trajectory will always converge to the goal, maintaining a similar shape to the learned trajectory.
Third, the learned trajectory can be executed at different speed simply by changing a single parameter.
Finally, DMPs have been proven to be flexible enough to being extended in multiple ways: for instance, the formulation can be modified to deal periodic movements \cite{INS03,UGAM10}, to learn sensory experience \cite{PRKS11,PKRS12}, and to work in unit quaternion space (in order to model orientations) \cite{UNPM14,SFL19}.
Another extension, that is the topic treated in this paper, is the inclusion of obstacle avoidance in the DMP framework \cite{PHPS08,HPPS09,GMCDSF19}.

In this Section, we recall the DMP formulation given in \cite{PHPS08,HPPS09,PHAS09} upon which our work is based.
Such formulation is an improvement of the original formulation by \cite{INS02,INS03,Sch06,INHPS13}.
Subsequently, in Section~\ref{subsec:sota_obstacle} we will present the state of the art of obstacle avoidance methods for DMPs, highlighting their strengths and weaknesses.

Dynamic Movement Primitives consist of the following system of Ordinary Differential Equations:
\begin{subnumcases}{\label{eqs:dmps_eq}}
    \tau \dot{\vv} = \mK (\vg - \vx) - \mD \vv - \mK (\vg - \vx_0) s + \mK \vf (s) \label{eq:dmp_acc}\\
    \tau \dot{\vx} = \vv \label{eq:dmp_vel}
\end{subnumcases}
Vectors $ \vx, \vv \in \RR ^ d $ are, respectively, the \emph{position} and \emph{velocity} of the system;
and $ \vx_0, \vg \in \RR^d $ are, respectively, the \emph{starting} and \emph{goal positions}.
Matrices $ \mK, \mD \in \RR_+^{d \times d} $ are, respectively, the \emph{elastic} and \emph{damping terms} of the system.
Both are diagonal matrices, \( \mK = \diag(K_1, K_2, \ldots, K_d) \), \( \mD = \diag(D_1, D_2, \ldots, D_d) \), and satisfy the critical dumping relation $ D_i = 2 \sqrt{K_i} $, so that the un-perturbed system, i.e. when $\vf \equiv \vzero$, converges as fast as possible to the unique equilibrium $ (\vx, \vv) = (\vg, \vzero) $.
Scalar $ \tau \in \RR_+ $ is a \emph{temporal scaling factor} which can be used to make the execution of the trajectory faster or slower.
Function $ \vf : \RR \to \RR^d $ is the \emph{forcing} (also called \emph{perturbation}) \emph{term}.
Scalar $s \in (0, 1]$ is a re-parametrization of time $ t \in [0, T] $ governed by the so called \emph{canonical system}
\begin{equation}
    \label{eq:can_sys}
    \tau \dot{s} = -\alpha s,
\end{equation}
where $ \alpha \in \RR_+ $ and the initial state is $ s(0) = 1 $.\\
The forcing term $\vf(s) = [f_{1}(s), f_{2}(s), \ldots, f_{d}(s)] \transpose$ is written in term of basis functions.
Each component $ f_{p} (s) $, $ p=1,2,\ldots,d $ has then the form
\begin{equation}
    \label{eq:forcing_term}
    f_{p}(s) = \frac{ \sum_{i=0}^N {}^p\omega_i \, \psi_i(s) }{ \sum_{i=0}^N \psi_i(s) } \, s,
\end{equation}
where $ {}^p\omega_i \in \RR $ is called \emph{weigth}, and $\psi_i(s)$ is a \emph{Gaussian Radial Basis} (GRB) \emph{function} defined as
\begin{equation}
    \label{eq:gaussian_bf}
    \psi_i(s) = \exp \br{ - h_i (s - c_i) ^ 2 },
\end{equation}
with \emph{centers} $c_i$ defined as
\begin{equation}
    \label{eq:gbf_center}
    c_i = \exp \br{ -\alpha \, i\, \frac{T}{N} }, \quad i = 0,1,\ldots,N,
\end{equation}
and \emph{widths} defined as
\begin{equation}
    \label{eq:bgf_width}
    \begin{aligned}
        h_i & = \frac{ 1 }{ \br{ c_{i+1} - c_i } ^ 2 }, \quad i = 0, 1, \ldots, N - 1, \\
        h_{N} & = h_{N-1}.
    \end{aligned}
\end{equation}

\begin{figure}[htb]
    \centering
    \includegraphics[width=0.8\linewidth]{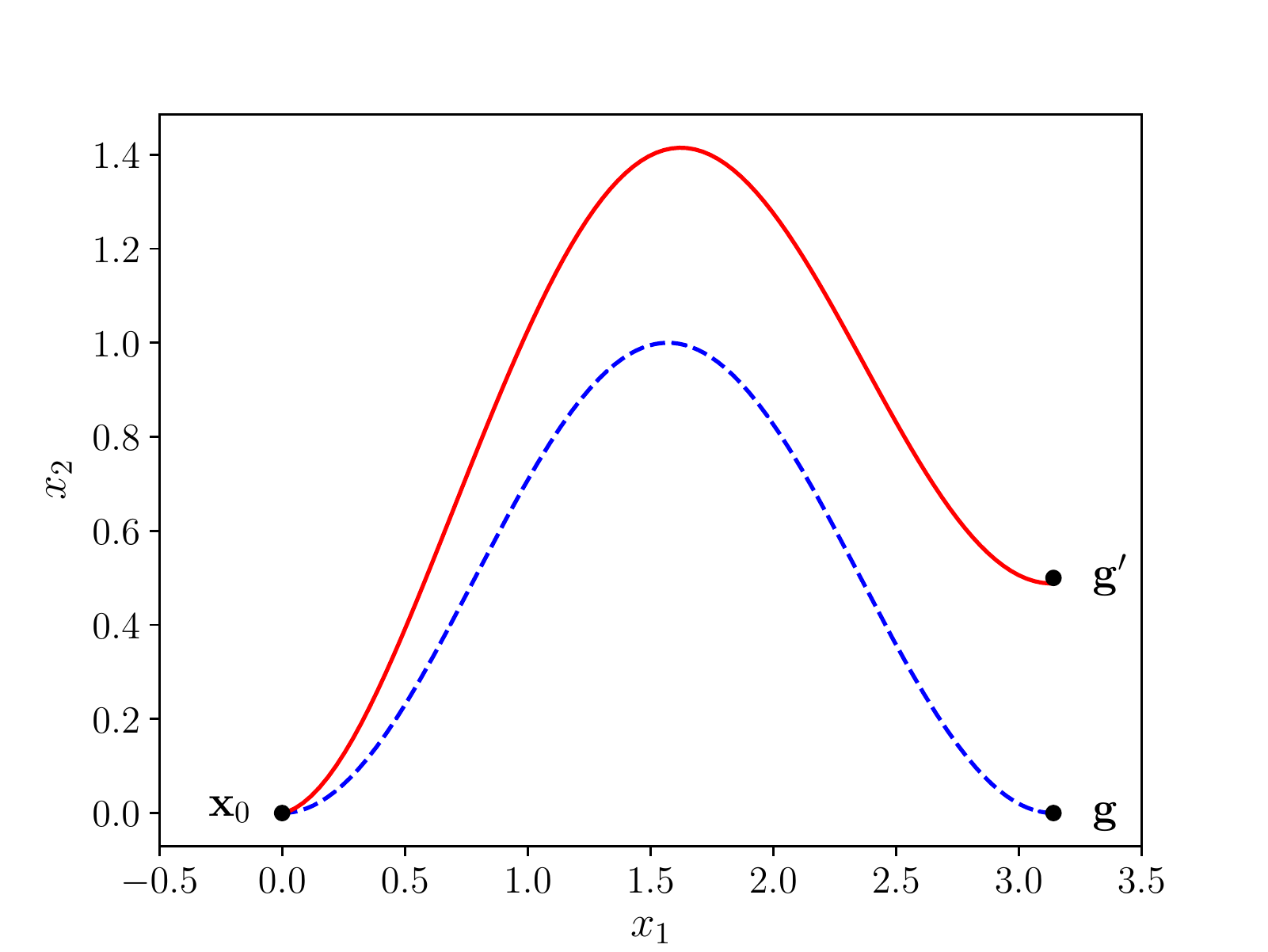}
    \caption{Example of execution of a DMP in $ \RR^2 $.
    The blue dashed line shows the desired trajectory, which start at $ \vx_0 = [0, 0]\transpose $ and ends at $ \vg = [\pi, 0]\transpose $.
    The solid red line shows the execution of the learned DMP when changing goal position to $ \vg ' = [\pi, 0.5]\transpose $.}
    \label{fig:dmp_exp}
\end{figure}

During the \emph{learning phase}, a desired trajectory $ \widetilde{\vx} (t) $ and its velocity $ \widetilde{\vv} (t) $ are recorded.
Then, from \eqref{eq:dmp_acc}, the desired forcing term $ \widetilde{\vf} (s(t)) $ is computed (after fixing matrices $ \mK $ and $\mD$).
Finally, the weights $ {}^p\omega_i $, $ i=0,1,\ldots,N $, $ p=1,2,\ldots, d $ that best approximate the desired forcing term $ \widetilde{\vf} $ using formulation \eqref{eq:forcing_term} are computed.\\
During the \emph{execution phase}, starting and goal positions $ \vx_0, \vg $ are set, and the forcing term $\vf$ is computed using \eqref{eq:forcing_term} with the weights computed before.
Solving the dynamical system \eqref{eqs:dmps_eq} will give a trajectory of similar shape to the learned one, that start from $\vx_0$ and converges to $\vg$.
In Figure~\ref{fig:dmp_exp} an example of the spatial generalization property of the DMP framework is shown.


\subsection{Methods for Obstacle Avoidance}
\label{subsec:sota_obstacle}

In the literature, there exist two main ways to implement obstacle avoidance in the DMP framework.
The first approach is the so-called \emph{Stylistic DMPs} \cite{MHM10}
in which a probability distribution $ q ( {}^p\omega_i | \zeta ) $ of the weights, conditioned to a \emph{style parameter} $\zeta$ is learned, instead of the set of weights $ \{{}^p\omega_i $\}.
The style parameter can be, for instance, the size of an obstacle.
The second approach, instead, consists in adding a \emph{repulsive term} $ \vvphi (\vx, \vv) \in \RR ^ d $, that 'pushes' the trajectory away from the obstacle, to \eqref{eq:dmp_acc}, that then reads
\begin{equation}
    \label{eq:dmp_acc_obst}
    \tau \dot{\vv} = \mK (\vg - \vx) - \mD \vv - \mK (\vg - \vx_0) s + \mK \vf (s) + \vvphi (\vx, \vv).
\end{equation}
In full generality, the repulsive term $\vvphi$ depends on both position $\vx$ and velocity $\vv$ of the system, but we will see that for some methods, it depends only on the position.
This second approach can be further subdivided into two sub-categories.
The first includes all those approaches that require an additional learning phase, in which executions both with and without obstacle are recorded, to model $\vvphi$.
For instance, in \cite{RSSM17} and \cite{SSSM18} a Neural Network is used to model the perturbation term.
In \cite{RMIS14} an analytical formulation is presented, but, the number of free parameters that has to be tuned requires an additional learning process.
The second sub-category, in which our approach fits, comprehends all approaches in which there is no need for any additional learning phase.
This is a great advantage, since the DMP can be used in virtually any situation, while the learning approaches may fail in situations too dissimilar to the ones shown during the learning phase.

The proposed method enters the `designed by hand' approaches, therefore we will recall here, and compare our approach to later, only the methods that do not require any additional learning phase.

A potential field approach for point obstacles is proposed in \cite{Kha85} where an obstacle creates a potential field $ U(\vx) $ at the system position $ \vx $.
The perturbation term $ \vvphi (\vx, \vv) $ in this case depends only on the position (and not on the velocity) and is the negative gradient of the potential:
\begin{equation}
    \label{eq:static_gradient}
    \vvphi (\vx, \vv) \equiv \vvphi (\vx) = - \nabla_\vx U(\vx),
\end{equation}
with the potential defined as
\begin{equation}
    \label{eq:static_potential_khatib}
    U_s(\vx) =
        \begin{cases}
            \frac{ \eta }{ 2 } \br{ \frac{ 1 }{ p(\vx) } - \frac{ 1 }{ p_0 } } ^ 2 & \text{if }p(\vx) \le p_0 \\
            0 & \text{if } p(\vx) > p_0
        \end{cases},
\end{equation}
where $ \eta \in \RR_+ $ is a constant gain, $ p_0 \in \RR_+ $ is the influence radius of the obstacle, and $ p(\vx) \in \RR_+ $ is the distance between the obstacle and the system's position.

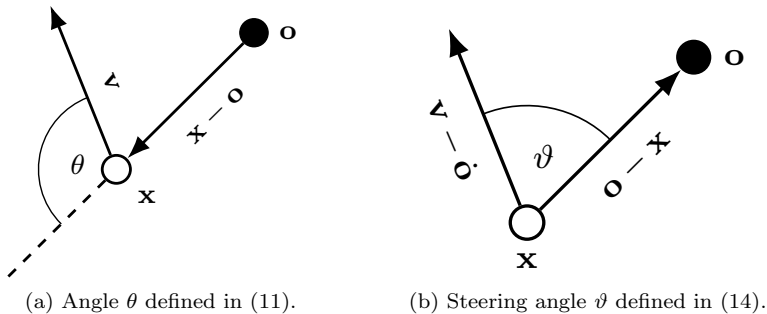
\begin{figure}[htb]
    \centering
    \subfloat[Angle $\theta$ defined in \eqref{eq:cos_theta_def_point}.\label{subfig:theta}]{
    \resizebox{!}{0.3\linewidth}
    {
        \input{angle_scheme.tex}
    }
    }
    \hspace{0.05\linewidth}
    \subfloat[Steering angle $\vartheta$ defined in \eqref{eq:steering_angle_def}.\label{subfig:steering_angle}]{
    \resizebox{!}{0.3\linewidth}
    {
        \input{steering_angle_scheme.tex}
    }
    }
    \caption{Depiction on the definition of angle $\theta$ and $\vartheta$ in \eqref{eq:cos_theta_def_point} and \eqref{eq:steering_angle_def} respectively.
    We remark that the two main differences.
    Firstly, in the steering angle (on the right) the velocity vector takes into consideration also the velocity of the obstacle.
    Secondly, the two angles are complementary, assuming same $\vx - \vo, \vv$, and $\dot{\vo} = \vzero$, since in this case the cosines are opposite: \( \cos \theta = - \cos \vartheta \).
    }
    \label{fig:angle}
\end{figure}

It was pointed out in \cite{PHPS08} that the perturbation term \eqref{eq:static_gradient} obtained using \eqref{eq:static_potential_khatib} as potential may result in non-smooth obstacle behaviors since it does not depend on the velocity $ \vv $ of the system.
Thus, the following `dynamic' (i.e. velocity dependent) potential is proposed
\begin{equation}
    \label{eq:dynamic_potential_park}
    U_d (\vx , \vv) =
    \begin{cases}
        \lambda (- \cos \theta) ^ \beta \frac{ \norm{\vv} }{ p (\vx) } & \text{if } \theta \in \left( \frac{\pi}{2} , \pi \right] \\
        0 & \text{if } \theta \in \sbr{ 0 , \frac{\pi}{2} }
    \end{cases} ,
\end{equation}
where $\lambda, \beta \in \RR_+$ are constant gains, and $ \theta $, depicted in Figure~\ref{subfig:theta}, is the angle taken between the current velocity $\vv$ and the system's position $\vx$ relative to the position $\vo$ of the obstacle:
\begin{equation}
    \label{eq:cos_theta_def_point}
    \cos \theta = \frac{\scalarp{\vv}{\vx - \vo}}{ \norm{\vv} p(\vx) },
\end{equation}
where $ \scalarp{\cdot}{\cdot} $ denotes the standard scalar product in $\RR^d$, and $ p(\vx) $ still denotes the distance between $\vx$ and the obstacle.\\
For potentials depending on both position $\vx$ and velocity $\vv$ of the system, the perturbation term is defined as the negative gradient with respect to the position:
\begin{equation}
    \label{eq:dynamic_gradient}
    \vvphi (\vx, \vv) = - \nabla_\vx U(\vx, \vv).
\end{equation}

The following perturbation term was proposed in \cite{HPPS09}:
\begin{equation}
    \label{eq:perturb_steering_angle}
    \vvphi (\vx, \vv) = \gamma \, \mR \, \vv \, \vartheta \, \exp \br{ -\beta \vartheta },
\end{equation}
where $ \gamma, \beta \in \RR_+ $ are constant gains.
The \emph{steering angle} $ \vartheta $ (depicted in Figure~\ref{subfig:steering_angle}) is defined as
\begin{equation}
    \label{eq:steering_angle_def}
    \vartheta = \arccos \br{ \frac{ \scalarp{ \vo - \vx }{ \vv - \dot{\vo} } }{ \norm{\vo - \vx} \norm{\vv - \dot{\vo}} } },
\end{equation}
where $ \vo $ and $ \dot{\vo} $ are position and velocity of the point obstacle.
Matrix $ \mR $ is defined as the rotation matrix of angle $ \pi / 2 $ with respect to the axis generated by $ (\vo - \vx) \times \vv $, where $\times$ denotes the cross product in $\RR^3$.
This formulation presents an important advantage and two important shortcomings with respect to the previous two approaches.
The advantage is that this formulation guarantees convergence to the goal position if the obstacles are still.
On the other hand, using potential functions \eqref{eq:static_potential_khatib} and \eqref{eq:dynamic_potential_park}, there may be cases in which the system remains `trapped' in a local minima.
However, as defined in \cite{HPPS09}, the matrix $ \mR $ makes sense only in $ \RR^3 $ (and $\RR^2$).
Thus this approach can be used only when DMPs are used in ambient space, and not joint space.
Moreover, formulation \eqref{eq:perturb_steering_angle} does not depend on the distance from the obstacle, and the same `importance' is given to close and far obstacles: this may result in oscillatory behaviors, as pointed out in \cite{GMCDSF19}.

The presented methods work only on point obstacles.
Volumetric obstacles can be modeled using point clouds or by choosing a `critical point' on the surface of the obstacle itself.
However, both these strategies may generate odd behaviors: using a point cloud may result in high computational time, and it is in general hard to decide a priori how dense the point cloud should be; and the use of a critical point (e.g. the closer one) can result in non-smooth behaviors since this point is constantly changing.\\
For this reason, we proposed, in \cite{GMCDSF19}, a novel method to implement volumetric obstacle avoidance, based on the theory of \emph{superquadric potential functions} \cite{Vol90}.
In this approach, the following static potential function is defined
\begin{equation}
    \label{eq:static_potential_volume}
    U_S (\vx) = \frac{ A \exp \br{ - \eta \, C(\vx) } }{ C(\vx) },
\end{equation}
where $ A, \eta \in \RR_+ $ are gain parameters.
Functional $ C : \RR^d \to \RR $ is an \emph{isopotential} function that vanishes on the surface of the obstacle.
In $ \RR^3 $ we defined it as
\begin{equation}
    \label{eq:isopot_gen_ell}
    C (\vx) = \br{
        \br{ \frac{x_1}{ f_1(\vx) } } ^ {2n} +
        \br{ \frac{x_2}{ f_2(\vx) } } ^ {2n}
    } ^ \frac{2m}{2n} +
    \br{ \frac{x_3}{ f_3(\vx) } } ^ {2m} - 1.
\end{equation}
that vanishes on the surface of a \emph{generalized ellipsoid}.
By tuning parameters $m, n$ and functions $ f_1, f_2, f_3 $ it is possible to model obstacles of any shape (their boundary will be the zero-level set of \eqref{eq:isopot_gen_ell}).
We remark that any function $ C $ satisfying:
\begin{description}
    \item[I1.\label{isopot_prop_1}] The boundary of the obstacle is the zero-level set of the isopotential;
    \item[I2.\label{isopot_prop_2}] The value of $C$ increases when the distance from the obstacle increases;
\end{description}
can be used as isopotential in place of \eqref{eq:isopot_gen_ell} when defining \eqref{eq:static_potential_volume}.
This means that this approach can be theoretically used also in joint space.\\
The perturbation term in this approach is defined as in \eqref{eq:static_gradient}: \( \vvphi (\vx, \vv) = \vvphi(\vx) = - \nabla_\vx U_S(\vx) \).


\section{New Potential Function}
\label{sec:contribution}

In this work, we propose a dynamic potential function for volumetric (non-pointwise) obstacles, thus merging the frameworks \eqref{eq:dynamic_potential_park} and \eqref{eq:static_potential_volume}.

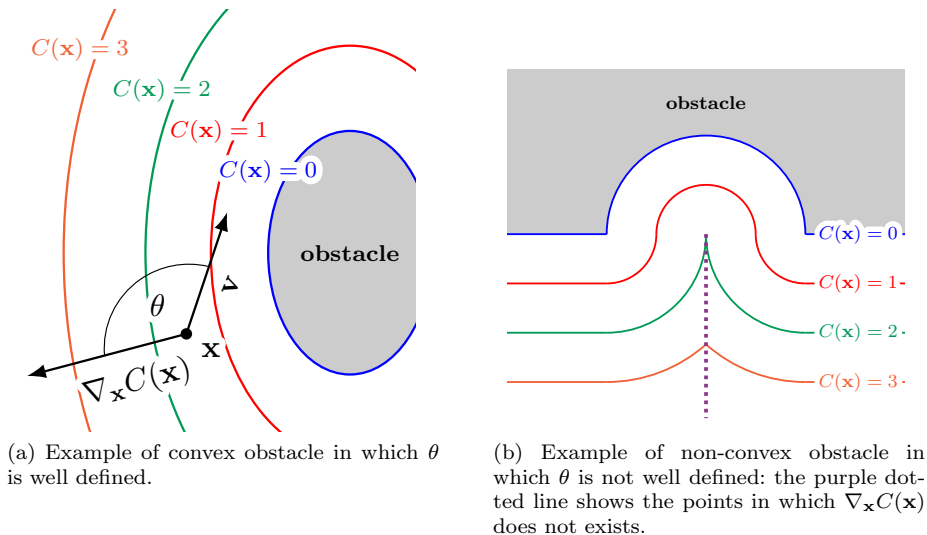
\begin{figure}[htb]
    \subfloat[Example of convex obstacle in which $ \theta $ is well defined. \label{subfig:convex_obst}]{
        \resizebox{0.45\linewidth}{!}{
        \input{grad_orth_convex.tex}
        }
    }
    \hspace{0.05\linewidth}
    \subfloat[Example of non-convex obstacle in which $ \theta $ is not well defined: the purple dotted line shows the points in which $ \nabla_\vx C(\vx) $ does not exists.\label{subfig:nonconvex_obst}]{
        \resizebox{0.45\linewidth}{!}{
        \input{grad_orth_concave.tex}
        }
    }
    \caption{Figure~\ref{subfig:convex_obst} shows how the angle $\theta$ is defined when the gradient $ \nabla_\vx C(\vx) $ of the isopotential exists.
    Figure~\ref{subfig:nonconvex_obst}, instead, shows an example on how non-convex obstacles result in non differentiable isopotentials, and thus it is not possible to define the angle $ \theta $.}
    \label{fig:cos_theta_volume}
\end{figure}

Similarly to \cite{PHPS08}, we aim at designing a potential that satisfies the following three properties:
\begin{description}
    \item[P1.\label{prop_potential_1}] The magnitude of the potential decreases with the distance of the system from the obstacle;
    \item[P2.\label{prop_potential_2}] The magnitude of the potential increases with the velocity of the system $ \norm{ \vv } $ and is zero when the system is not moving;
    \item[P3.\label{prop_potential_3}] The magnitude of the potential decreases with the angle between current velocity direction $\vv / \norm{\vv}$, and the direction towards the obstacle; and, if the system is moving away from the obstacle, the potential should vanish.
\end{description}
To this end, mimiking \eqref{eq:dynamic_potential_park}, we define the dynamic potential function
\begin{equation}
    \label{eq:dynamic_potential_volume}
    U_D(\vx, \vv) =
    \begin{cases}
        \lambda (- \cos \theta) ^ \beta \dfrac{ \norm{\vv} }{ C ^ \eta (\vx) } & \text{if } \theta \in \left( \frac{\pi}{2} , \pi \right] \\
         0 & \text{if } \theta \in \sbr{ 0 , \frac{\pi}{2} }
    \end{cases},
\end{equation}
where $\lambda, \beta \in \RR_+$ are constant gains, and function $ C(\vx) $ is any ispotential satisfying Properties~\ref{isopot_prop_1}~and~\ref{isopot_prop_2} given in Section~\ref{subsec:sota_obstacle}.
The angle $\theta$ is taken between the system's velocity $\vv$ and the direction between system's position $\vx$ and the closer point of the obstacle.
Thanks to Property~\ref{isopot_prop_2}, we have that the gradient $ \nabla_\vx C(\vx) $ of the isopotential $ C(\vx) $, is always perpendicular to the obstacle surface.
Thus, at least for convex obstacles, the angle $ \theta $ can be computed using
\begin{equation}
    \label{eq:angle_for_volumes}
    \cos \theta = \frac{ \scalarp{ \nabla_\vx C(\vx) }{ \vv } }{ \norm{ \nabla_\vx C(\vx) } \norm{ \vv } },
\end{equation}
while it is not well defined for non-convex obstacles.
An intuition for these two observations are given in Figure~\ref{fig:cos_theta_volume}.

\begin{remark}
    For non convex obstacles, some workarounds can be used.
    First, if neither the starting position nor the goal are in the `holes' of the obstacle, that is they are not in the convex hull of the obstacle, then the convex hull itself can be used as obstacle.
    Second, one can think at relaxing the concept of gradient to allow sub-differentials.
    In such case, the sub-gradient exists but it is not unique.
    Thirdl a non-convex obstacle can be split in multiple convex components, and each component would generate its own potential.
\end{remark}

The potential defined in \eqref{eq:dynamic_potential_volume} clearly satisfies Properties~\ref{prop_potential_1}--\ref{prop_potential_3}.
Indeed, the potential is a decreasing function of both $ C(\vx) $ and $\theta$, thus it satisfies \ref{prop_potential_1} and \ref{prop_potential_3}.
Moreover, it is an increasing function of $ \| \vv \| $, and thus it satisfies also \ref{prop_potential_2}.

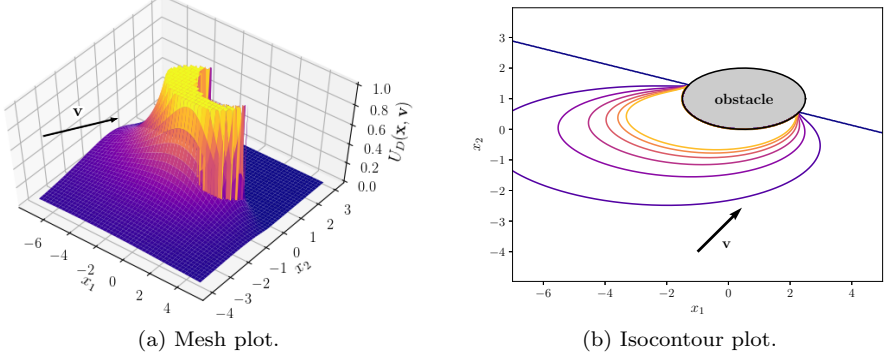
\begin{figure}[htb]
    \subfloat[Mesh plot.\label{subfig:pot_mesh}]{\resizebox{0.45\linewidth}{!}{\input{potential_function.tex}}}
    \hspace{0.05\linewidth}
    \subfloat[Isocontour plot.\label{subfig:pot_isoc}]{\resizebox{0.45\linewidth}{!}{\input{isopotential.tex}}}
    \caption{Example of the dynamic potential $ U_D (\vx ,\vv) $ given in \eqref{eq:dynamic_potential_volume} for an ellipse in $ \RR^2 $.
    The velocity vector $ \vv $ is set to $ \vv = [1, 1]\transpose $.
    The gains are set to $ \lambda = 2, \beta = 2 ,$ and $ \eta = 1 $.
    The ellipse has center in $\sbr{\nicefrac{1}{2}, 1}\transpose$, horizontal axis $2$, and vertical axis $1$.
    In both figures, the potential has been cropped at value $1$ for display purposes: it goes to infinity on half of the boundary of the obstacle (on the other half, the system goes away from the obstacle, so the potential is zero).
    }
    \label{fig:dyn_pot_vol_exp}
\end{figure}

\noindent
As an example, we show in Figure~\ref{fig:dyn_pot_vol_exp} the potential \eqref{eq:dynamic_potential_volume} for an elliptic obstacle in $ \RR ^ 2 $, whose isopotential is
\begin{equation*}
    C(\vx) = \br{ \frac{ x_1 - \widehat{x}_1 }{ \ell_1 } } ^ 2 + \br{ \frac{ x_2 - \widehat{x}_2 }{ \ell_2 } } ^ 2 - 1,
\end{equation*}
where the center of the ellipse is $ \widehat{\vx} = [\widehat{x}_1, \widehat{x}_2]\transpose $ and the horizontal and vertical axes are, respectively, $\ell_1$ and $\ell_2$.
The perturbation term is defined as in \eqref{eq:dynamic_gradient} and is computed as follows:
\begin{align*}
    \vvphi (\vx, \vv)
        & = - \nabla_\vx \big( U_D(\vx, \vv ) \big) \\
        & = - \nabla_\vx \br{\lambda (- \cos \theta) ^ \beta \frac{ \norm{\vv} }{ C ^ \eta (\vx) }} \\
        & = - \frac{\lambda \norm{\vv} (- \cos \theta)^{\beta - 1}}{ C ^ \eta (\vx) } \br{ - \beta \, \nabla_\vx (\cos \theta) + \frac{\eta \cos\theta}{ C(\vx) } \nabla_\vx (C(\vx)) }.
\end{align*}
The term $ \nabla_\vx (\cos\theta) $ can be computed as
\begin{align}
    \nabla_\vx (\cos\theta)
        & = \nabla_\vx \br{ \frac{ \scalarp{ \nabla_\vx C(\vx) }{ \vv } }{ \norm{ \nabla_\vx C(\vx) } \norm{ \vv } } } \nonumber\\
        & \!\begin{multlined}
            = \frac{ 1 }{ \norm{\vv} \norm{C(\vx)}^2 } \Big( \norm{\nabla_\vx C(\vx)} \, \nabla_\vx \big( \scalarp{ \nabla_\vx C(\vx) }{ \vv } \big) - \qquad\qquad \\
            \scalarp{ \nabla_\vx C(\vx) }{ \vv } \, \nabla_\vx \big( \norm{ \nabla_\vx C(\vx) } \big) \Big).
        \end{multlined}
        \label{eq:nabla_cos}
\end{align}

\noindent
For instance, let us consider the case in which the isopotential $ C(\vx) $ is an ellipsoid in $ \RR ^ 3 $ with center $ \widehat{\vx} = [\widehat{x}_1, \widehat{x}_2, \widehat{x}_3]\transpose $, and axes $(\ell_1, \ell_2, \ell_3)$,
\begin{equation*}
    C (\vx) =
        \br{ \frac{ x_1 - \widehat{x}_1 }{\ell_1} } ^ 2 +
        \br{ \frac{ x_2 - \widehat{x}_2 }{\ell_2} } ^ 2 +
        \br{ \frac{ x_3 - \widehat{x}_3 }{\ell_3} } ^ 2.
\end{equation*}
In this case, the gradient is
\begin{equation*}
    \nabla_\vx C(\vx) = 2
        \begin{bmatrix}
            \frac{ x_1 - \widehat{x}_1 }{ \ell_1 ^ 2 } \\[5pt]
            \frac{ x_2 - \widehat{x}_2 }{ \ell_2 ^ 2 } \\[5pt]
            \frac{ x_3 - \widehat{x}_3 }{ \ell_3 ^ 2 }
        \end{bmatrix}
        .
\end{equation*}
The quantities $ \nabla_\vx \big( \scalarp{ \nabla_\vx C(\vx) }{ \vv } \big) $ and $ \nabla_\vx \big( \norm{ \nabla_\vx C(\vx) } \big) $ in \eqref{eq:nabla_cos} read, respectively,
\begin{equation*}
    \nabla_\vx \big( \scalarp{ \nabla_\vx C(\vx) }{ \vv } \big) = 2
        \begin{bmatrix}
            \frac{v_1}{\ell_1 ^ 2} \\[5pt]
            \frac{v_2}{\ell_2 ^ 2} \\[5pt]
            \frac{v_3}{\ell_3 ^ 2}
        \end{bmatrix},
    \quad
    \text{and}
    \quad
    \nabla_\vx \big( \norm{ \nabla_\vx C(\vx) } \big) =
    \frac{4}{\norm{ \nabla_\vx C(\vx) }}
    \begin{bmatrix}
        \frac{ x_1 - \widehat{x}_1 }{ \ell_1 ^ 4 } \\[5pt]
        \frac{ x_2 - \widehat{x}_2 }{ \ell_2 ^ 4 } \\[5pt]
        \frac{ x_3 - \widehat{x}_3 }{ \ell_3 ^ 4 }
    \end{bmatrix}.
\end{equation*}

\begin{table}[htb]
    \centering
    \renewcommand{\arraystretch}{1.1}
    \begin{tabular}{cccccc}
        \hline
        \multirow{2}{*}{\textbf{Method}} & \textbf{Type of}& \textbf{Space of} & \textbf{Type of} & \textbf{Distance} & \textbf{Guaranteed}\\
        & \textbf{obstacle} & \textbf{definition} & \textbf{potential} & \textbf{dependent} & \textbf{convergence}\\ \hline \hline
        Static& \multirow{2}{*}{Point} & \multirow{2}{*}{\underline{$\RR^d, \, d\in \NN$}}& \multirow{2}{*}{Static}& \multirow{2}{*}{\underline{Yes}} & \multirow{2}{*}{No}\\ 
        potential \eqref{eq:static_potential_khatib}\\\hline
        Dynamic& \multirow{2}{*}{Point} & \multirow{2}{*}{\underline{$\RR^d, \, d\in \NN$}}& \multirow{2}{*}{\underline{Dynamic}} & \multirow{2}{*}{\underline{Yes}} & \multirow{2}{*}{No}\\
        potential \eqref{eq:dynamic_potential_park} \\\hline
        Steering & \multirow{2}{*}{Point} & \multirow{2}{*}{$\RR^2, \, \RR^3$} & \multirow{2}{*}{\underline{Dynamic}} & \multirow{2}{*}{No}& \multirow{2}{*}{\underline{Yes}}\\
        angle \eqref{eq:perturb_steering_angle}\\\hline
        Static& \multirow{2}{*}{\underline{Volumes}} & \multirow{2}{*}{\underline{$\RR^d, \, d\in \NN$}}& \multirow{2}{*}{Static}& \multirow{2}{*}{\underline{Yes}} & \multirow{2}{*}{No}\\
        potential \eqref{eq:static_potential_volume}\\\hline
        Dynamic& \multirow{2}{*}{\underline{Volumes}} & \multirow{2}{*}{\underline{$\RR^d, \, d\in \NN$}}& \multirow{2}{*}{\underline{Dynamic}} & \multirow{2}{*}{\underline{Yes}} & \multirow{2}{*}{No}\\
        potential \eqref{eq:dynamic_potential_volume}\\\hline
    \end{tabular}
    \caption{Summary of the properties of various methods for obstacle avoidance.
    The desired properties are underlined.}
    \label{tab:summary_oa_approaches}
\end{table}

We emphasize that the proposed approach encompass most of the desired properties of obstacle avoidance frameworks for DMPs since it is: dynamic, well defined in $ \RR^n $, volumetric, and distance dependent.
On the other hand, with this approach it is not guaranteed the convergence to the goal since local minima may arise.
However, as we already pointed out in \cite{GMCDSF19}, it is unlikely to encounter a local minimum, and if it happens, a perturbation term pushing the trajectory out of it can be easily added to the DMP formulation.\\
In Table~\ref{tab:summary_oa_approaches} a summary of the properties of both the approaches presented in Section~\ref{subsec:sota_obstacle} and the proposed approach is given.
From this, it is possible to observe that the proposed method is the one satisfying the greatest number of desirable properties.

\begin{flushleft}
    \emph{Remark.} Formulations \eqref{eq:cos_theta_def_point} and \eqref{eq:angle_for_volumes} do not take into account the velocity of the obstacle.
    However, it is straightforward to extend the definition to this case by simply substituting $ \vv $ with $ \vv - \dot{\vo} $, where $\dot{\vo}$ denotes the velocity of the obstacle.
\end{flushleft}


\section{Results}
\label{sec:results}


\subsection{Synthetic Experiments}
\label{subsec:result_synth}

\begin{table}[tb]
    \centering
    \renewcommand{\arraystretch}{1.2}
    \begin{tabular}{cc}\hline
        \textbf{Method} & \textbf{Hyper - parameters} \\ \hline \hline
        Static potential \eqref{eq:static_potential_khatib}     & $p_0 = 0.1, \, \eta = 1$ \\ \hline
        Dynamic potential \eqref{eq:dynamic_potential_park}     & $\lambda=0.2, \, \beta=2$ \\ \hline
        Steering angle \eqref{eq:perturb_steering_angle}        & $\gamma=20, \, \beta=3$ \\ \hline
        Static potential \eqref{eq:static_potential_volume}     & $A = 10, \, \eta = 1$ \\ \hline
        Dynamic potential \eqref{eq:dynamic_potential_volume}   & $\lambda=10, \, \beta = 2, \, \eta = \nicefrac{1}{2}$ \\ \hline
    \end{tabular}
    \caption{Hyper-parameters for obstacle avoidance methods.}
    \label{tab:hyperparam_one_obst}
\end{table}

In this Section, we test and compare the behaviors of the approaches recalled in Section~\ref{subsec:sota_obstacle} and our novel approach, presented in Section~\ref{sec:contribution}, performing the same test we performed in \cite{GMCDSF19}.
In the first test, we show the behaviors of all the methods in the presence of a single obstacle (an ellipse).
For the point obstacle methods, the obstacle is modeled using a point cloud on the boundary of the obstacle itself.
We then add a second obstacle (a circle) to the previous scenario.

\begin{figure}[htb]
    \centering
    \subfloat[Static potential \eqref{eq:static_potential_khatib}.
    \label{subfig:one_obst_point_static}]{\includegraphics[width=0.30\linewidth]{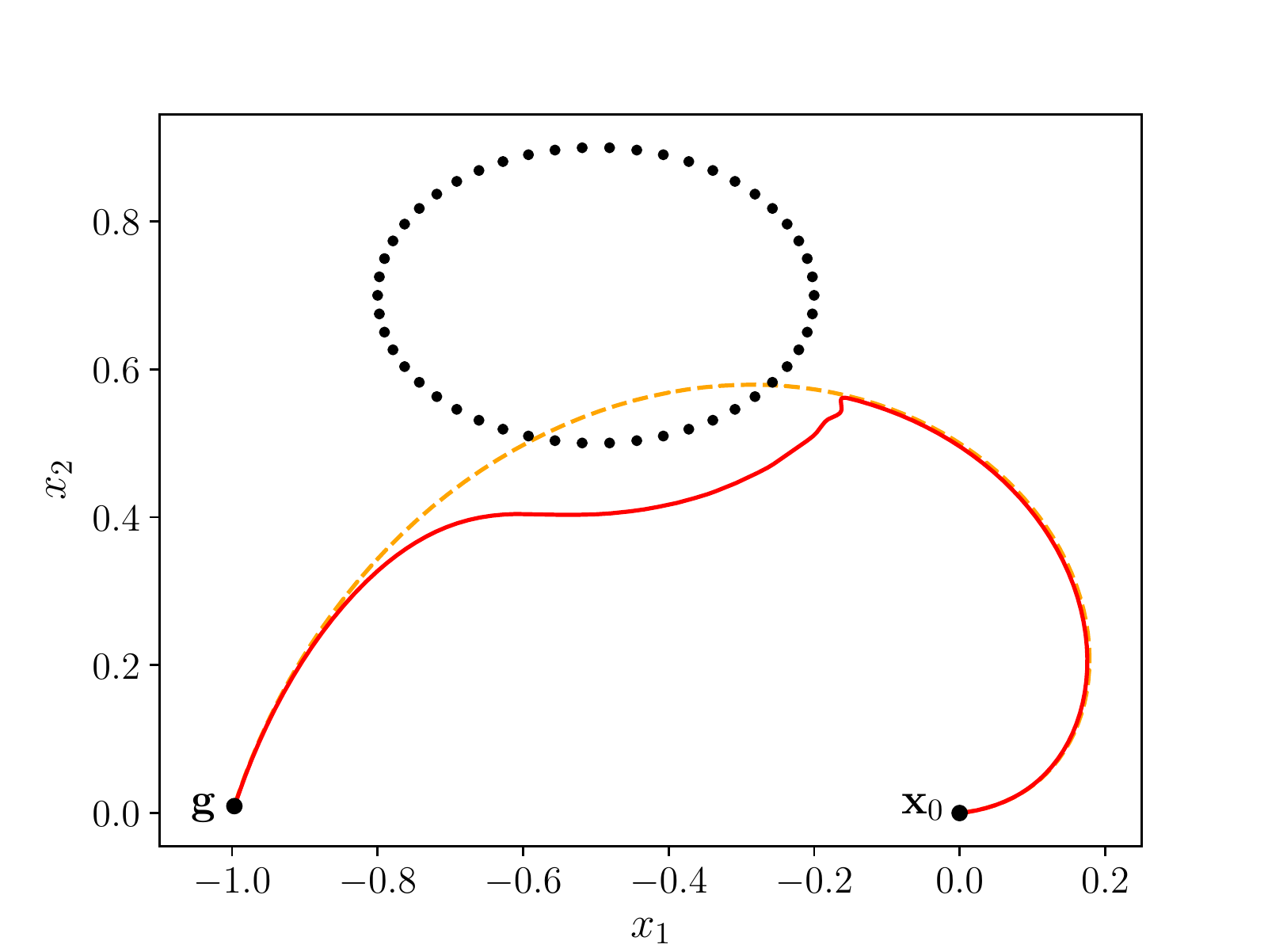}}
    \subfloat[Dynamic potential \eqref{eq:dynamic_potential_park}.
    \label{subfig:one_obst_point_dynamic}]{\includegraphics[width=0.30\linewidth]{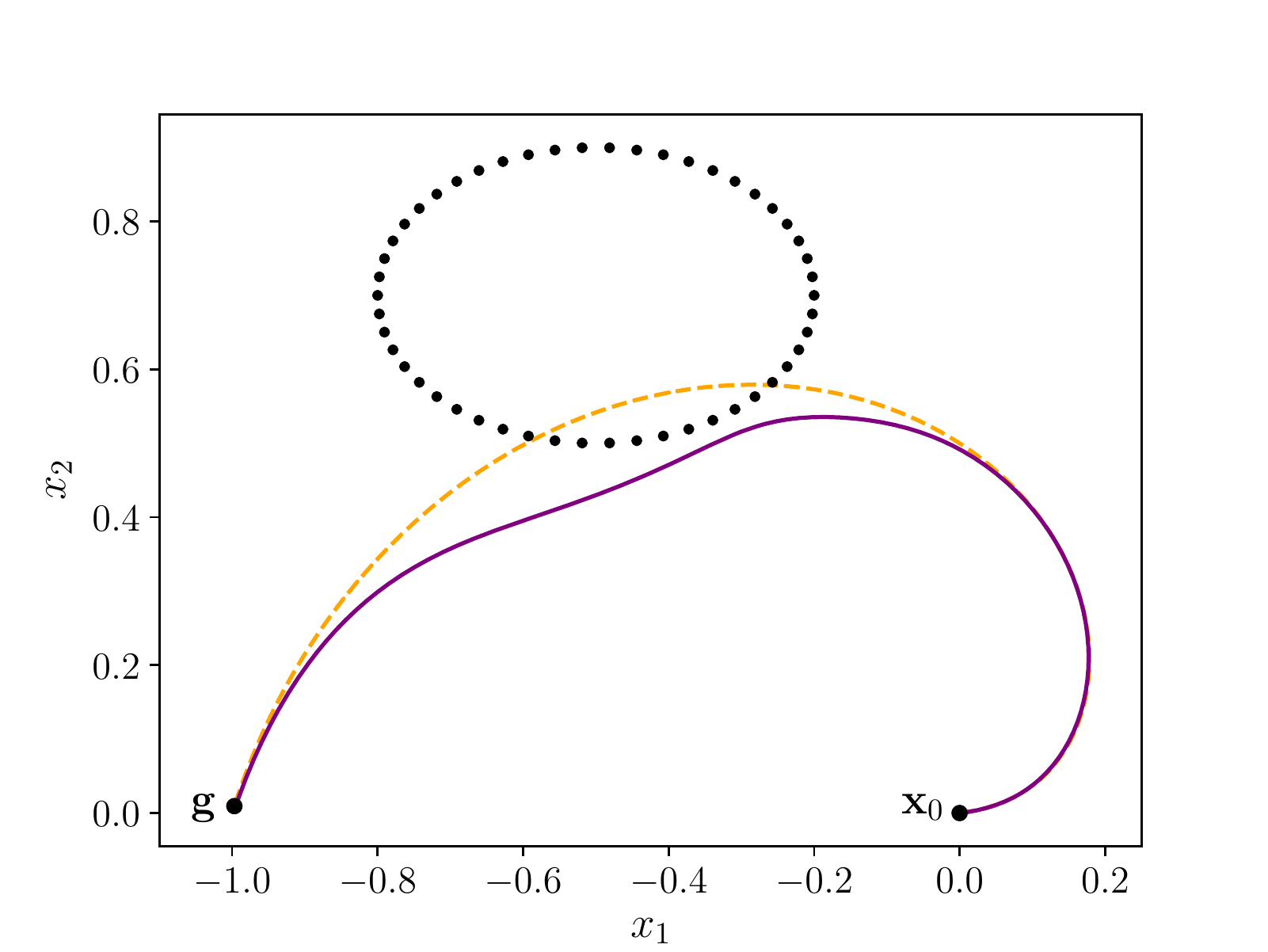}}
    \subfloat[Steering angle \eqref{eq:perturb_steering_angle}.
    \label{subfig:one_obst_point_steering}]{\includegraphics[width=0.30\linewidth]{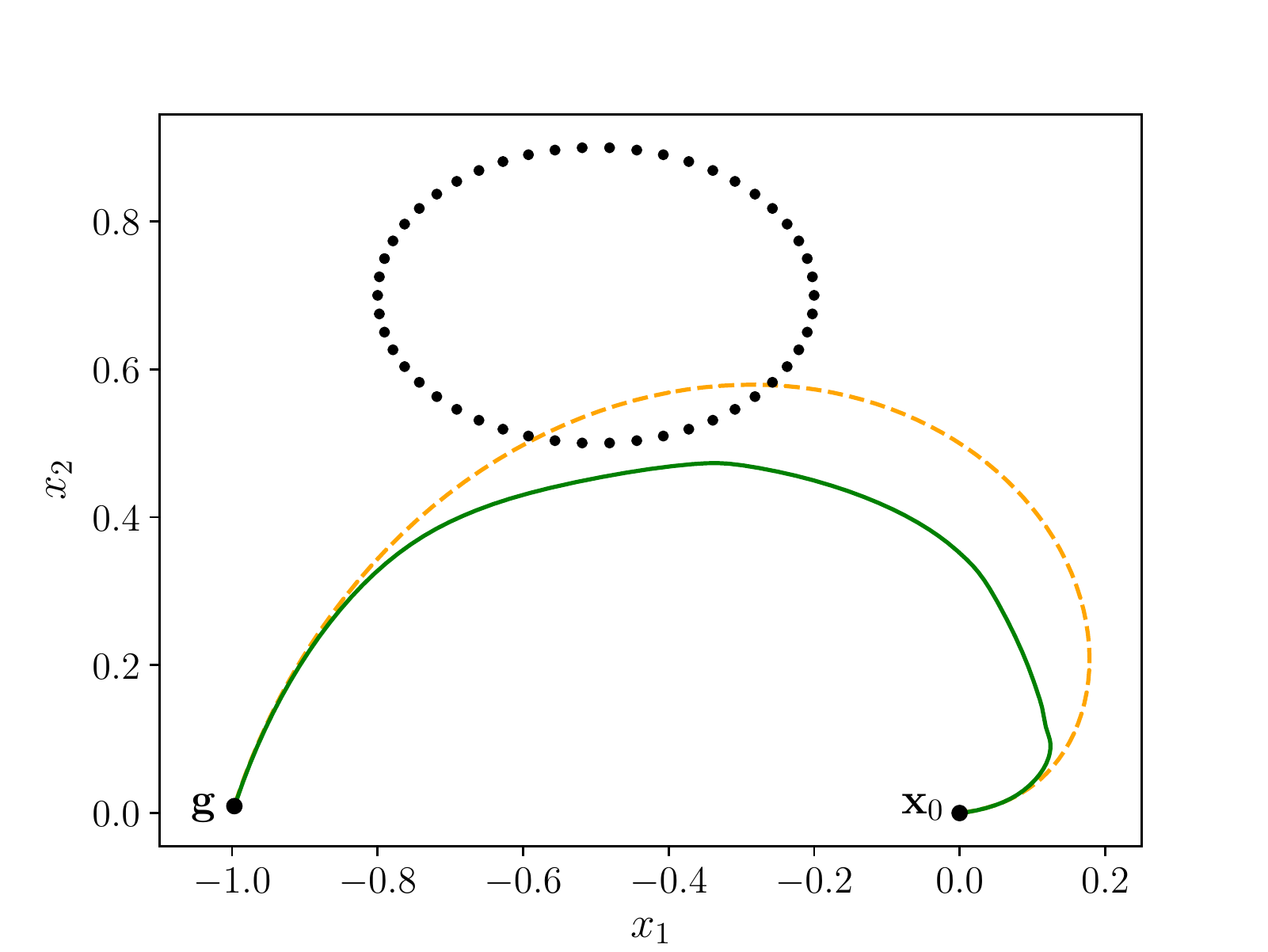}}\\
    \subfloat[Static potential \eqref{eq:static_potential_volume}.
    \label{subfig:one_obst_volume_static}]{\includegraphics[width=0.45\linewidth]{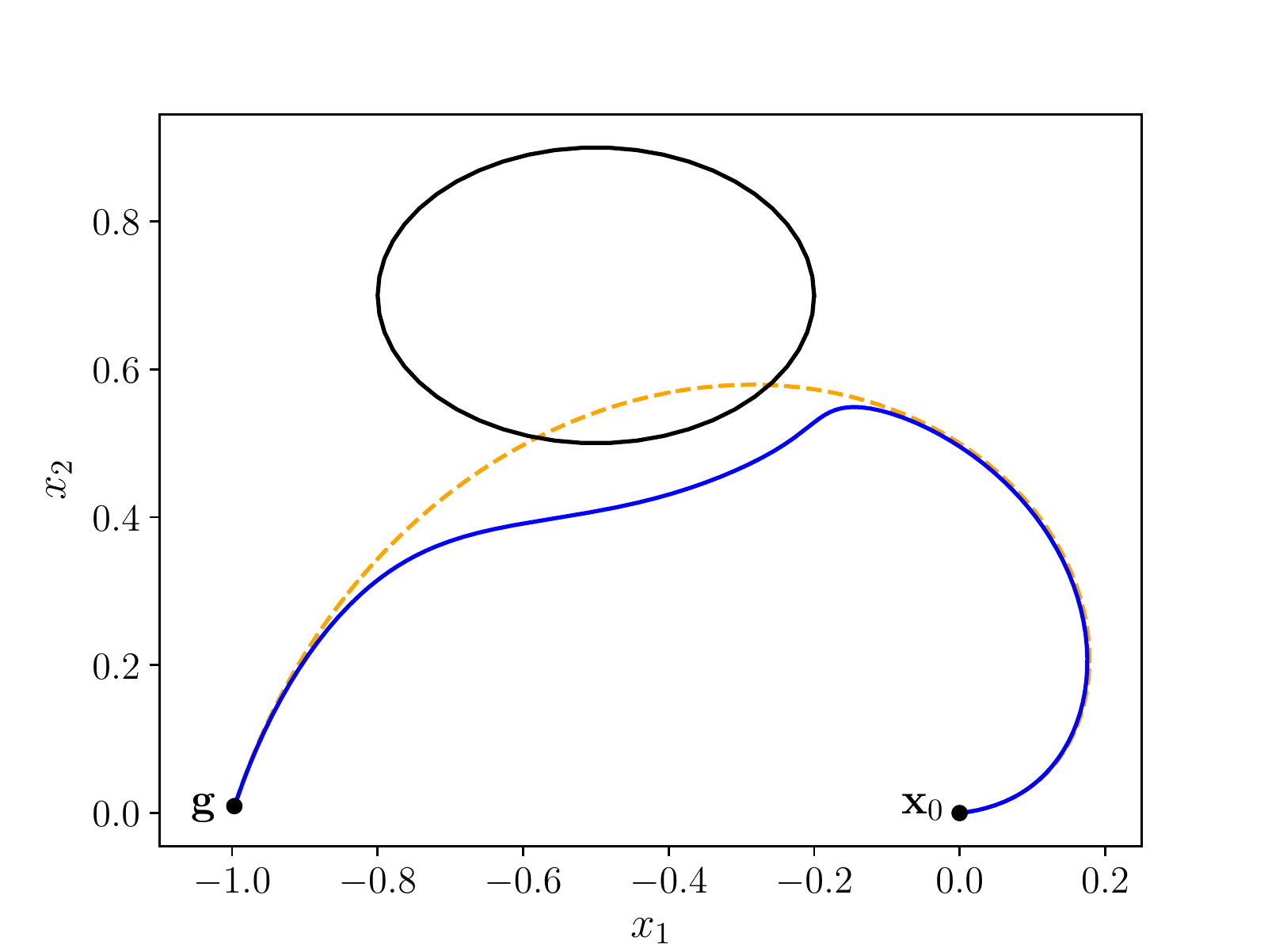}}
    \hspace{0.03\linewidth}
    \subfloat[Dynamic potential \eqref{eq:dynamic_potential_volume}.
    \label{subfig:one_obst_volume_dynamic}]{\includegraphics[width=0.45\linewidth]{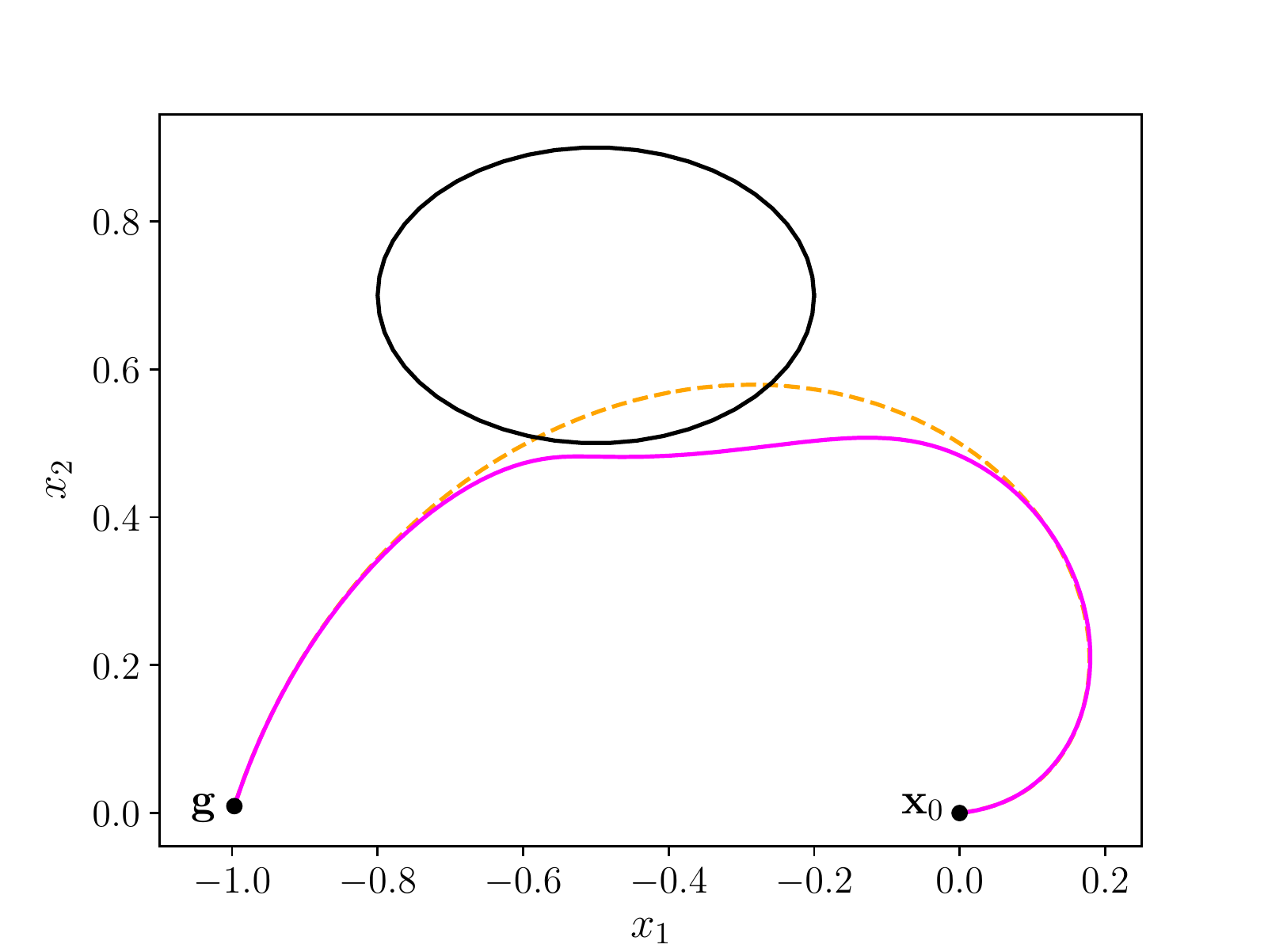}}
    \caption{Obstacle avoidance behavior for the methods recalled in Section~\ref{subsec:sota_obstacle} and the proposed method from Section~\ref{sec:contribution}.
    In all plots, the dashed orange line shows the desired trajectory, while the full colored line shows the adaptation of the DMP to the presence of the obstacle.
    In the three top figures, the black dots mark the point obstacles used as mesh.
    In the two bottom figures, the boundary of the obstacle is plotted using the full black line.}
    \label{fig:one_obst_test}
\end{figure}

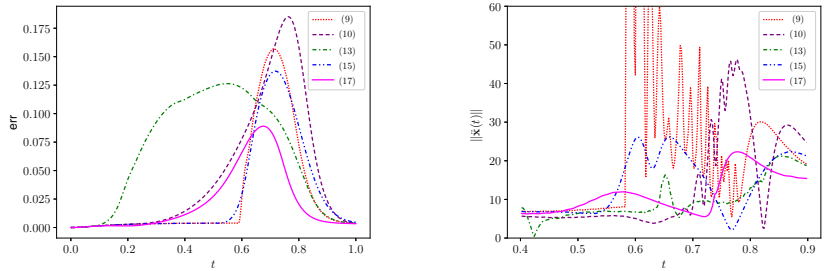
\begin{figure}[htb]
    \centering
    \subfloat[Plot of the distance (in 2-norm) between the desired trajectory and the executed one.\label{subfig:one_obst_err}]{\resizebox{0.45\linewidth}{!}{
        \input{one_obst_err.tex}
    }
    }
    \hspace{0.02\linewidth}
    \subfloat[Plot of the norm of the acceleration of the executed DMP.\label{subfig:one_obst_acc}]{\resizebox{0.45\linewidth}{!}{
        \input{one_obst_acc.tex}
    }
    }
    \caption{For tests depicted in Figure~\ref{fig:one_obst_test}, plot of the distance between desired and executed trajectory (left), and of the norm of the acceleration (right) as functions of time.}
\end{figure}

In the first test, we generate the following trajectory in the plane: \( (x_1(t), x_2(t)) = (t \, \cos(\pi\, t), t \, \sin(\pi\, t)) \), $ t \in [0, 1] $.
Then, we learn a DMP with elastic and damping constants, respectively, $ \mK = K\, \idmtrx_2 $ and $ \mD = D\,\idmtrx_2 $, where $\idmtrx_2$ denotes the $ 2\times 2 $ identity matrix, and $K$ and $D$ have values $ K = 1050 $ and $ D  =2 \, \sqrt{K} \approx 65 $.
In this test, the obstacle is an ellipse centered in $ (-0.5, 0.7) $ with semi-axis $0.3$ and $0.2$.
For the tests done using point-wise obstacle avoidance methods \eqref{eq:static_potential_khatib}, \eqref{eq:dynamic_potential_park}, and \eqref{eq:perturb_steering_angle}, the boundary of the obstacle is discretized using fifty equally distributed points.
The hyper-parameters for all the methods are given in Table~\ref{tab:hyperparam_one_obst}.
The resulting trajectories are shown in Figure~\ref{fig:one_obst_test}.
From this first test, we compute, at each time $t$ how much the trajectory deviate from the learned behavior in order to avoid the obstacle.
This ``error'' is computed as
\begin{equation*}
    \varepsilon (t) = \| \vect{x}_{\text{true}}(t) - {\vect{x}}(t) \|,
\end{equation*}
where $ \vect{x}_{\text{true}}(t) $ is the learned trajectory, and $ {\vect{x}}(t) $ is the adapted behavior.
In Figure~\ref{subfig:one_obst_err} it is possible to observe that the proposed method results in the trajectory that deviate less from the learned one.\\
To discuss the smoothness of the different behaviors, we compute, at each time $t$, the norm $ \norm{\ddot{\vx}(t)} $ of the acceleration $ \ddot{\vx}(t) $ of the adapted trajectory.
As shown in Figure~\ref{subfig:one_obst_acc}, we see that the proposed method results in the less oscillatory behavior of $ \norm{\ddot{\vx}(t)} $.
This last aspect makes the proposed method the most stable one when controlling the position of a robot.

In summary, the proposed dynamic potential \eqref{eq:dynamic_potential_volume} gives both the smoother behavior and the trajectory that remains closer to the learned one between al the methods we presented in Section~\ref{subsec:sota_obstacle}, thus making it the most suitable in real applications.


\begin{figure}[htb]
    \centering
    \subfloat[Static potential \eqref{eq:static_potential_khatib}.
    \label{subfig:two_obst_point_static}]{\includegraphics[width=0.30\linewidth]{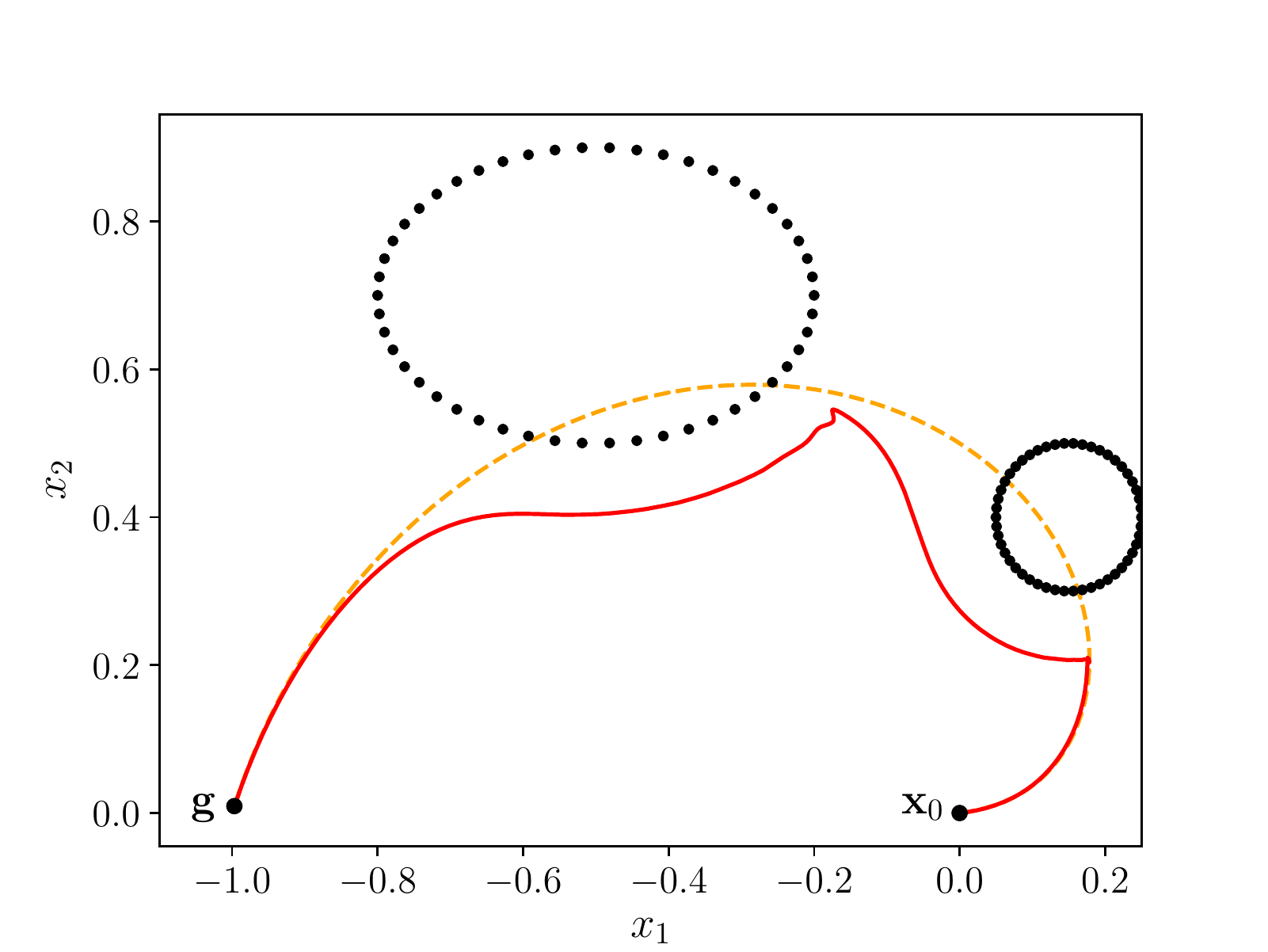}}
    \subfloat[Dynamic potential \eqref{eq:dynamic_potential_park}.
    \label{subfig:two_obst_point_dynamic}]{\includegraphics[width=0.30\linewidth]{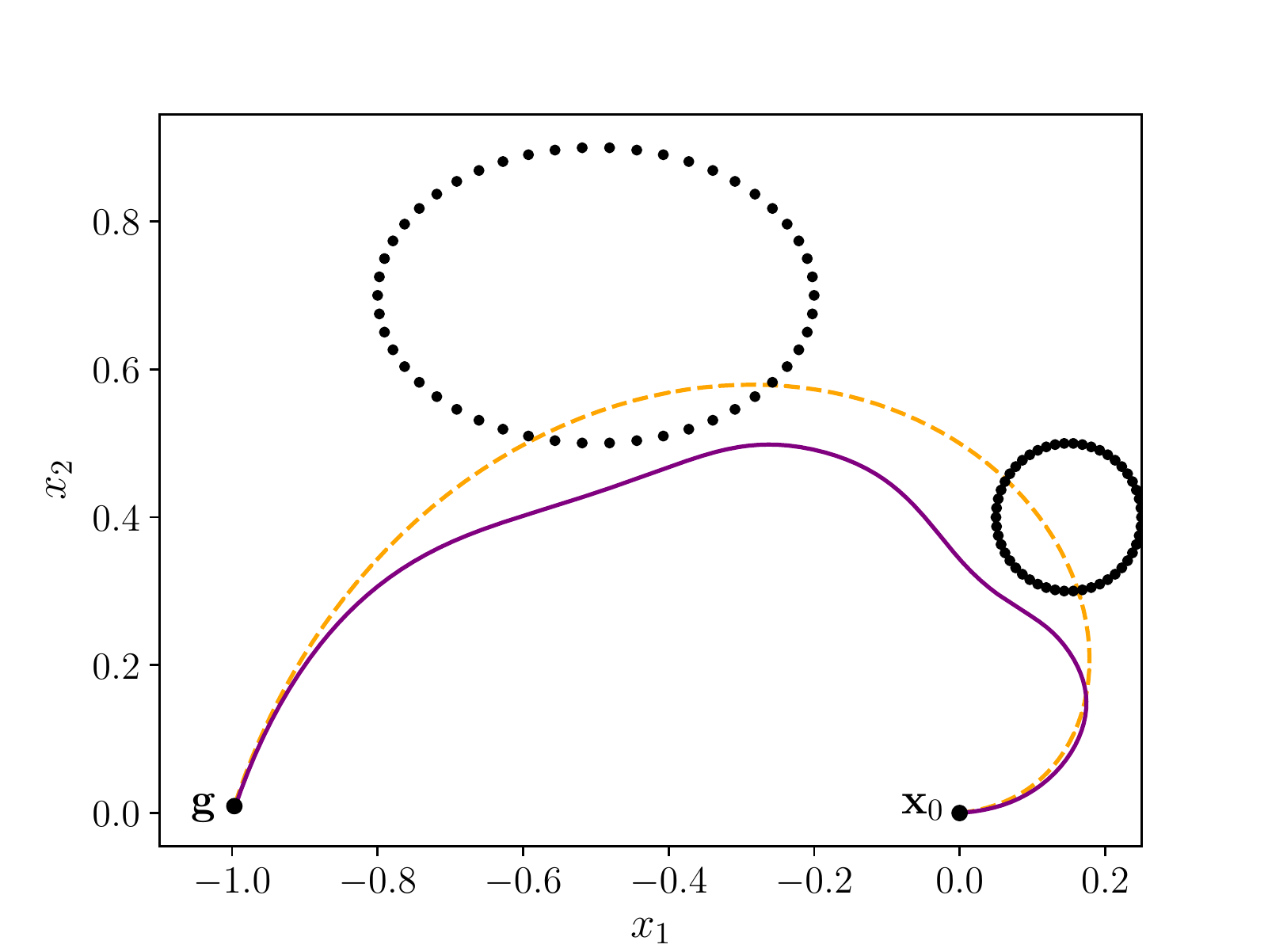}}
    \subfloat[Steering angle \eqref{eq:perturb_steering_angle}.
    \label{subfig:two_obst_point_steering}]{\includegraphics[width=0.30\linewidth]{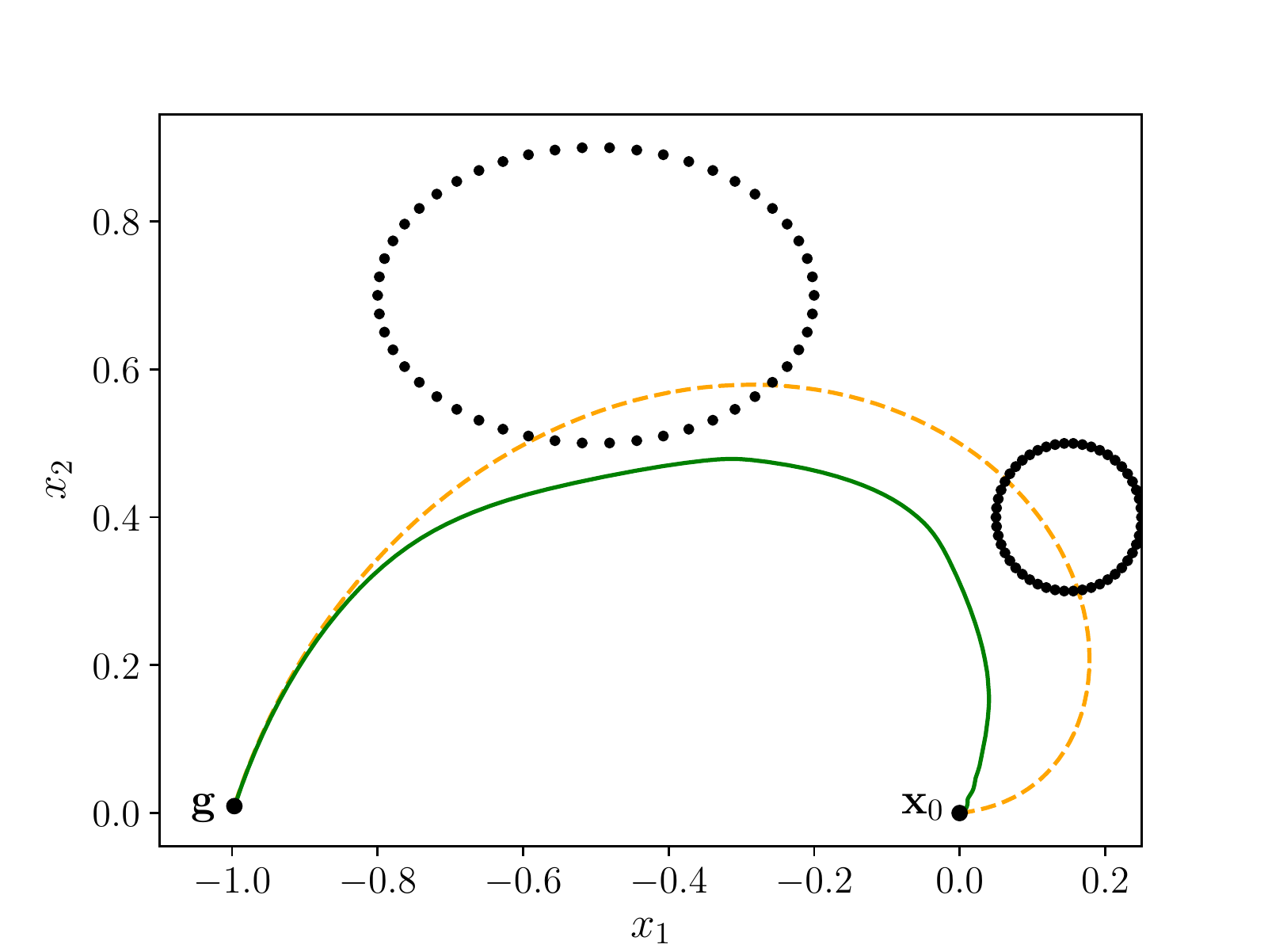}}\\
    \subfloat[Static potential \eqref{eq:static_potential_volume}.
    \label{subfig:two_obst_volume_static}]{\includegraphics[width=0.45\linewidth]{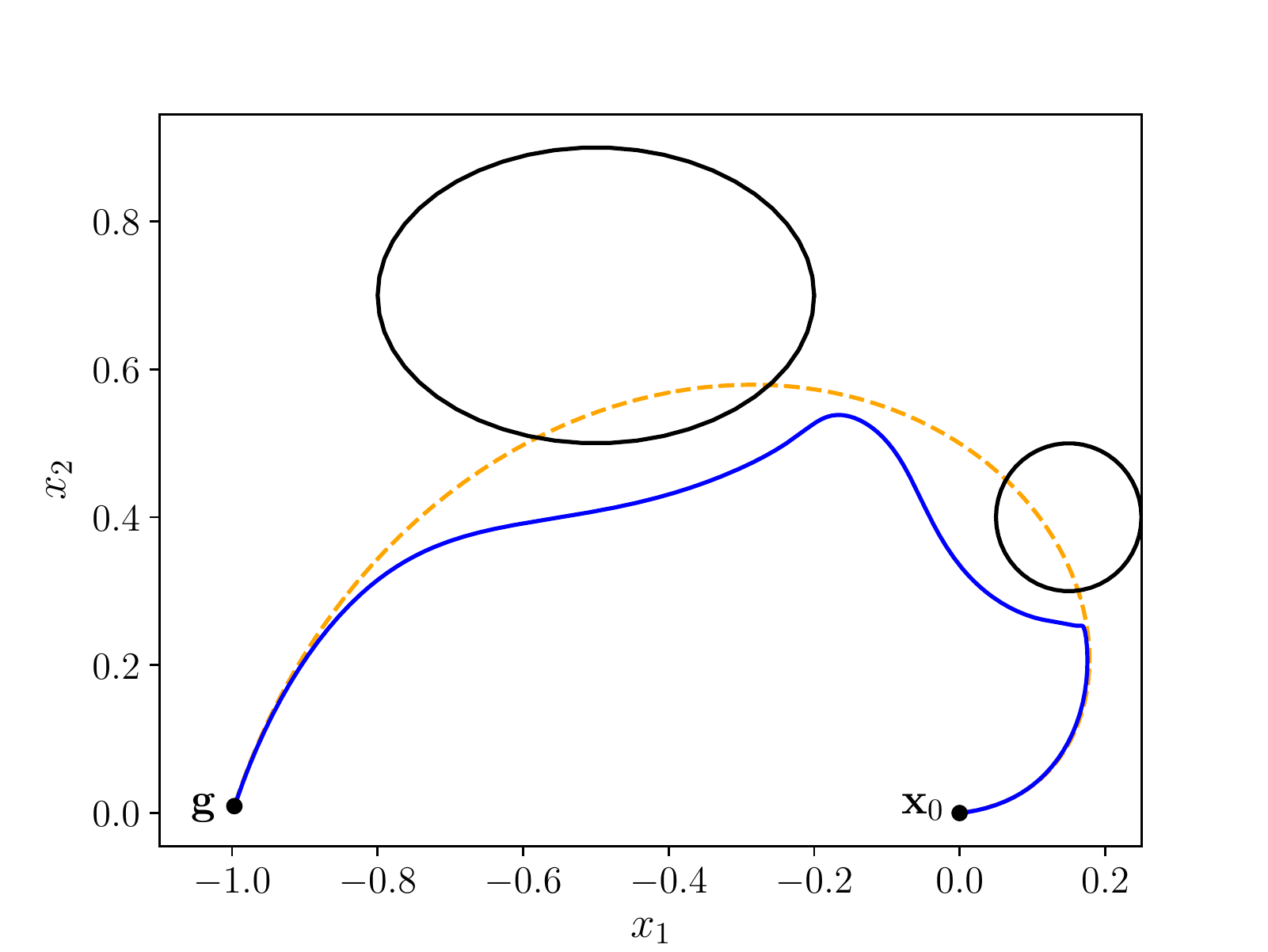}}
    \hspace{0.03\linewidth}
    \subfloat[Dynamic potential \eqref{eq:dynamic_potential_volume}.
    \label{subfig:two_obst_volume_dynamic}]{\includegraphics[width=0.45\linewidth]{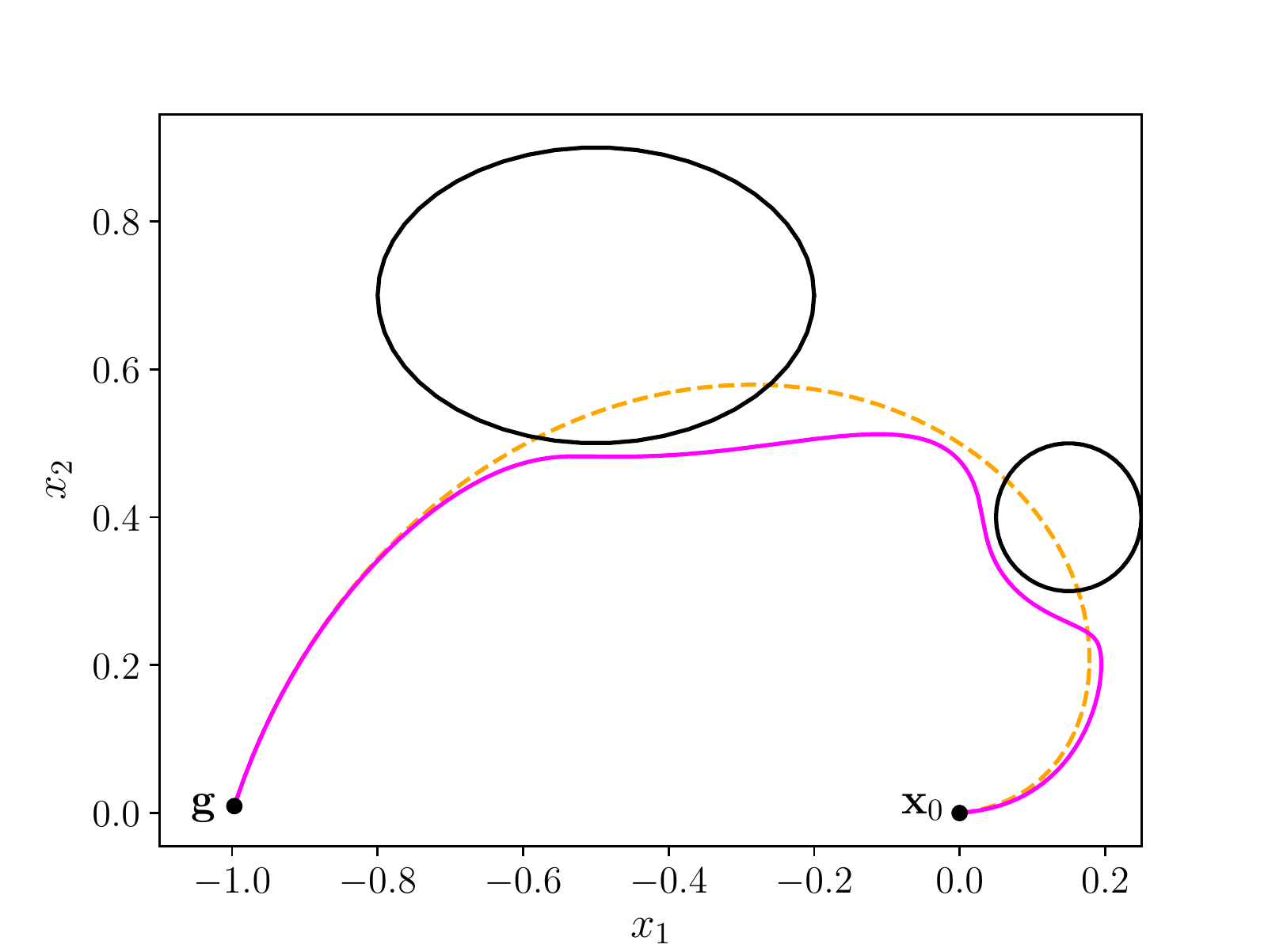}}
    \caption{Obstacle avoidance behavior for the methods recalled in Section~\ref{subsec:sota_obstacle} and the proposed method from Section~\ref{sec:contribution}.
    In all plots, the dashed orange line shows the desired trajectory, while the full colored line shows the adaptation of the DMP to the presence of the obstacle.
    In the three top figures, the black dots mark the point obstacles used as mesh.
    In the two bottom figures, the boundary of the obstacle is plotted using the full black line.}
    \label{fig:two_obst_test}
\end{figure}

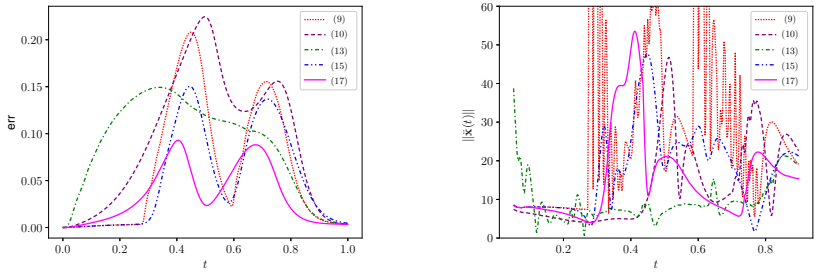
\begin{figure}[htb]
    \centering
    \subfloat[Plot of the distance (in 2-norm) between the desired trajectory and the executed one.\label{subfig:two_obst_err}]{\resizebox{0.45\linewidth}{!}{
        \input{two_obst_err.tex}
    }
    }
    \hspace{0.02\linewidth}
    \subfloat[Plot of the norm of the acceleration of the executed DMP.\label{subfig:two_obst_acc}]{\resizebox{0.45\linewidth}{!}{
        \input{two_obst_acc.tex}
    }
    }
    \caption{For tests depicted in Figure~\ref{fig:two_obst_test}, plot of the distance between desired and executed trajectory (left), and of the norm of the acceleration (right) as functions of time.}
\end{figure}

As second synthetic experiment, we maintain the same conditions of the experiment before (desired curve, as well as DMP and obstacles' hyper-parameters), and we add a second obstacle.
This new obstacle is a circle centered in $ (0.15, 0.4) $ and with radius $0.1$.
For the point-wise obstacle avoidance methods, the circumference is discretized with fifty equally distributed nodes.
Figure~\ref{fig:two_obst_test} shows the adaptation of the DMP to the presence of the obstacles, Figure~\ref{subfig:two_obst_err} shows the distance between desired trajectory and DMP, and Figure~\ref{subfig:two_obst_acc} shows the 2-norm of the acceleration of the DMP as function of time.

\noindent
Also in this test, it is possible to observe that the proposed method still gives the trajectory that less deviate from the learned one, while maintaining the less oscillatory behavior at acceleration level.


\subsection{Experiments with robots in simulation}
\label{subsec:sim_robot}

\begin{figure}
    \centering
    \includegraphics[scale=0.2]{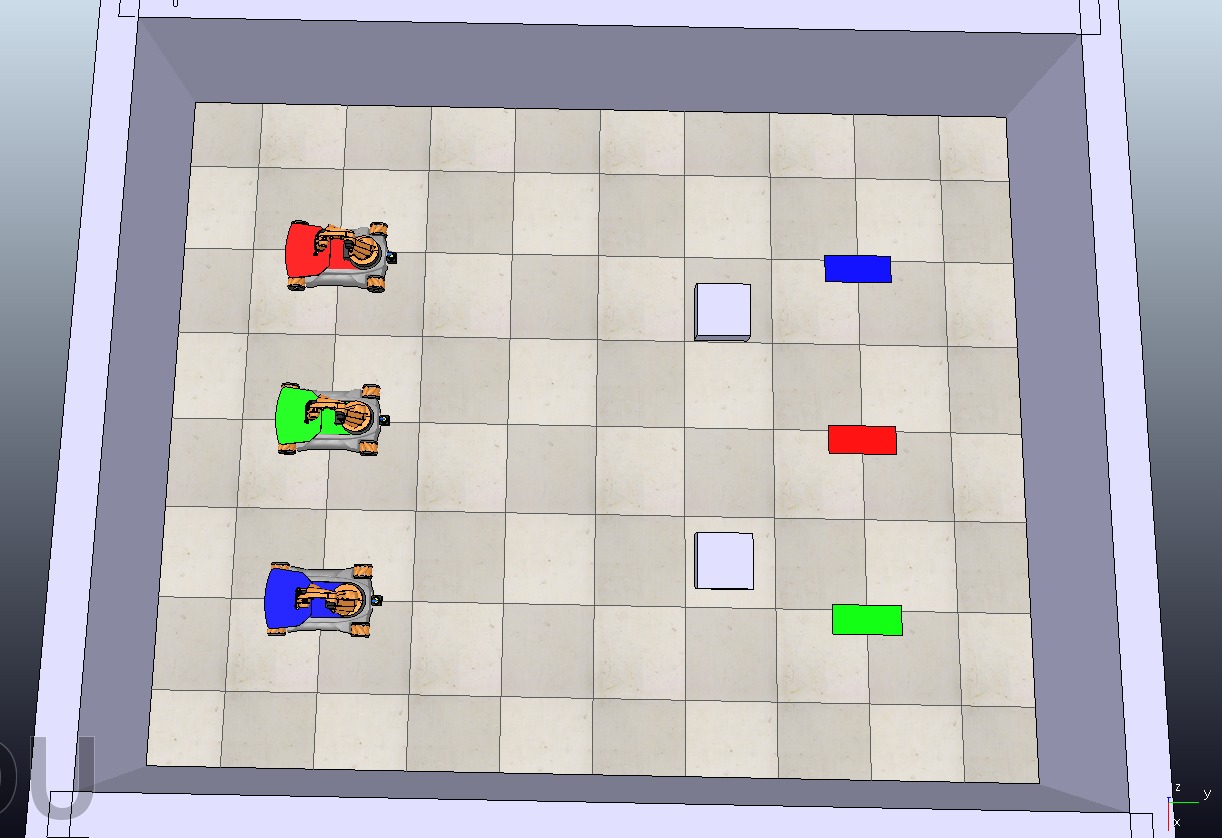}
    \caption{The simulation scene in CoppeliaSim for the three YouBots.}
    \label{fig:robot_scene_sim}
\end{figure}

In this Section, we describe experiments performed with Kuka YouBot models in a simulated environment.
These experiments are useful to validate the results highlighted in the previous section with an application of our framework to a more realistic use case.
The simulation scene is shown in Figure \ref{fig:robot_scene_sim}.
It includes three YouBots which can move in a rectangular region defined by four walls (treated as obstacles), with fixed cubes as obstacles on the way.
Each robot must reach a specific target position, defined by a platform with the same color as the robot. 
We assume that the geometry and the positions of the obstacles in the scene is known in advance.
The scene is built in the popular CoppeliaSim simulation environment from Coppelia Robotics \cite{coppeliaSim}, which allows to simulate the dynamics of the robots and to control them through ROS topics as in real applications.

\begin{figure}
    \subfloat[Null weights, static potential\label{subfig:sim_static_null}]{
        \includegraphics[width=0.45\linewidth]{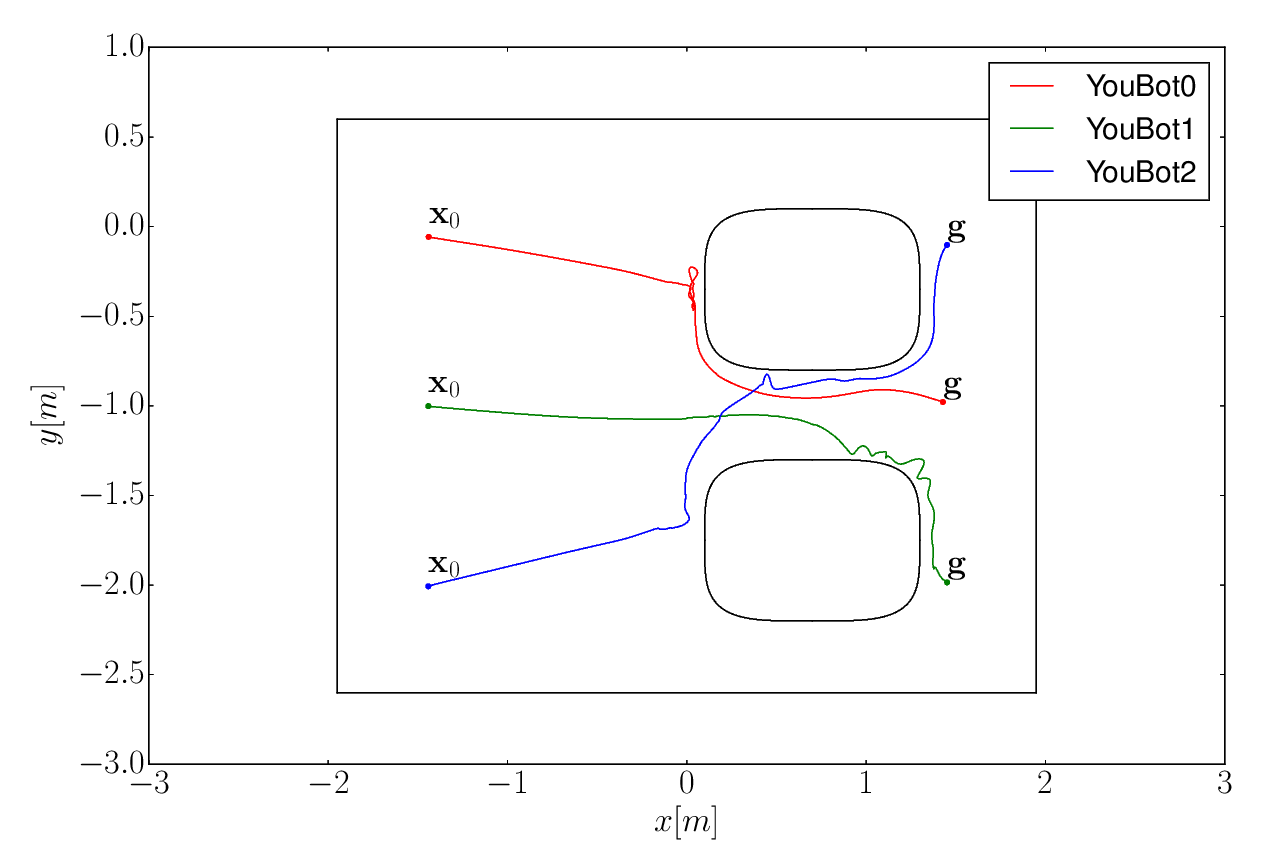}
    }
    \subfloat[Constant speed, static potential\label{subfig:sim_static_const}]{
        \includegraphics[width=0.45\linewidth]{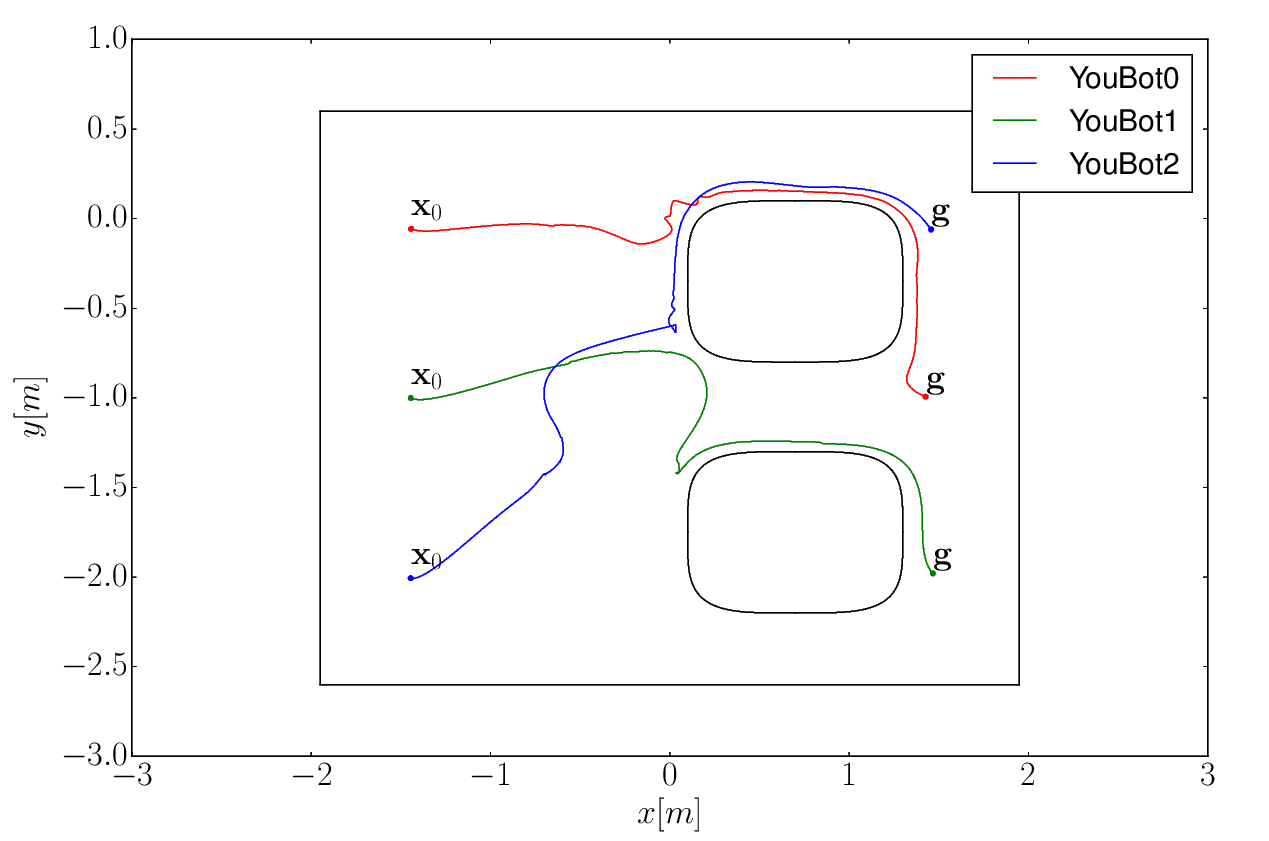}
    }
    \hfill
    \subfloat[Null weights, dynamic potential\label{subfig:sim_dyn_null}]{
        \includegraphics[width=0.45\linewidth]{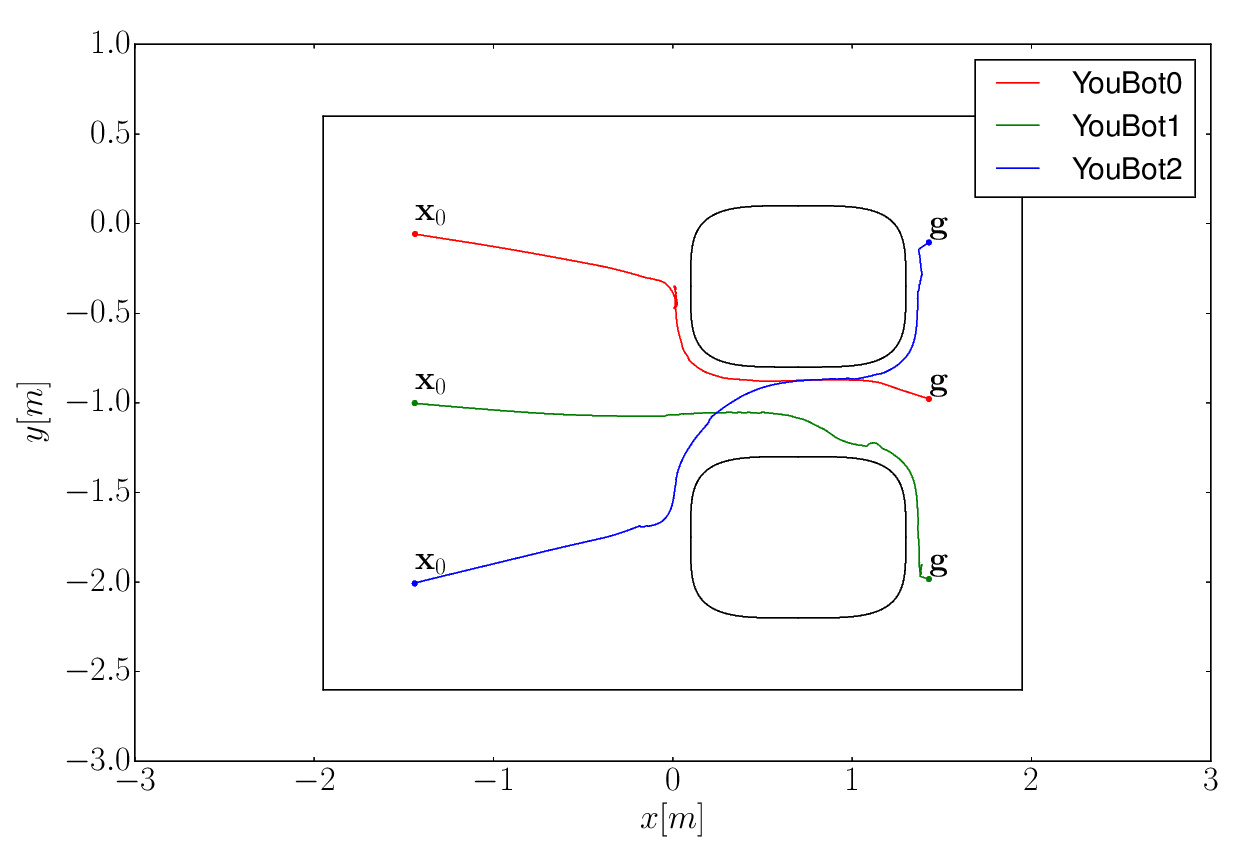}
    }
    \subfloat[Constant speed, dynamic potential\label{subfig:sim_dyn_const}]{
        \includegraphics[width=0.45\linewidth]{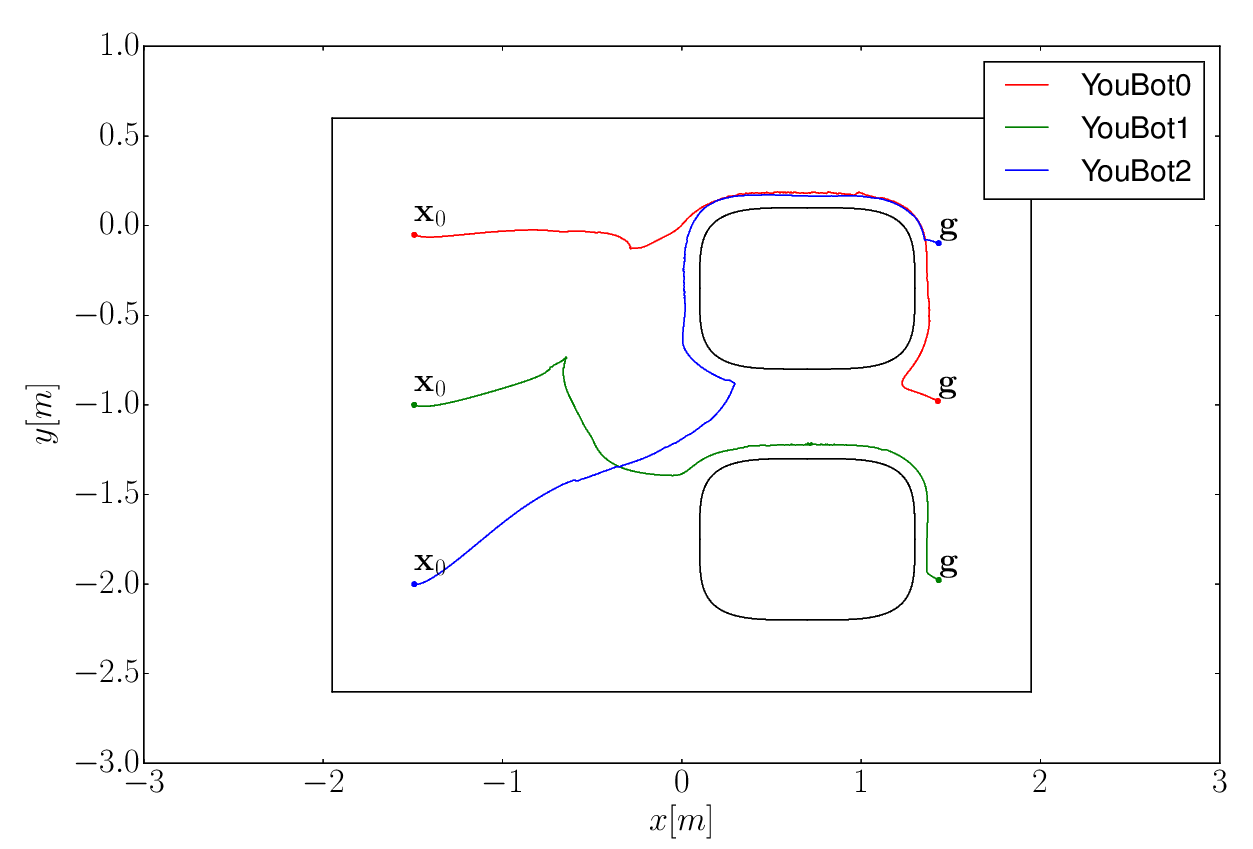}
    }
    \caption{DMPs with constant speed and with null weights of the three YouBots in simulation. Obstacles are represented in the scene with the superquadric isopotential approximation, enlarged of the dimensions of the YouBots. The walls are represented as a rectangle containing the robots and the other obstacles for simplicity. Trajectories are referred to the center points of the robots.}
    \label{fig:robot_plot_sim}
\end{figure}

\noindent
Each Youbot is controlled in position by a DMP with $\vx, \vv \in \RR^2$; we do not control the orientation of the robots along their normal axis, since we are interested in the obstacle avoidance problem for Cartesian DMPs.
In order to guarantee the synchronization between the robots, we construct a 6-dimensional DMP, concatenating the components $\vx, \vv, \dot{\vv} \in \RR^2$ of position, velocity and acceleration of each YouBot in a single array. In this way, the robots share the same canonical system.
The obstacle-free trajectory of each robot towards its target is a straight line. In this way, it is clear from the scene that the robots would collide during their motion.
Since the objects in the scene are known a priori, one could argue that the collision between the robots could be avoided computing the trajectories in advance, and coordinating the motion of the robots (e.g. tuning the speed of each of them appropriately). Multi-robot motion coordination has been extensively studied, and it is out of the scope of this paper. We refer the reader to \cite{yan2013survey} for a recent survey.
In our experiments, we have decided to simulate a more realistic multi-robot task, in which the robots do not know the trajectory of each other in advance. 
Hence, we model each YouBot as a dynamic potential using our formulation as in \eqref{eq:dynamic_potential_volume}, so that it influences the forcing terms of the other robots.
In this way, we show how our framework for obstacle avoidance is suitable for reactive motion planning.
At each time step, we build an ellipsoid around each YouBot, setting $m=n=1$ in \eqref{eq:isopot_gen_ell}. We control the center point of the YouBots, therefore the semi-axes of the ellipsoid are set as the full dimension of the robot (width $\times$ length) to avoid collisions. The parameters for the dynamic potential function are set as $\lambda=60, \eta=0.2, \beta=2$ after empirical evaluation.
When computing the forcing term for each robot, we compute the velocity term in \eqref{eq:dynamic_gradient} as the relative velocity between the robots.
We test two different straight line trajectories, one with null forcing term and the other with constant speed, to verify the independency of our framework with respect to the specific trajectory to be executed. 
The constant speed trajectory is first generated synthetically; then, the weights are learned as explained in Section \ref{sec:dmp_general}.
The DMP parameters are set as as $K= 3050, \alpha=4, D=\sqrt{K}$ for both sets of weights.
The trajectories are computed at $1 \ ms$ step of integration.
We model the walls and the fixed obstacles as generalized ellipsoids (enlarged of the dimension of the YouBots), setting $n=m=2$ in \eqref{eq:isopot_gen_ell} to better approximate the sharp edges.
We compare the performance of our previous static obstacle formulation \eqref{eq:static_potential_volume} with our novel one, modeling the fixed obstacles with both methods.
The results are shown in Figure \ref{fig:robot_plot_sim}.

Figures \ref{subfig:sim_static_null}-\ref{subfig:sim_static_const} are obtained setting $A=60, \eta=2$ in \eqref{eq:static_potential_volume}.
Figures \ref{subfig:sim_dyn_null}-\ref{subfig:sim_dyn_const} are obtained setting $\lambda=60, \beta=2, \eta=2$ in \eqref{eq:dynamic_potential_volume}.
Parameters are set after empirical evaluation.
We notice that the dynamic potential formulation results in smoother trajectories as the robots move close to the cubes in the scene.
This is due to the dependency of the forcing term on the velocity. 
In fact, the forcing term in \eqref{eq:dynamic_gradient} deviates the trajectory earlier depending on the module of the velocity when the robot moves in the direction of the obstacle, and not only on the position of the obstacle as in the static gradient formulation \eqref{eq:static_gradient}.
We also notice that the shape of the trajectory does not change significantly with different potential models for the obstacles.
Hence, we conclude that it is convenient to model even fixed obstacles with dynamic potential functions.

\subsection{Experiments on Real Setups}
\label{subsec:result_real}
We now show the results of the tests performed on different robots.
At first, we tested our obstacle avoidance framework on an industrial manipulator Panda from Franka Emika, studying a standard pick-and-place task with pegs and rings.
Then, we replicated the same task on a smaller setup with a surgical robot da Vinci from Intuitive Surgical, showing that our framework is able to scale with the dimension of the setup.
Finally, a scenario with a YouBot in a partially unstructured environment is tested, showing how our framework can be easily integrated with scene reconstruction techniques through vision sensors.

\subsubsection{Experiments with Panda robot} \label{subsec:panda}

\begin{figure}[htb]
    \centering
    \subfloat[Initial setup.\label{subfig:panda_setup}]{
        \includegraphics[height=0.40\linewidth]{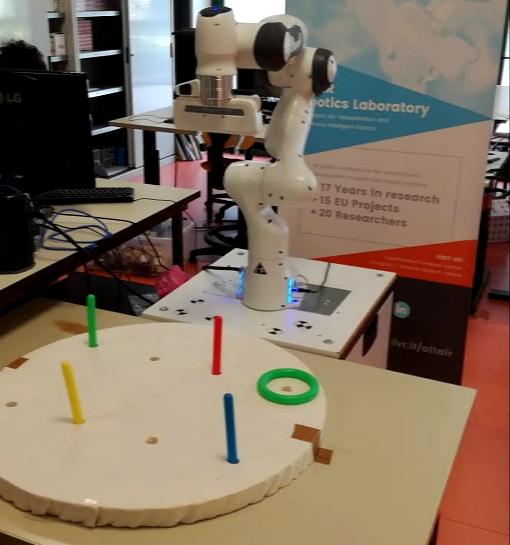}
    }
    \subfloat[Grasped ring.\label{subfig:panda_grasped}]{
        \includegraphics[height=0.40\linewidth]{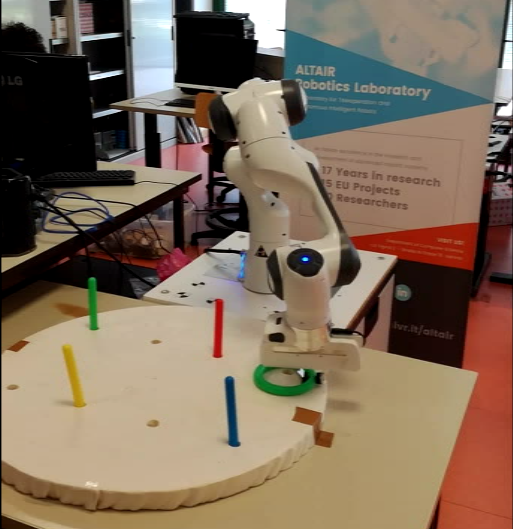}
    }
    \hfill
    \subfloat[Avoided obstacle.\label{subfig:panda_avoided}]{
        \includegraphics[height=0.40\linewidth]{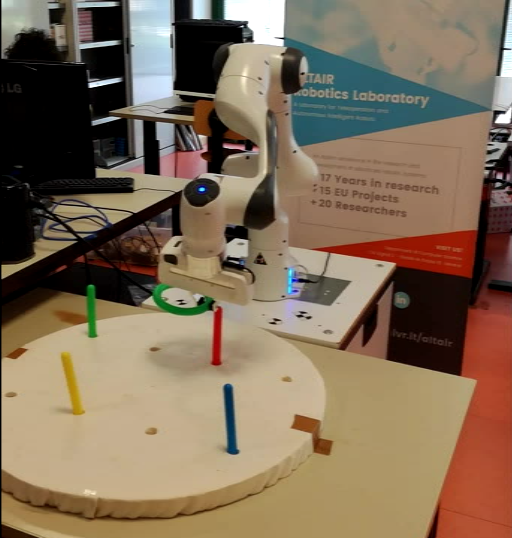}
    }
    \subfloat[Released ring.\label{subfig:panda_released}]{
        \includegraphics[height=0.40\linewidth]{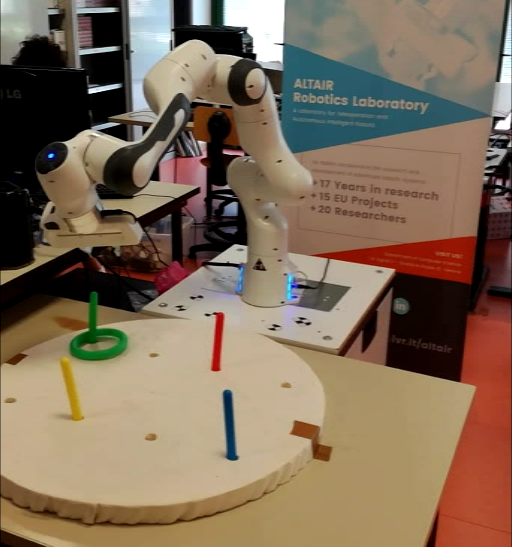}
    }
    \caption{The pick-and-place task with the Panda robot.}
    \label{fig:panda_frames}
\end{figure}

\begin{figure}[htb]
    \centering
    \subfloat[Move to ring (dynamic potential).\label{subfig:move_panda}]{
        \includegraphics[width=0.45\linewidth]{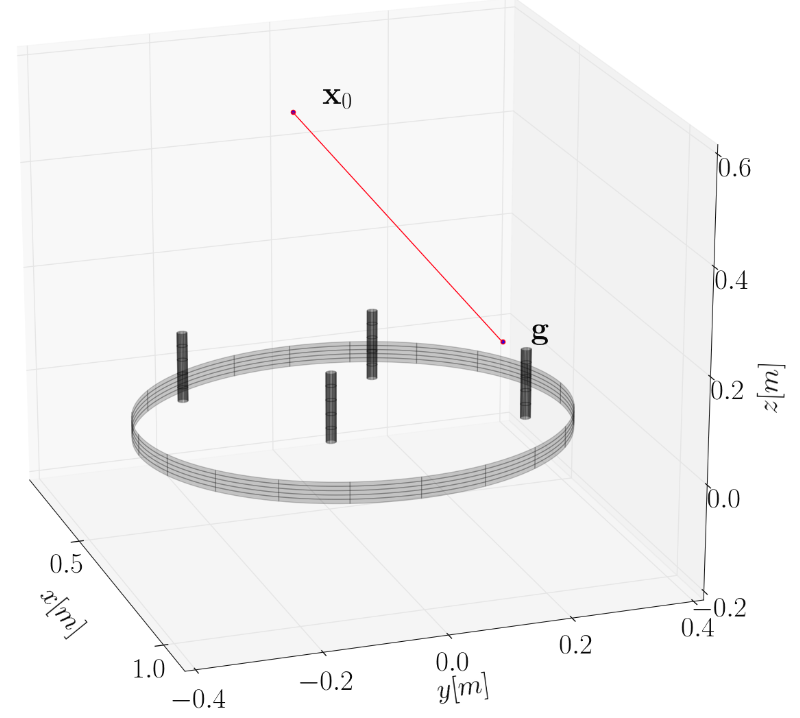}
    }
    \hfill
    \subfloat[Move to peg (static potential)\label{subfig:carry_st_panda}]{
        \includegraphics[width=0.45\linewidth]{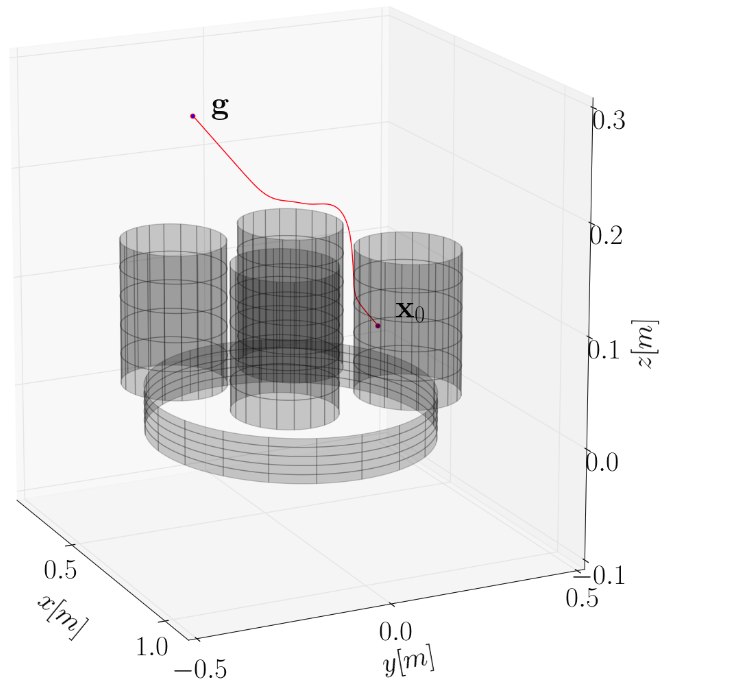}
    }
    \subfloat[Move to peg (dynamic potential)\label{subfig:carry_dyn_panda}]{
        \includegraphics[width=0.45\linewidth]{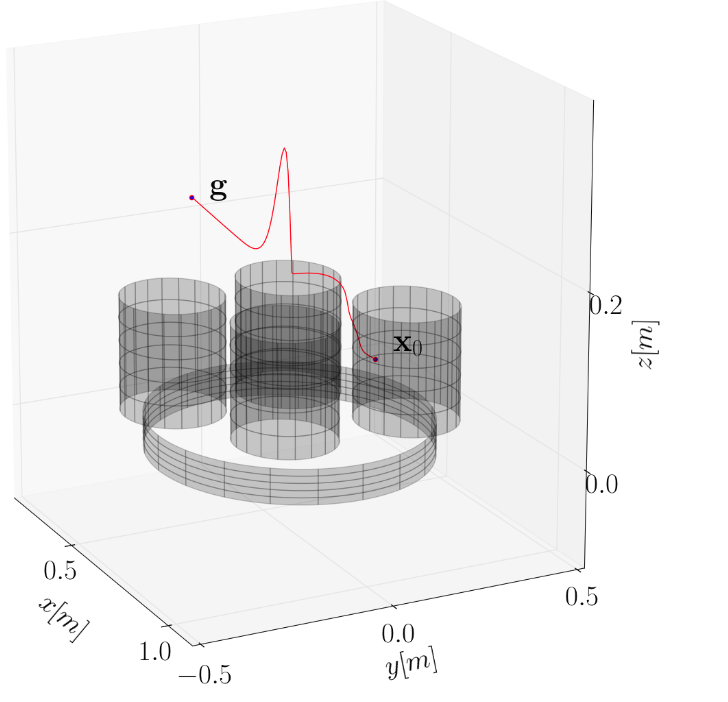}
    }
    \caption{Moving trajectories of the pick-and-place task with Panda robot. Axes coordinates are referred to the frame of the base link of the robot.}
    \label{fig:panda_exp}
\end{figure}

The setup for the Panda robot is shown in Figure \ref{subfig:panda_setup}.
The robot must pick the green ring and place it on the green peg.
On the way to the peg, the robot has to avoid the red peg, i.e. neither the end effector nor the grasped ring have to hit the peg.
The task can be described by a simple state machine with four actions / states: \emph{move to ring, grasp, move to peg} and \emph{release gripper}.
The moving actions are kinematically described with two DMPs in Cartesian space with null weights, i.e. straight line trajectories, with $K=3050, \alpha=4, D=\sqrt{2}K$.
The trajectories describe the motion of the center of the gripper of the robot.
Notice from Figure \ref{subfig:panda_setup} that the encumbrance of the end effector is significant, and controlling only the center of the gripper does not guarantee safe collision avoidance.
As explained in our previous work \cite{GMCDSF19}, there are two solutions to this issue.
One is to enlarge the radial dimension of the pegs according to the size of the end effector. The second solution is to exploit the kinematic redundancy of the 7-DOF manipulator and compute an obstacle-free joint configuration for each point in the DMP.
We have chosen the latter approach to limit the size of the obstacles and, hence, maximize the reachable workspace for the robot.
We control the robot through its standard MoveIt / ROS interface, setting TRAC-IK \cite{beeson2015trac} as inverse kinematics solver.
TRAC-IK is a state-of-the-art library for this purpose, and it has been chosen because it allows to define optimal metrics to compute the joint configuration from a given pose.
We set the solver to compute an inverse kinematics solution which maximizes the manipulability of the robot, defined as $\sqrt{det(JJ^T)}$ \cite{GMCDSF19}.
Though we do not control the orientation of the end effector with our DMP formulation, we constrain the orientation to be within $5\degree$ (along each axis) from the initial orientation for each Cartesian waypoint (shown in Figure \ref{subfig:panda_setup}).
Then, we gradually relax this tolerance if no inverse kinematcs solution is found.
We also constrain two consecutive joint configurations to differ no more than $45\degree$ on each joint, so that we are able to avoid abrupt movements during the execution.
The scene (location of the peg base, the pegs and the ring) is assumed to be known in advance.
Hence, obstacles (the base and the pegs) are represented as superquadric potential function shaped as cylinders (assuming the $z-$axis as the normal to the base, exponents in \eqref{eq:isopot_gen_ell} are set as $m=n=1, p=2$).
Figure \ref{fig:panda_frames} shows the main steps in the task execution. 
In Figure \ref{fig:panda_exp} we show the trajectories for the two actions. 
Notice that the radial dimension of the pegs is enlarged when moving to the peg.
In fact, we need to avoid that the grasped ring hits the obstacles. Hence, the radius of the base of the pegs is augmented of the diameter of the ring.
Obstacles are modelled with our dynamic potential formulation when moving to the ring, setting $\lambda=10, \eta=2, \beta=2$ in \eqref{eq:dynamic_potential_volume}.
When moving to the peg, we compare our approach with the static potential formulation in \eqref{eq:static_potential_volume}, setting $A=10, \eta=2$ (Figures \ref{subfig:carry_st_panda}-\ref{subfig:carry_dyn_panda}).
We notice that the dynamic potential determines a glitch of the trajectory along the $z-$axis in correspondence of the peg, while this phenomenon does not occur with the static potential.
This is due to the dependency of \eqref{eq:dynamic_gradient} on the velocity of the robot. Since there is a difference between the $z-$coordinates of the starting and goal position, the natural attraction of the dynamical system in \eqref{eqs:dmps_eq} towards its goal sums to the repulsive potential of the obstacle induced by the relative position and velocity of the robot moving upwards.
Our Online resource shows the full execution of the task with the static potential formulation.

\subsubsection{Experiments with the da Vinci surgical robot}\label{subsec:dvrk}

\begin{figure}[htb]
    \centering
    \includegraphics[scale=0.25]{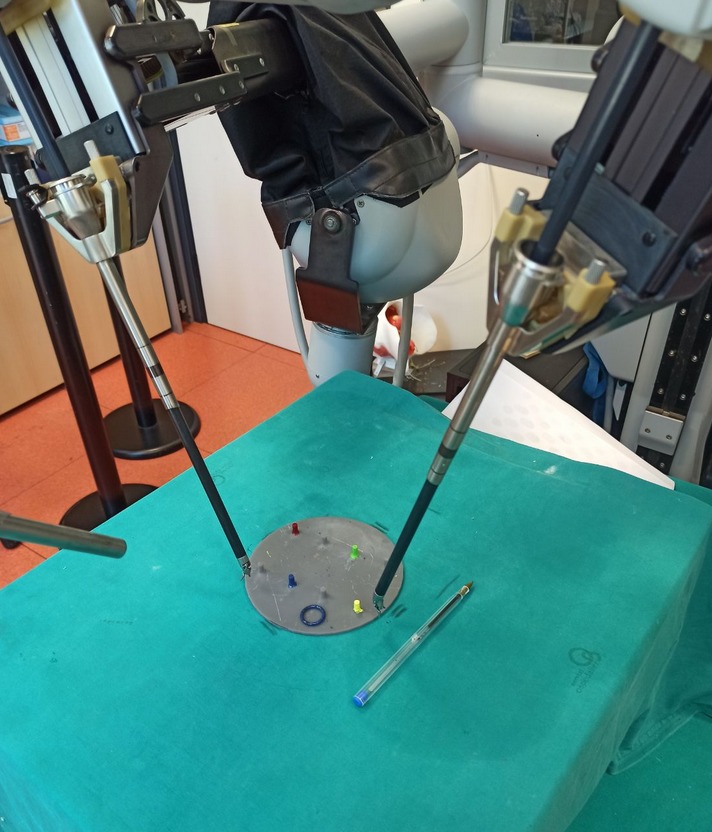}
    \caption{The setup for the peg transfer task with the da Vinci surgical robot: PSM1 on the right and PSM2 on the left.}
    \label{fig:dvrk_setup}
\end{figure} 

\begin{figure}[htb]
    \centering
    \subfloat[Move to ring with PSM1.\label{subfig:move_dvrk}]{
        \includegraphics[width=0.45\linewidth]{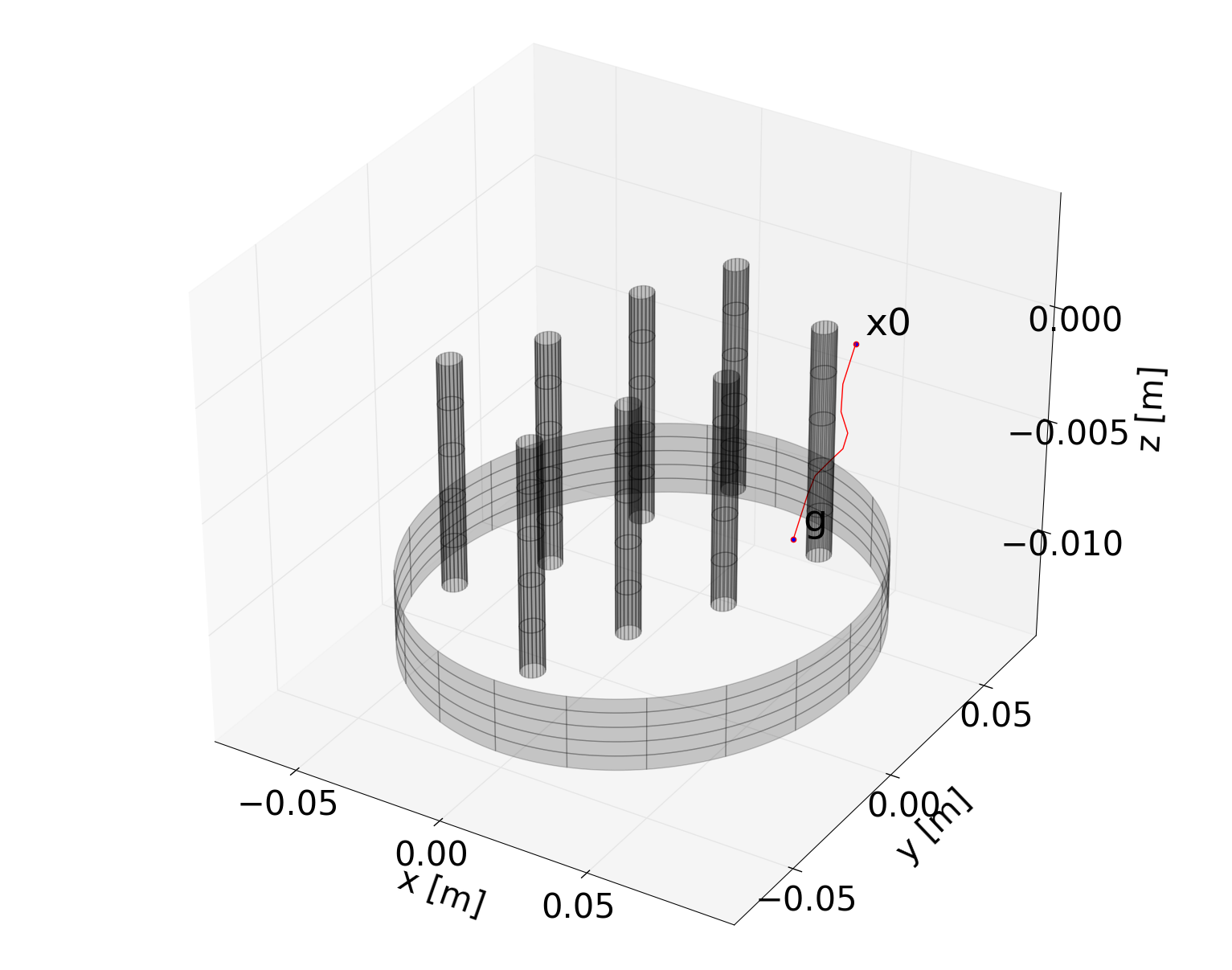}
    }
    \hfill
    \subfloat[Transfer between PSMs (PSM1 in red, PSM2 in green).\label{subfig:transfer_dvrk}]{
        \includegraphics[width=0.45\linewidth]{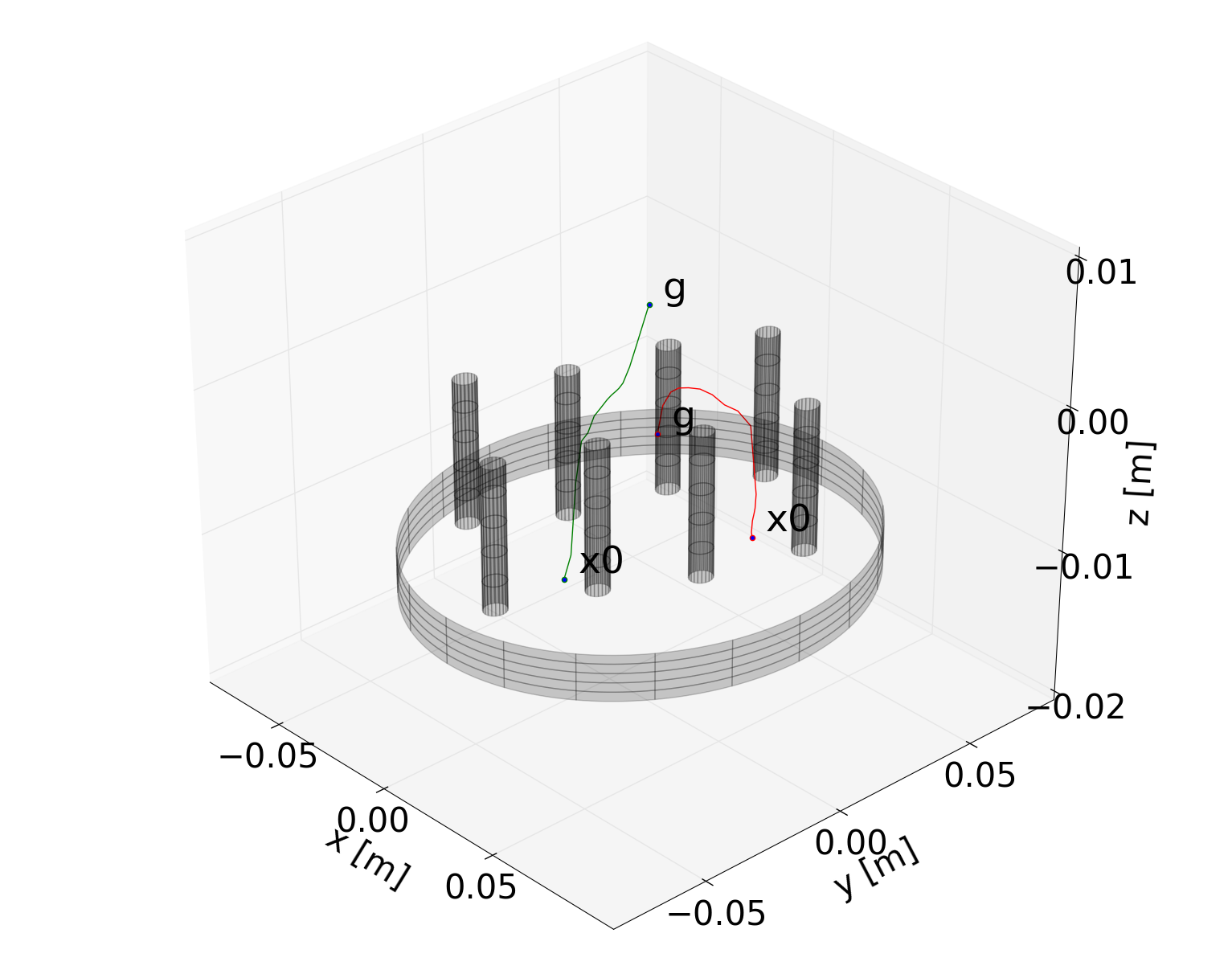}
    }
    \subfloat[Move to peg with PSM2.\label{subfig:carry_dvrk}]{
        \includegraphics[width=0.45\linewidth]{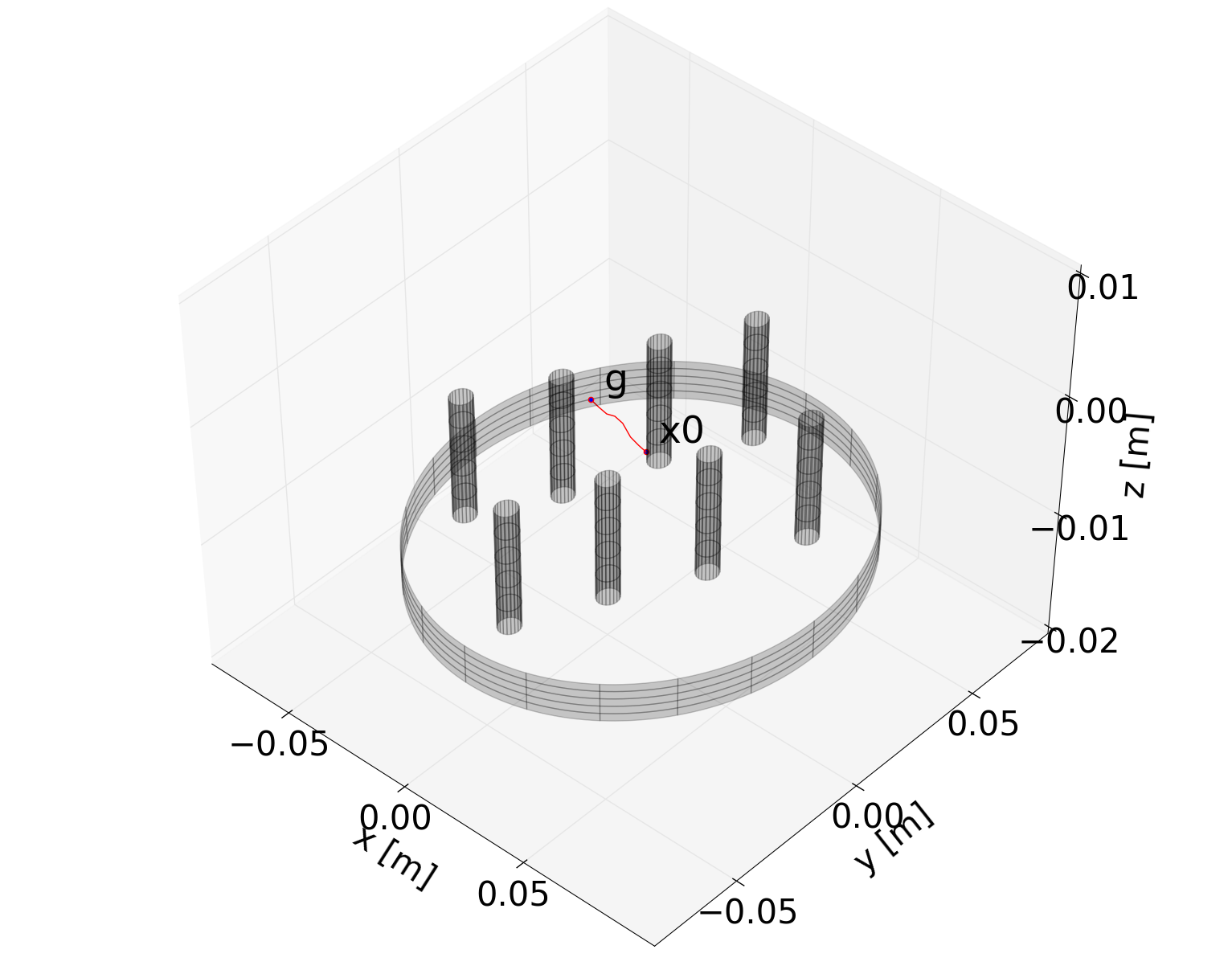}
    }
    \caption{Trajectories executed by the da Vinci arms. Trajectories are referred to the center of the grippers, and they are expressed in a fictitious coordinate frame common to the PSMs, obtained using our calibration procedure presented in \cite{calibration}.}
    \label{fig:dvrk_traj}
\end{figure}

\begin{figure}[htb]
    \centering
    \subfloat[Initial condition.\label{subfig:initial_dvrk_frame}]{
        \includegraphics[height=0.2\linewidth]{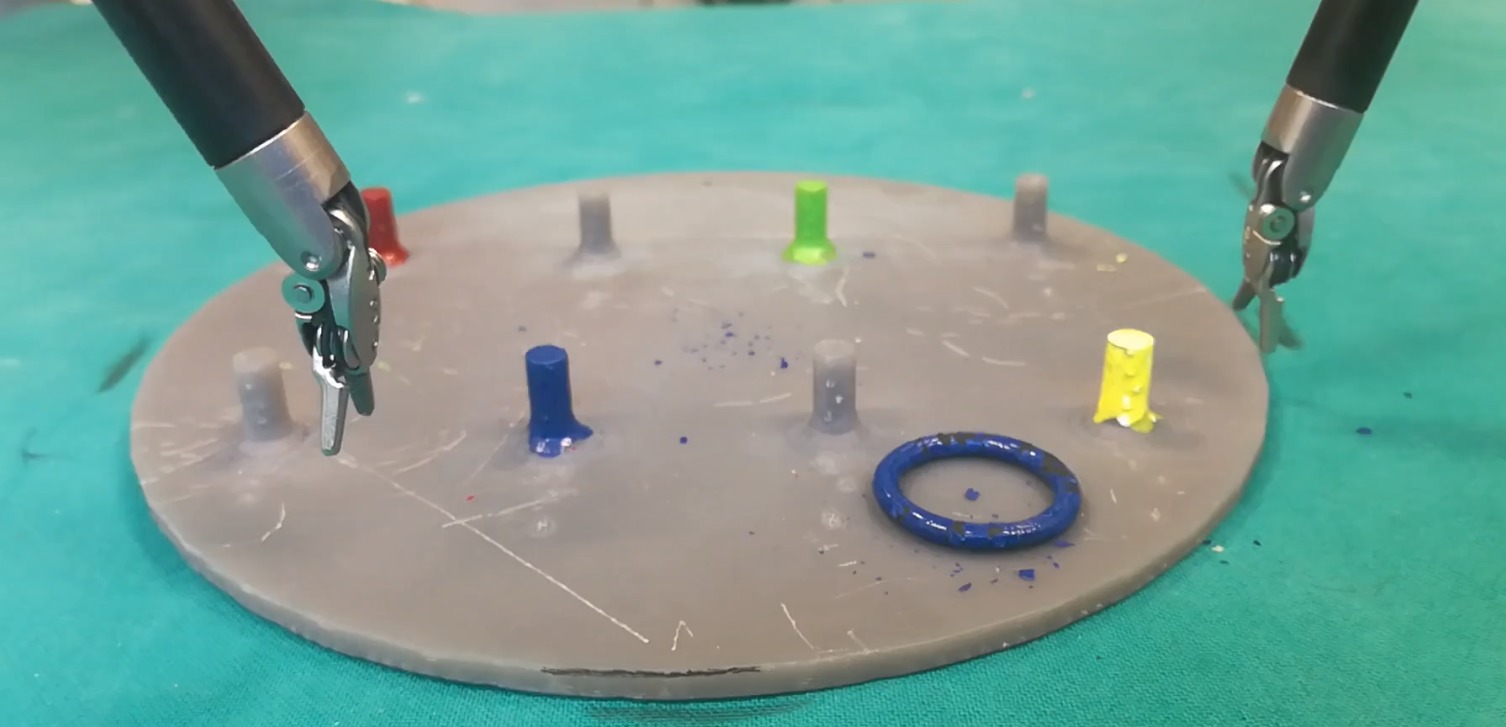}
    }
    \subfloat[Ring grasped by PSM1.\label{subfig:grasped_dvrk_frame}]{
        \includegraphics[height=0.2\linewidth]{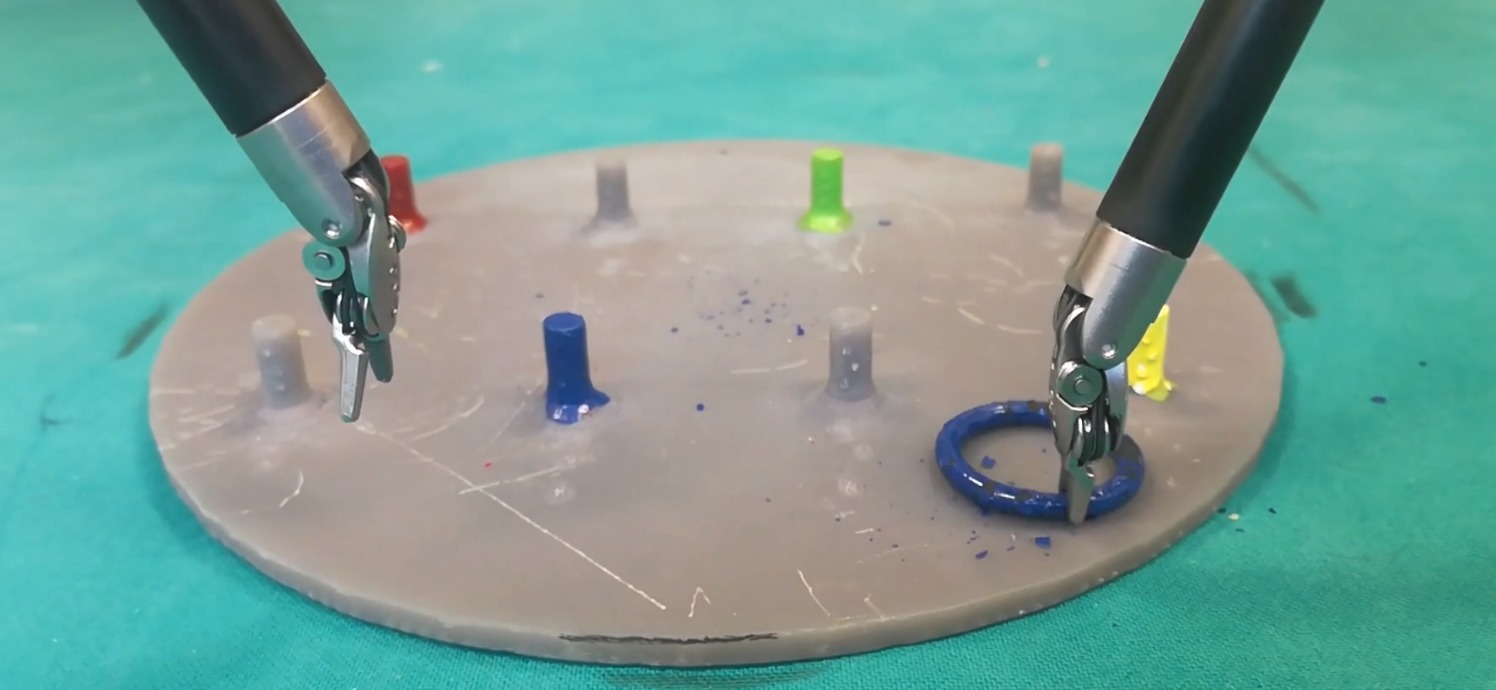}
    }
    \hfill
    \subfloat[Ring transferred.\label{subfig:transferred_dvrk_frame}]{
        \includegraphics[height=0.205\linewidth]{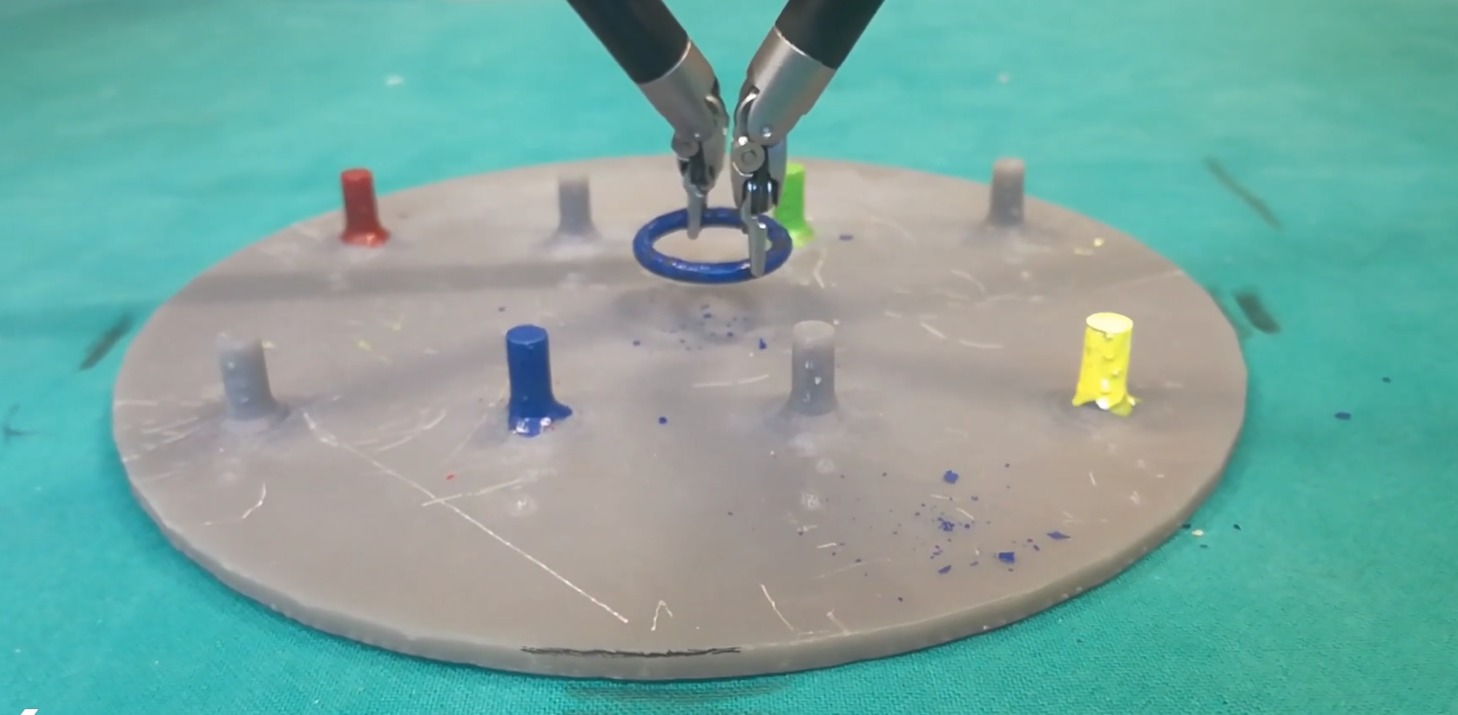}
    }
    \subfloat[Ring placed on the peg by PSM2.\label{subfig:carried_dvrk_frame}]{
        \includegraphics[height=0.205\linewidth]{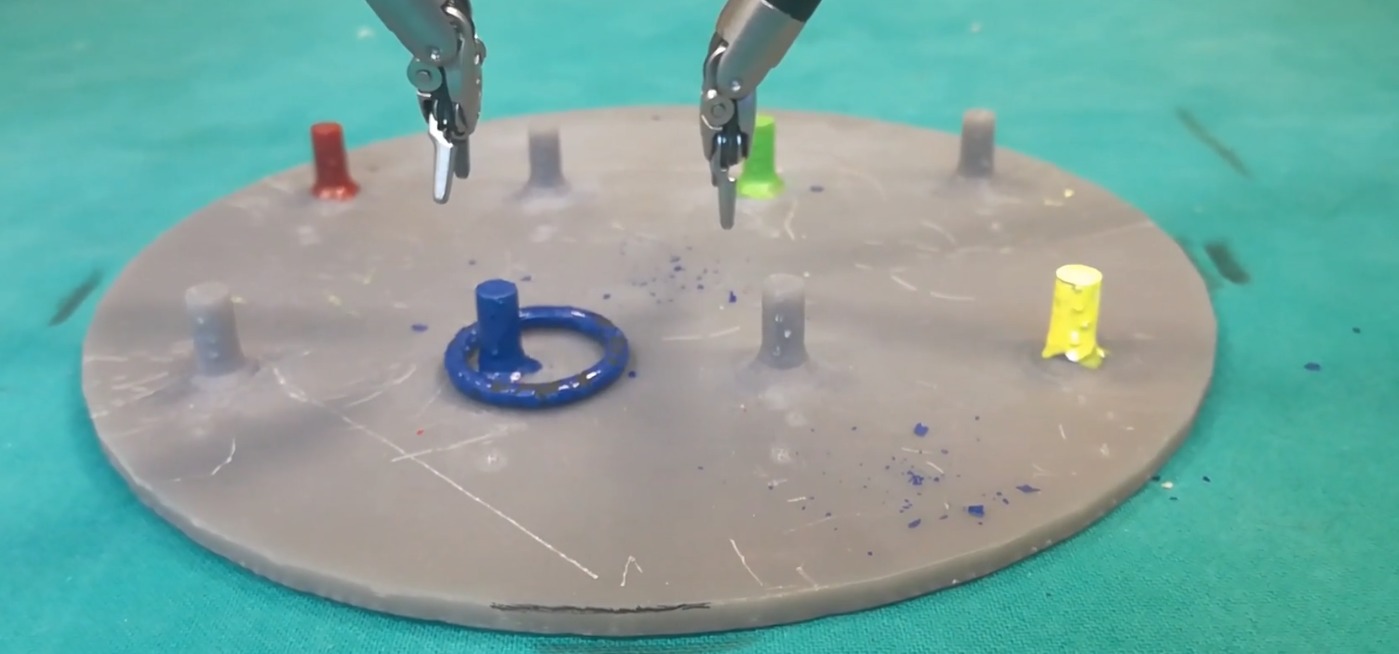}
    }
    \caption{Main steps of the peg transfer task with the da Vinci surgical robot.}
    \label{fig:dvrk_steps}
\end{figure}

We replicate the peg transfer task using the da Vinci surgical robot from Intuitive Surgical controlled through the da Vinci Research kit \footnote{\url{https://github.com/jhu-dvrk/dvrk-ros/tree/master/dvrk_python}} and ROS, with the setup shown in Figure \ref{fig:dvrk_setup}.
The robot has two arms, named PSM1 and PSM2. Hence, we modify the state machine for the task.
The PSM1 must move to the blue ring, grasp it, move the ring to the center of the base and exchange it with the PSM2.
Then, the PSM2 carries the ring to the blue peg and the task ends.
The scene description (locations of pegs, the ring and the base) is assumed as a prior.
We have designed the initial location of the arms and the ring in such a way that the pegs act as obstacles for the robot.
In order to make a comparison with the task with the Panda robot, the \emph{transfer} action can be seen as a combination of a \emph{move to ring} action for the PSM2 and a \emph{move to peg} action for the PSM1, where the goal is actually the center of the base instead of a real peg.
Hence, the actions executed by the surgical robot can be interpreted as a replication of the actions of the industrial manipulator, just scaled on a smaller size of the setup.
For this reason, we represent the obstacles with the same superquadric potentials (i.e. the same parameters) as in the Panda task. 
The trajectories of the robot are again described by Cartesian DMPs with null weights, and we build a single 6-dimensional DMP in order to share the same canonical system for the arms.
We first test our novel dynamic potential formulation to model the obstacles.
However, the DMPs does not converge to the goal in the transfer and when moving to the peg. 
On the contrary, our static potential formulation converges smoothly, as shown in Figure \ref{fig:dvrk_traj}.
Figure \ref{fig:dvrk_steps} shows the main steps of the task execution. The reader is referred to our Online resource for the full execution.
Notice that we do not need to compute inverse kinematics solutions from the DMP waypoints as with the industrial manipulator.
In fact, the arms of the surgical robot have 6 degrees of freedom, and we force the orientation of the grippers to stay constant during the task.
Moreover, Figure \ref{fig:dvrk_setup} shows that the encumbrance of the grippers is minor, hence they safely avoid obstacles.

\subsubsection{Experiments with real YouBot} \label{subsec:real_youbot}

\begin{figure}[htb]
    \centering
    \includegraphics[scale=0.2]{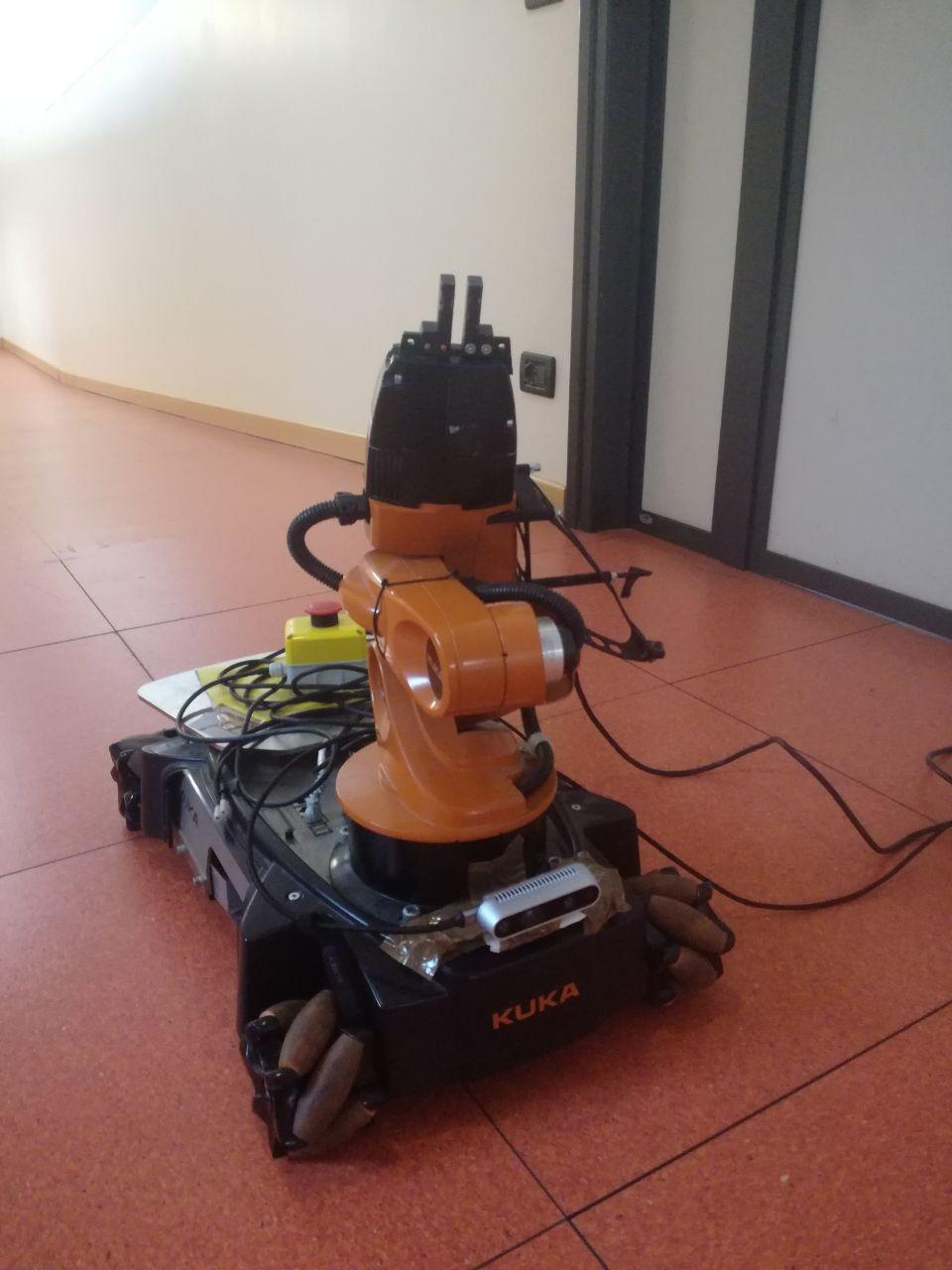}
    \caption{The YouBot with the Realsense D435 camera on its front.}
    \label{fig:youbot}
\end{figure} 

\begin{figure}[htb]
    \centering
    \includegraphics[scale=0.8]{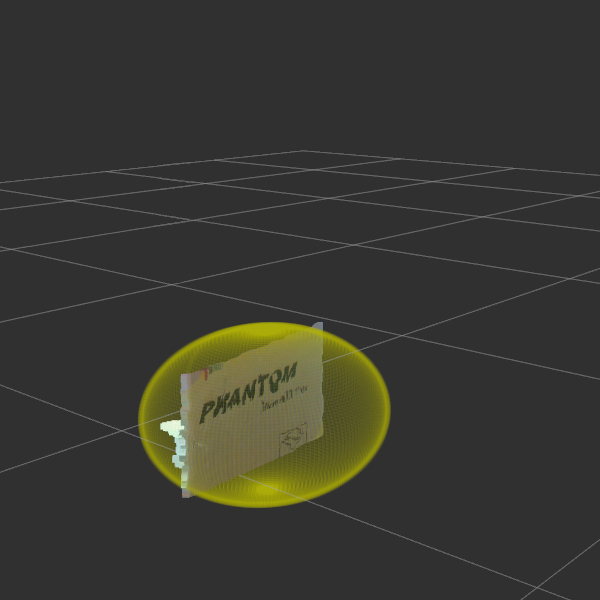}
    \caption{Point cloud filtered with the ellipsoid created around the object}
    \label{fig:youbot_pcd}
\end{figure} 

\begin{figure}[htb]
    \centering
    \subfloat[Added obstacle, static $A=10$.\label{subfig:yb_static_1}]{
        \includegraphics[width=0.5\linewidth]{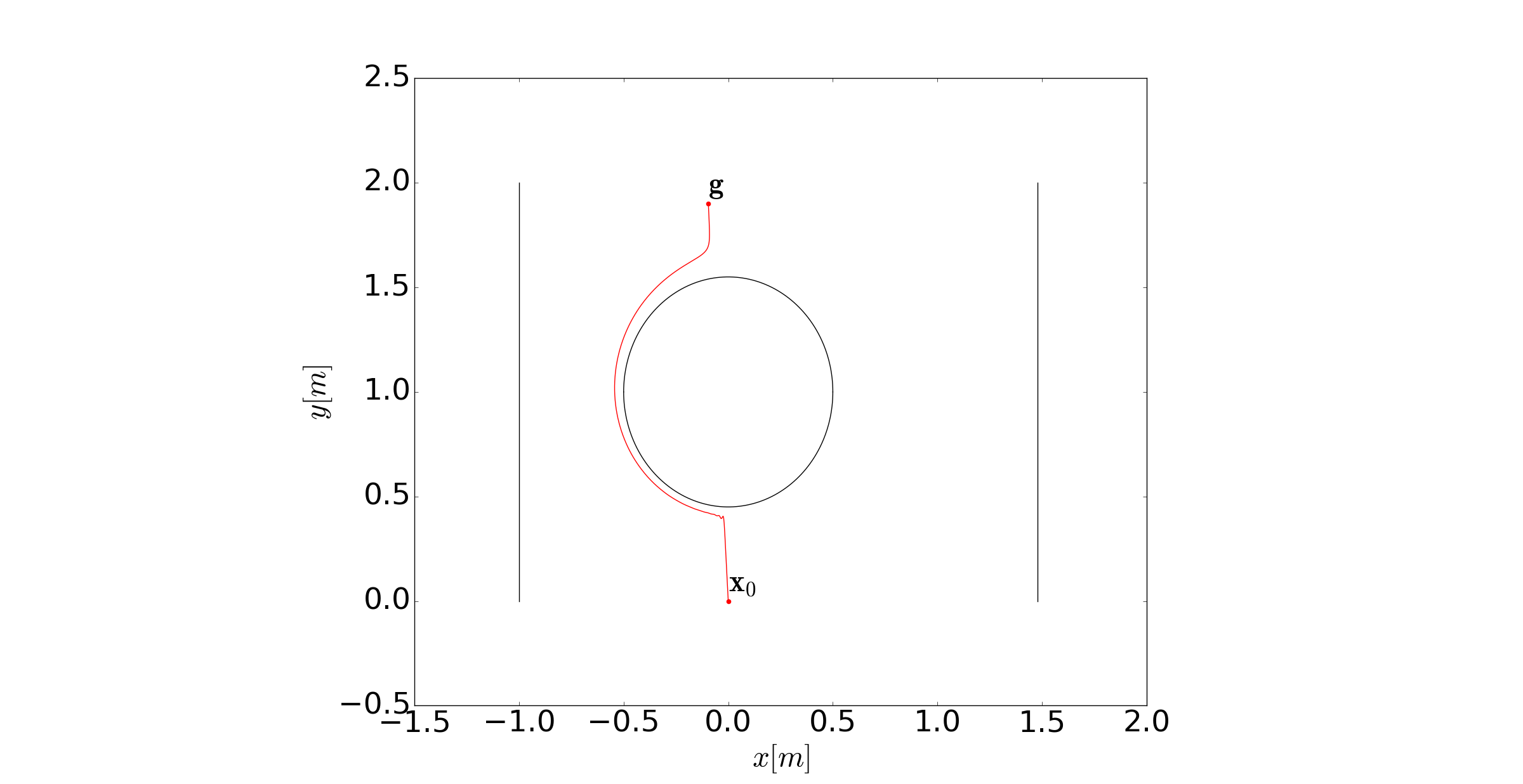}
    }
    \subfloat[Added obstacle, dynamic $\lambda=1$.\label{subfig:yb_dynamic_1}]{
        \includegraphics[width=0.5\linewidth]{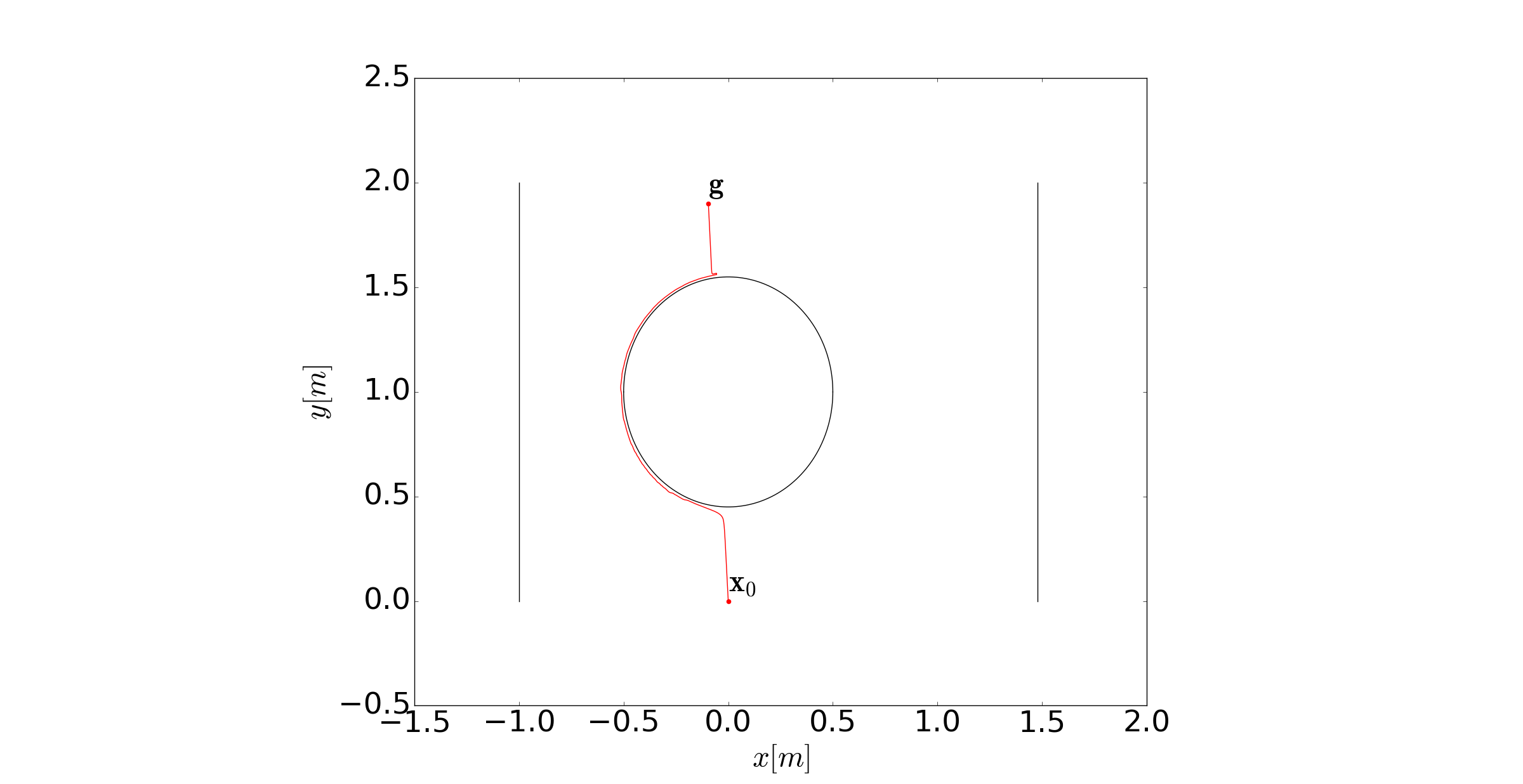}
    }
    \hfill
    \subfloat[Temporary obstacle, static $A=10$.\label{subfig:yb_static_2}]{
        \includegraphics[width=0.5\linewidth]{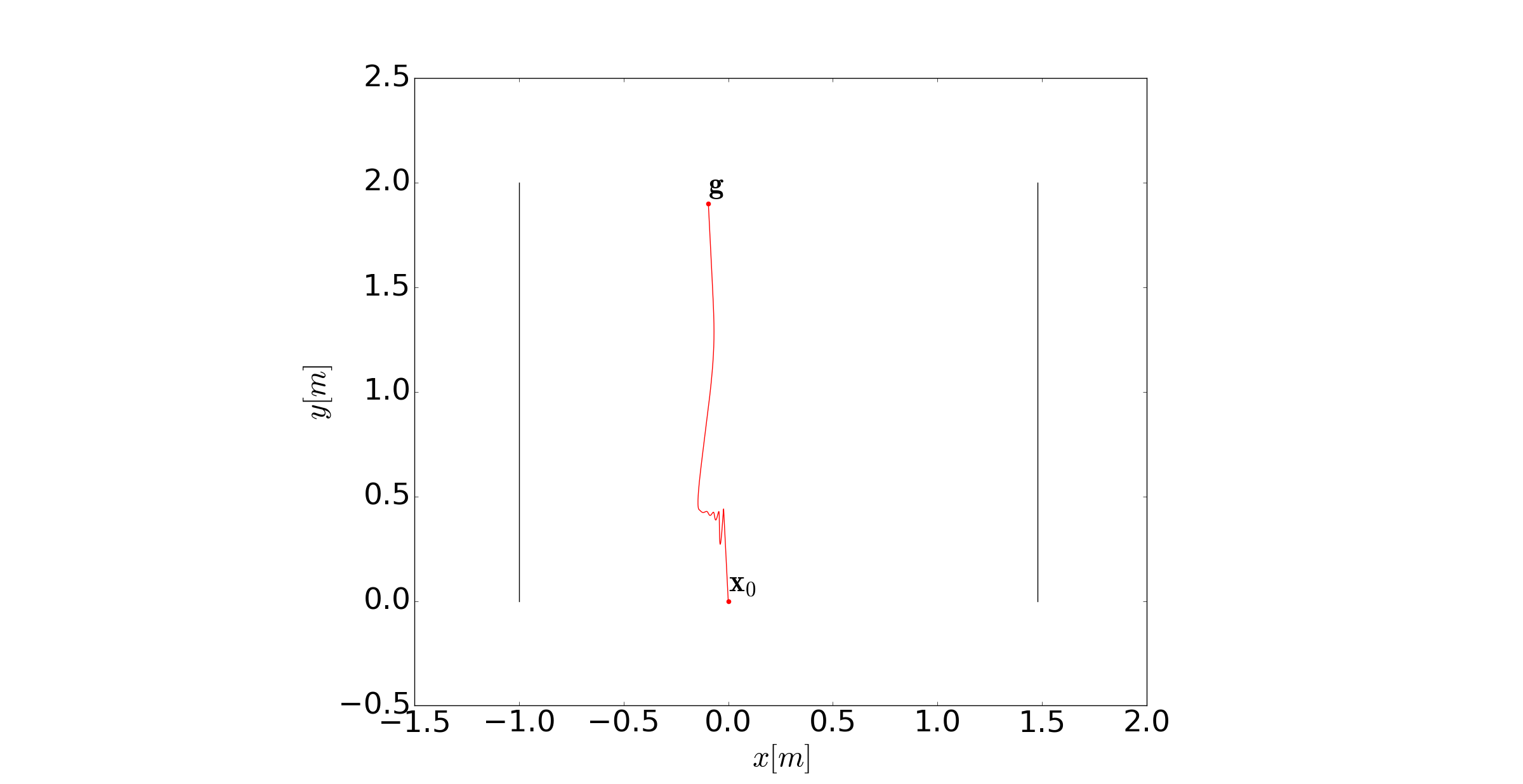}
    }
    \subfloat[Temporary obstacle, dynamic $\lambda=1$.\label{subfig:yb_dynamic_2}]{
        \includegraphics[width=0.5\linewidth]{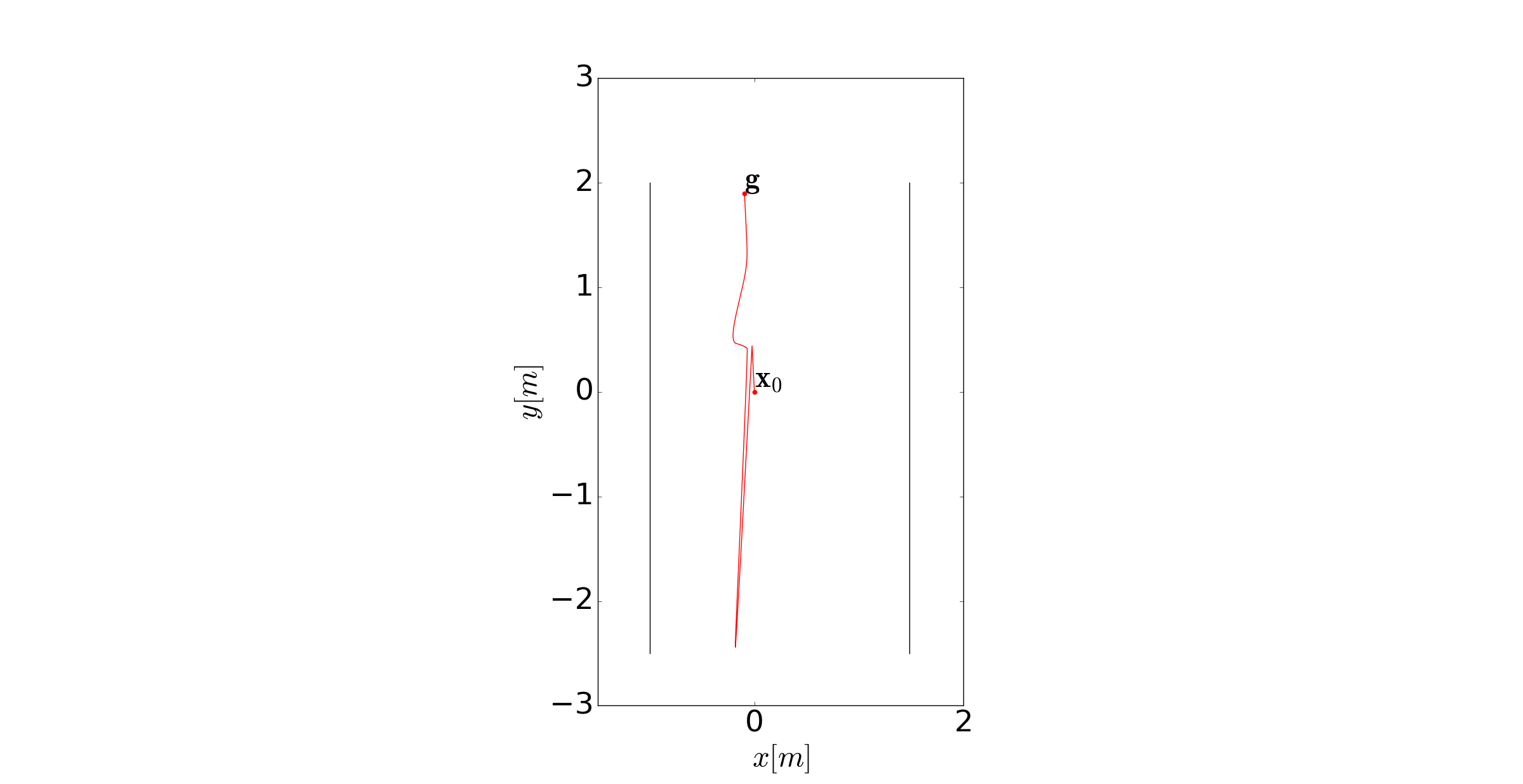}
    }
    \caption{Trajectories for the tests with the real YouBot. The shape of the ellipsoid of the obstacle is shown in the plot when it is added permanently. Walls are represented as straight lines. Units are referred to the initial position of the YouBot.}
    \label{fig:yb_plots}
\end{figure}

\begin{figure}[htb]
    \centering
    \subfloat[Start with no obstacle.\label{subfig:initial_youbot_frame1}]{
        \includegraphics[height=0.5\linewidth]{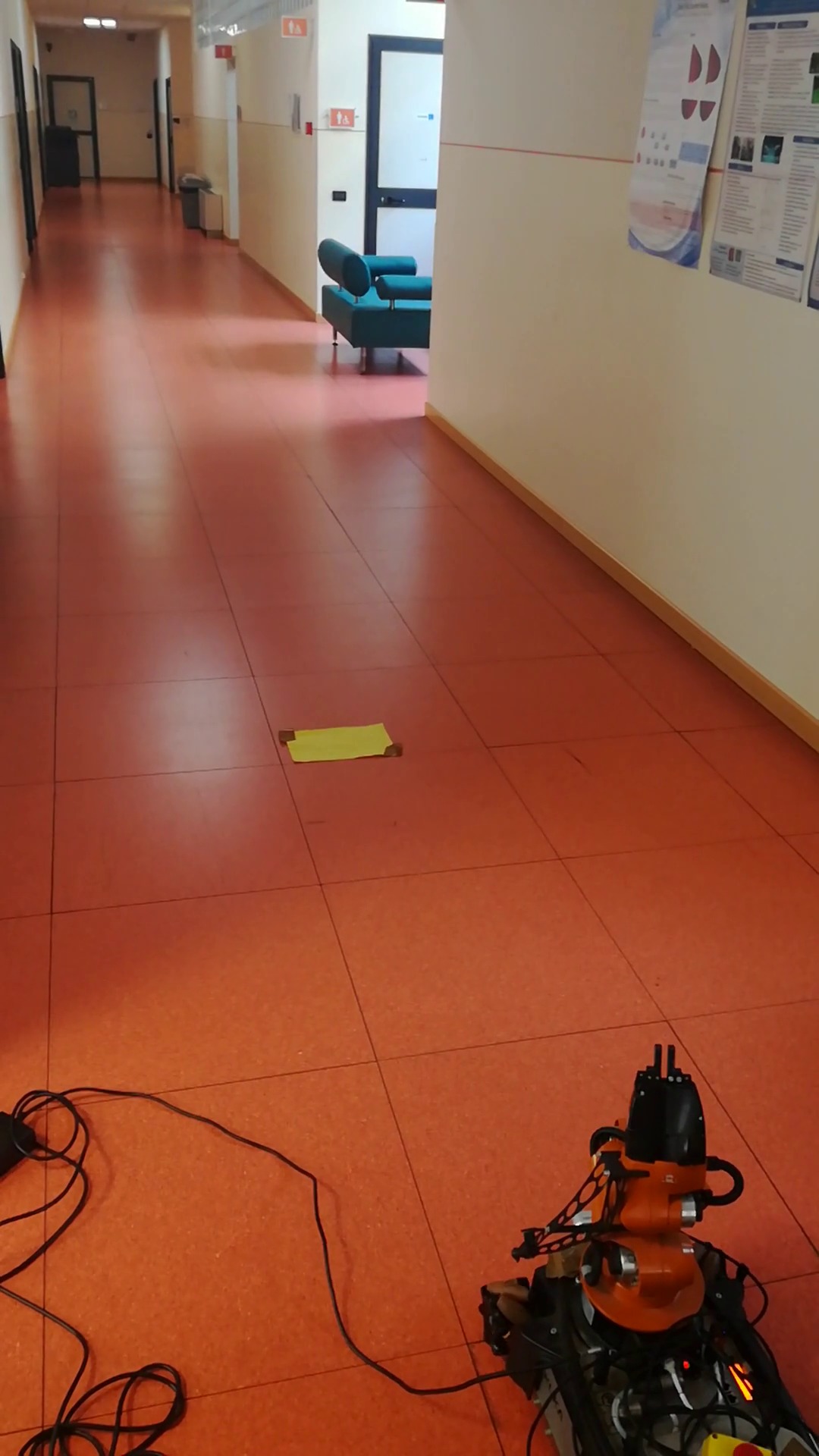}
    }
    \subfloat[Avoiding obstacle.\label{subfig:avoiding_youbot_frame1}]{
        \includegraphics[height=0.5\linewidth]{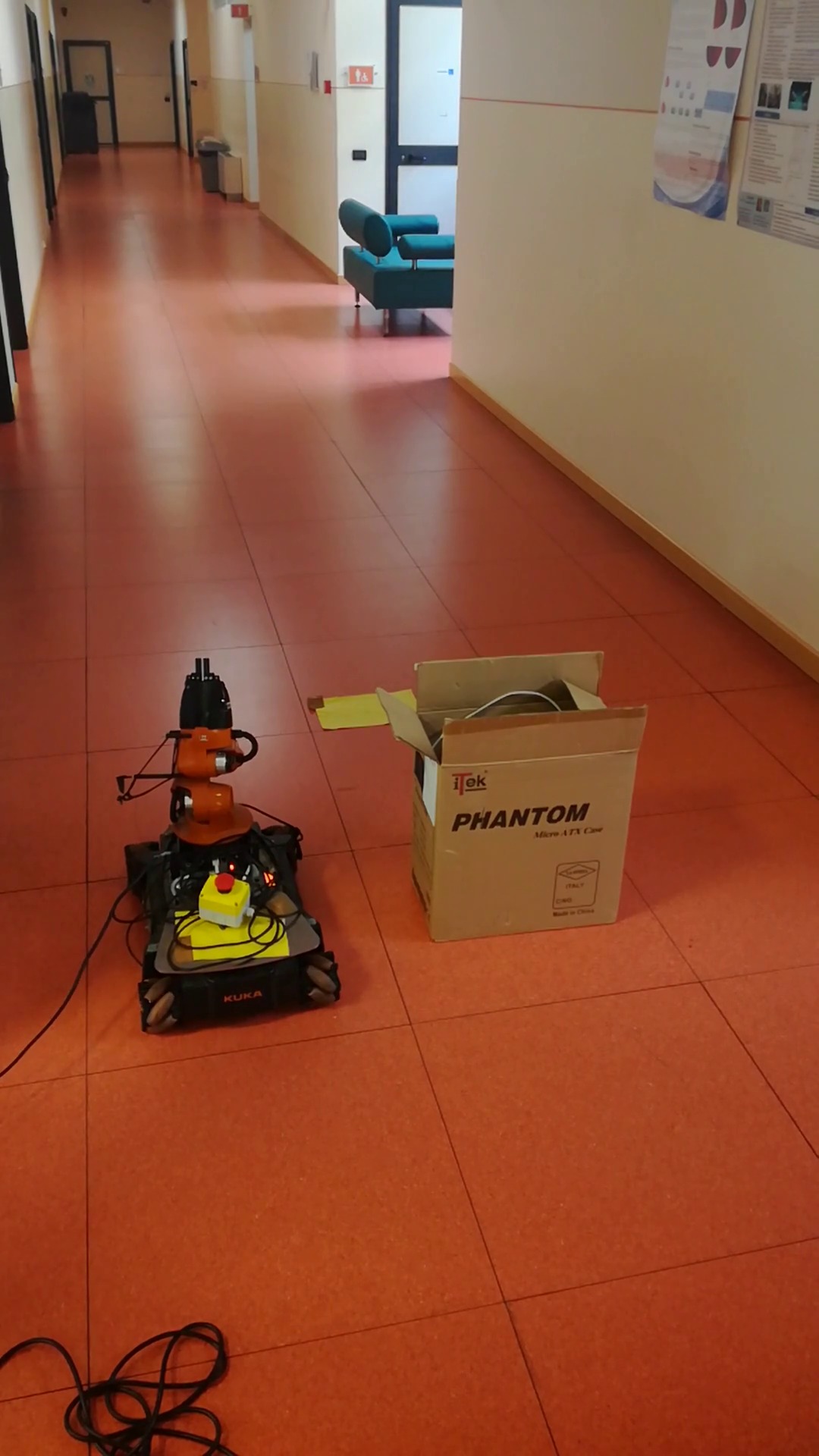}
    }
    \subfloat[Goal reached.\label{subfig:goal_youbot_frame1}]{
        \includegraphics[height=0.5\linewidth]{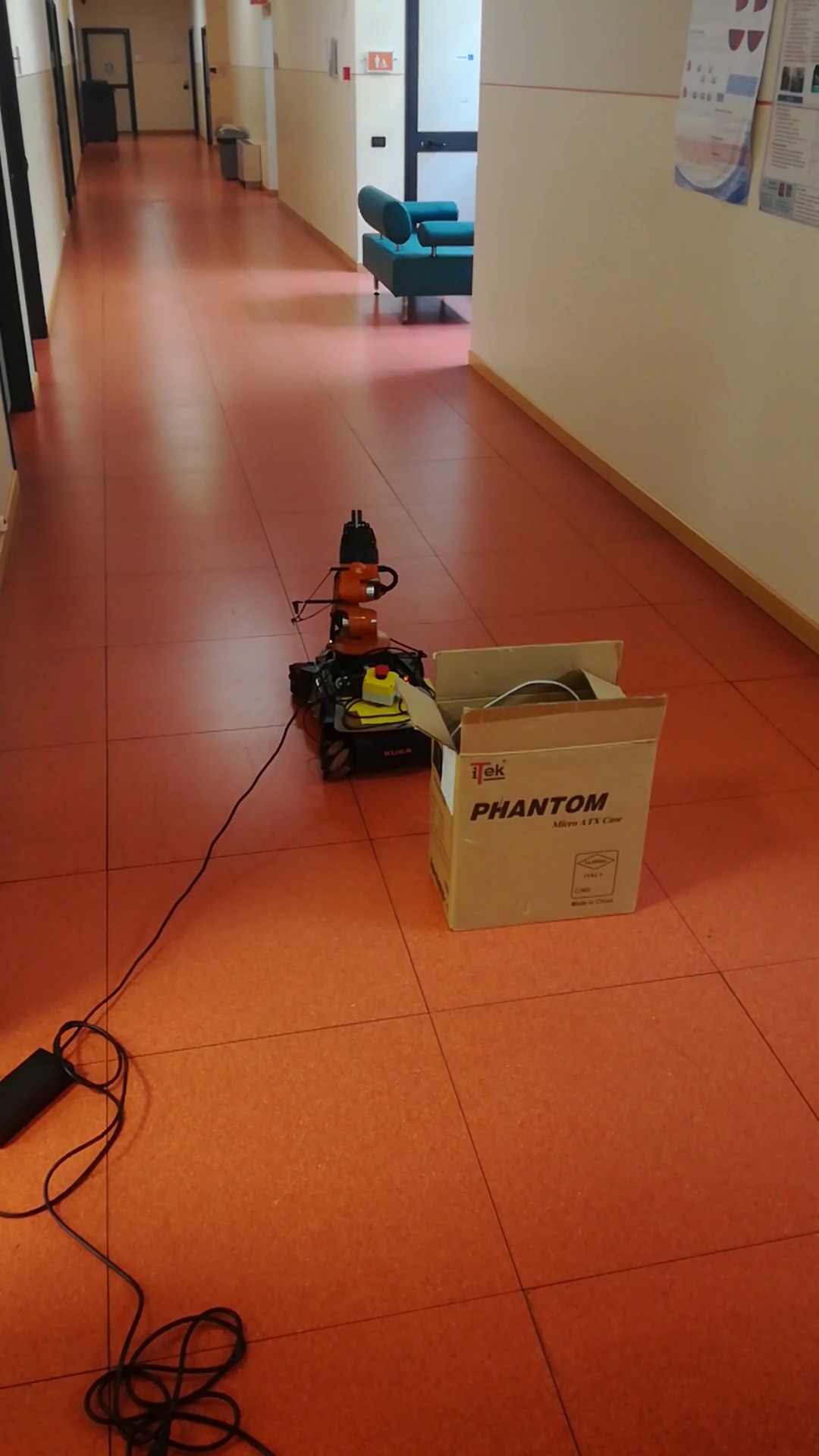}
    }
    \caption{Main steps of the YouBot task with the obstacle added to the scene during the execution.}
    \label{fig:youbot_exec1}
\end{figure}

\begin{figure}[htb]
    \centering
    \subfloat[Start with no obstacle.\label{subfig:initial_youbot_frame2}]{
        \includegraphics[height=0.5\linewidth]{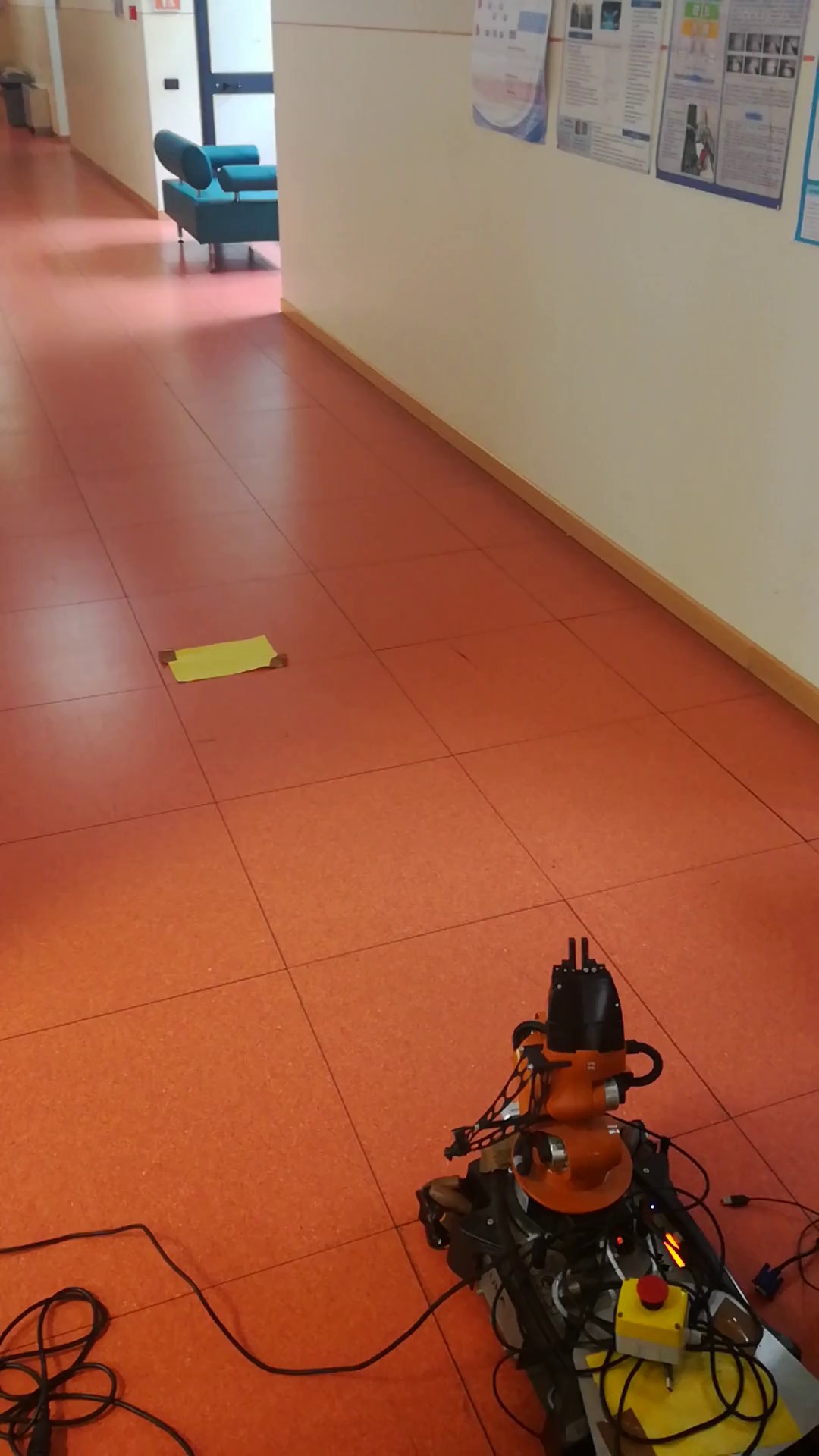}
    }
    \subfloat[Avoiding obstacle right before it is removed.\label{subfig:avoiding_youbot_frame2}]{
        \includegraphics[height=0.5\linewidth]{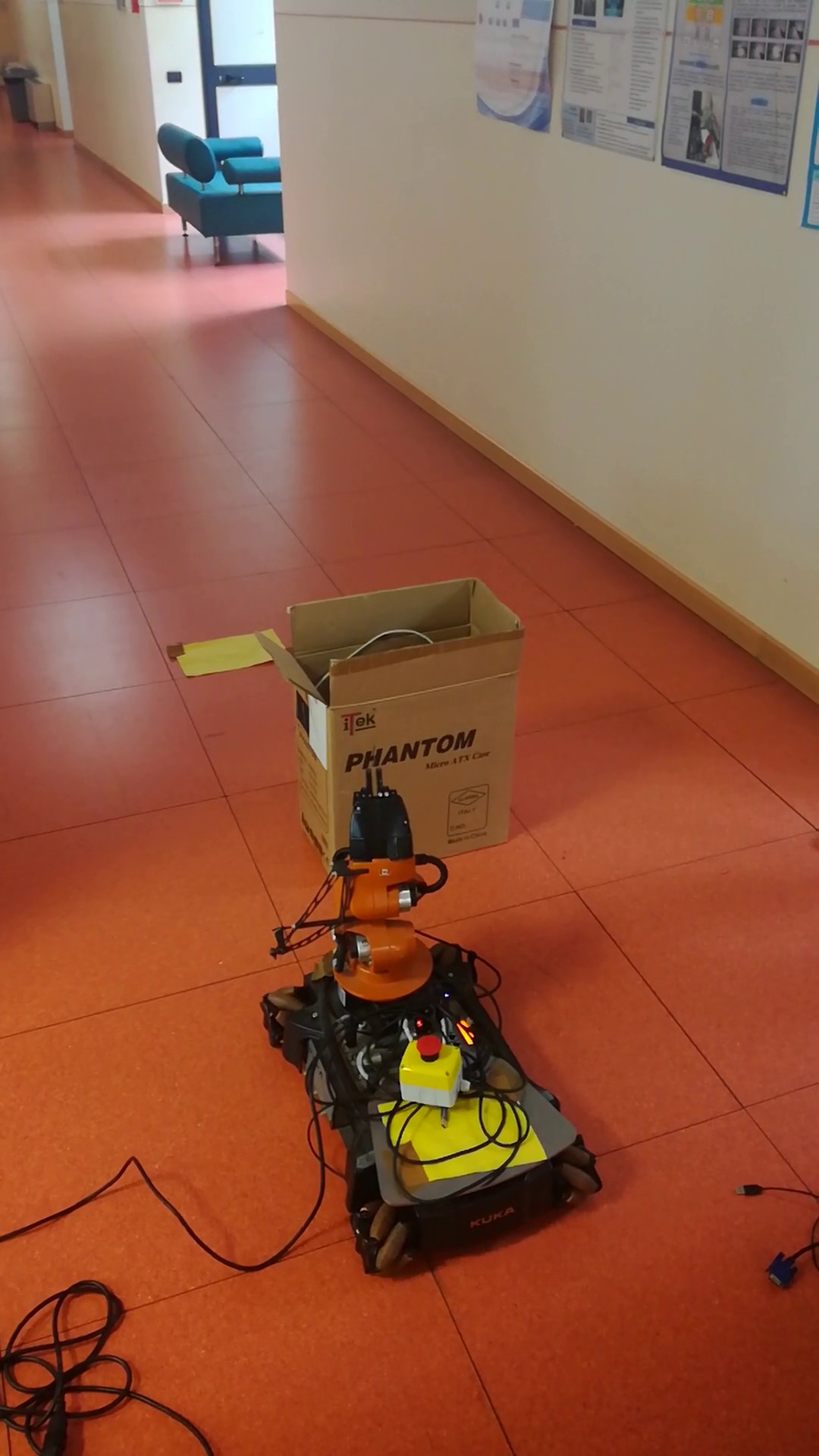}
    }
    \subfloat[Goal reached.\label{subfig:goal_youbot_frame2}]{
        \includegraphics[height=0.5\linewidth]{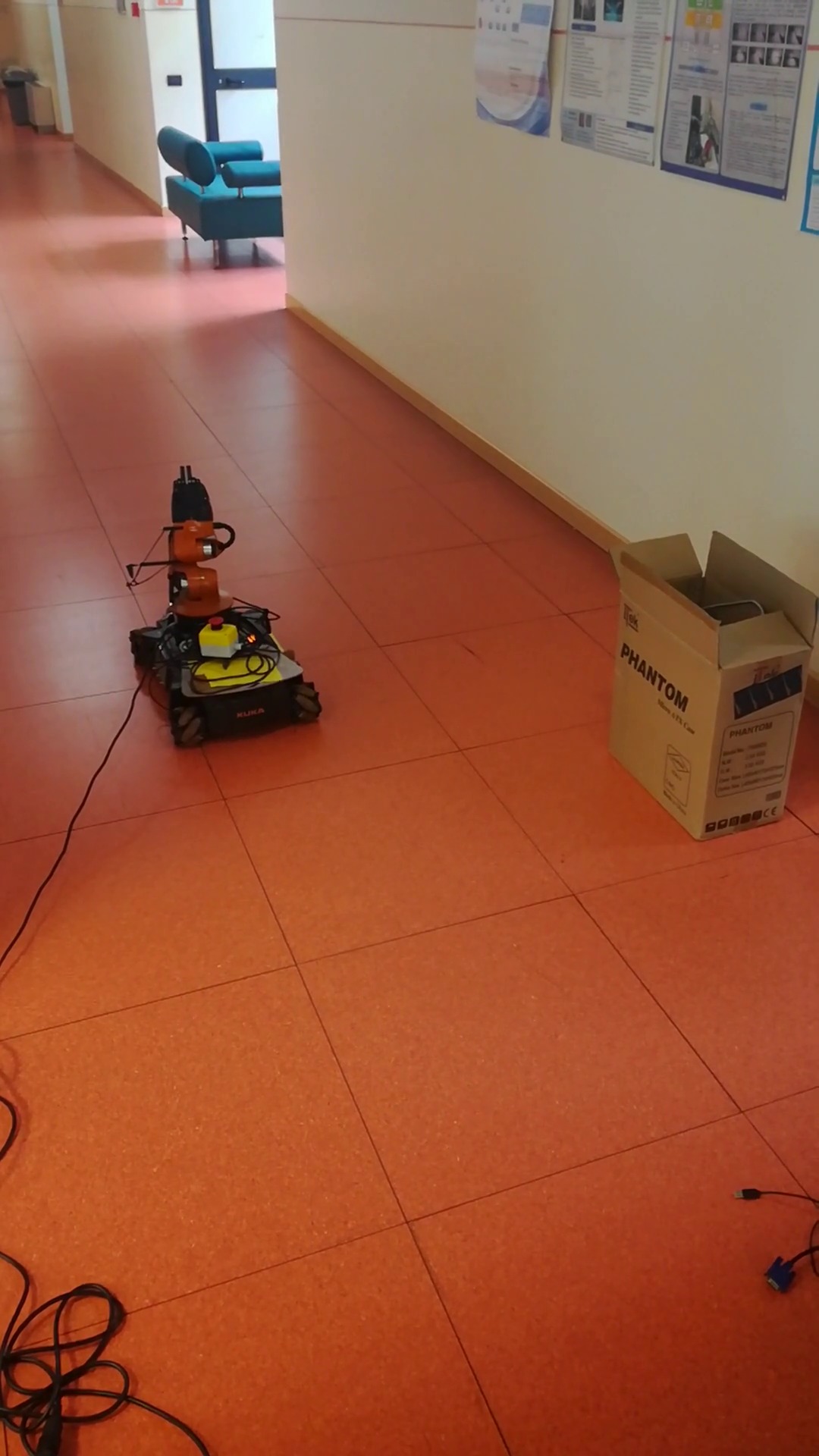}
    }
    \caption{Main steps of the YouBot task with the obstacle added to and removed from the scene during the execution.}
    \label{fig:youbot_exec2}
\end{figure}

We test our obstacle avoidance framework with a real YouBot.
The robot must move forward in a corridor for 2 meters to a pre-defined target, with an obstacle on its way.
Differently from simulations presented in Section \ref{subsec:sim_robot}, we only assume that the walls are known in advance and modeled as superquadric potentials.
On the contrary, the obstacle on the path of the robot is unknown, and it can be added to and moved away from the scene during the execution.
Hence, the YouBot is equipped with a Realsense D435 RGB-depth camera from Intel as shown in Figure \ref{fig:youbot}, in order to record the point cloud of the scene in real time.
At each time step the point cloud is filtered along the world z-axis to remove the floor and on its own depth to remove points beyond the target. 
Then it is clustered into separate point clouds for each object in the scene and is registered with the previous point cloud in a common reference frame to update the scene \ref{fig:youbot_pcd}. 
Finally an ellipsoid as in \eqref{eq:isopot_gen_ell} is fitted with $n=m=1$, enlarging axes of the dimensions of the robot (since the motion of the robot is 2-dimensional, we consider only the planar coordinates of the ellipsoid).
Fitting a pure ellipsoid rather than a pseudo-ellipsoid ($n=m=2$) guarantees a smoother perturbation to the trajectory of the robot and leverages the real-time computational complexity. 
The robot is controlled by a 2-dimensional DMP with null weights, with the same parameters as in the simulated scenario.
The camera and the YouBot controller communicate through a ROS network. 
We test two different scenarios. At first (Figure \ref{fig:youbot_exec1}) we add a box as an obstacle on the way to the goal right after the YouBot has started moving
We model the box both with the static \eqref{eq:static_potential_volume} and the dynamic \eqref{eq:dynamic_potential_volume} potential formulations.
The plots in Figures \ref{subfig:yb_static_1}-\ref{subfig:yb_dynamic_1} show that both formulations guarantee smooth trajectory adaptation, though the proportional parameter $\lambda$ in the dynamic potential must be reduced (see Figure \ref{subfig:yb_dynamic_1}).
Slight oscillations can be observed with the static potential when approaching the obstacle, since this formulation does not take into account the speed of the robot.
On the contrary, the dynamic potential ensures smoother approach to the obstacle, but slight oscillations can be observed when moving far from the obstacle.
In the second scenario, we add the box in the scene later in time, in order to further challenge the reactivity of the DMP framework.
Then, we remove the obstacle before the robot has overcome it, so that the trajectory adapts to the original straight line again.
Main steps in this scenario are shown in Figure \ref{fig:youbot_exec2}.
The plots in Figures \ref{subfig:yb_static_2}-\ref{subfig:yb_dynamic_2} compare the static and the potential formulation for the obstacle.
We notice that the static formulation guarantees smooth convergence to the goal with few oscillations, even if the obstacle is added when the robot is closer to it.
On the contrary, the dynamic formulation generates a backward oscillation when approaching the obstacle.
Parameters for DMPs and the potentials are the same as in the simulation experiments described in Section \ref{subsec:sim_robot}, except for the proportional coefficients in \eqref{eq:static_potential_volume}-\eqref{eq:dynamic_potential_volume} which are specified in the plots.
Our Online resource shows the execution of the task in the first scenario with the dynamic potential, in the second scenario with the static potential.

\section{Conclusions}

In this paper we have presented a new dynamic potential formulation for obstacle avoidance with DMPs in the Cartesian space.
This formulation extends our previous static potential one based on position, in that it takes into account the velocity of the system governed by the DMP and of the obstacle.
We have designed synthetic experiments and tests with simulated and real robots to compare our frameworks with the state-of-the-art ones existing in literature about DMPs.
Experiments with real robots are performed with an industrial manipulator, a surgical robot and a mobile robot, in order to show the generality of our framework.
One advantage of our formulations is that they consider volumetric obstacles, instead of point-like obstacles as other state-of-the-art methods, guaranteeing a more stable behavior.
Volumes are modeled with superquadric functions, which allow to describe shapes of real objects with arbitrary degree of approximation.
The synthetic experiments show that our potential formulations guarantee smoother acceleration behavior and minimal deviation from the obstacle-free trajectory defined by the forcing term of the DMP.
Moreover, the simulation experiments with three mobile robots show that our formulations can cope with multi-robot obstacle avoidance in real time, without any predefined coordination strategy between the robots.
Our new dynamic potential formulation generates fewer oscillations in proximity of the obstacles with respect to the static potential one.
In fact, the dynamic potential depends on the relative speed of the robot with respect to the obstacles, hence it deviates the trajectory earlier when the obstacle is approached.
However, the experiments with real robots show that the dynamic potential can result in higher deviations from the original trajectory, depending on the forcing term of the DMP and the position of obstacles in the scene.
On the contrary, the static potential performs better in all the experiments with the real robots, including the scenario with the mobile robot when an obstacle is added on its way during the execution.
The experiments with the industrial manipulator and the surgical robot on a pick-and-place task show that our frameworks can scale to different dimensions of the setup.
The major drawback of our formulations is that they do not guarantee convergence to the goal, which is a typical issue with potential-based formulations.

Future research will focus on the extension of our frameworks to the quaternion space. 
In fact, while the superquadric description of the volumes allows to approximate the shapes of real objects and to save more of the available workspace, the obstacle-aware adaptation of the orientation of the robot is not implemented at the moment.
Hence, the obstacles should be enlarged of the dimension of the end effector.
This is particularly evident in the scenario with the industrial manipulator, which has an encumbrant end effector.
We have partially solved this issue in our experiments, exploiting the kinematic redundancy of the robot and an efficient inverse kinematics solver to generate obstacle-free joint configurations from the Cartesian DMP waypoints.
However, this yields to higher computational time and slower execution.
We believe that representing the obstacles directly in the quaternion space at the DMP level would improve the performances.

%
\section*{Conflict of interest}

The authors declare that they have no conflict of interest.

\bibliographystyle{spmpsci}      
\bibliography{root.bib}   


\end{document}

%% file: packages.tex
\usepackage{amsmath, amsfonts, amssymb}
\usepackage[dvipsnames]{xcolor}
\usepackage{graphicx, tikz, subfig}
\usetikzlibrary{arrows.meta}
\usetikzlibrary{decorations.pathreplacing,angles,quotes}
\usetikzlibrary{shapes}
\usepackage{wrapfig}
\usepackage{bm}
\usepackage{cases}
\usepackage{verbatim}
\usepackage{nicefrac}
\usepackage{mathtools}
\usepackage{algpseudocode}
\usepackage{algorithm}
\usepackage{cool}
\usepackage{multirow}
\usepackage{nameref}
\usepackage{gensymb}

\usepackage[outline]{contour} 

%% file: math_shortcut.tex
\newcommand{\br}[1]{\left(#1\right)}
\newcommand{\sbr}[1]{\left[#1\right]}

\newcommand{\norm}[1]{\left\|{#1}\right\|}

\newcommand{\scalarp}[2]{\left\langle #1, #2 \right \rangle}

\newcommand{\vect}[1]{\mathbf{#1}}
\newcommand{\mtrx}[1]{\mathbf{#1}}

\def\idmtrx{\textbf{\text{I\small{d}}}}
\def\transpose{^\intercal}

\def\diag{\text{\normalfont diag}}

\def\vx{\vect{x}}
\def\vv{\vect{v}}
\def\vg{\vect{g}}
\def\vf{\vect{f}}
\def\vo{\vect{o}}

\def\vzero{\vect{0}}
\def\vvphi{\bm{\varphi}}

\def\mD{\mtrx{D}}
\def\mK{\mtrx{K}}
\def\mR{\mtrx{R}}

\def\NN{\mathbb{N}}

\def\RR{\mathbb{R}}


%% file: angle_scheme.tex
\begin{tikzpicture}[scale = 0.55]
    \node [label=below right:{$\vx$}, circle, draw=black, thick] (human) at (0,0) {};
    \draw[dashed, thick] (human) -- (-2, -2);
    \node [label=right:{$\mathbf{o}$}, circle, draw=black, fill = black] (obstacle) at (2.5,2.5) {};
    \draw [-{Latex}, thick] (human) --++ (-1*1.2, 2.5*1.2);
    \draw [{Latex}-, thick] (human) -- (obstacle);
    \draw [] (-1,-1) arc (225:112:1.4142);
    \node () at (-0.7,0.1) {$\theta$};
    \node[rotate=-67] () at (-0.1, 1.6) {$\vv$};
    \node[rotate=45] () at (1.8,1.0) {$ \mathbf{x} - \mathbf{o} $};
  \end{tikzpicture}

%% file: steering_angle_scheme.tex
\begin{tikzpicture}[scale = 0.7]
    \node [label=below:{$\vx$}, circle, draw=black, thick] (human) at (0,0) {};
    \node [label=right:{$\mathbf{o}$}, circle, draw=black, fill = black] (obstacle) at (2,2) {};
    \node [] (vel) at (-1,2.5) {};
    \node[rotate = -67] () at (-0.9, 1.0) {$\vv - \dot{\vo}$};
    \draw [-{Latex}, thick] (human) -- (vel);
    \draw [-{Latex}, thick] (human) -- (obstacle);
    \draw [] (1,1) arc (45:112:1.4142);
    \node () at (0.2,0.8) {$\vartheta$};
    \node[rotate=45] () at (1.3,0.7) {$ \vo - \vx $};
  \end{tikzpicture}

%% file: grad_orth_convex.tex
\begin{tikzpicture}
    \clip (-4.0, -2.2) rectangle (0.8, 3);
    \fill[opacity = 0.2] (0,0) ellipse (1 and 1.5);
    \node () at (0,0) {\rotatebox{0}{\textbf{obstacle}}};
    \draw[thick, blue] (0,0) ellipse (1 and 1.5);
    
    \draw[thick, red] (0,0) ellipse (1 * 1.7 and 1.5 * 1.7);
    \draw[thick, ForestGreen] (0,0) ellipse (1 * 2.5 and 1.5 * 2.5);
    \draw[thick, RedOrange] (0,0) ellipse (1 * 3.5 and 1.5 * 3.5);

    \node[] () at (-1.0, 1.0) {\contour{white}{\textcolor{blue}{$ C(\vx) = 0 $}}};
    \node[] () at (-1.6, 1.5) {\contour{white}{\textcolor{red}{$ C(\vx) = 1 $}}};
    \node () at (-2.3, 2.0) {\contour{white}{\textcolor{ForestGreen}{$ C(\vx) = 2 $}}};
    \node () at (-3.3, 2.5) {\contour{white}{\textcolor{RedOrange}{$ C(\vx) = 3 $}}};

    \fill (-2, -1) circle (2pt);
    \node[] () at (-1.7, -1.2) {\contour{white}{\large$ \vx $}};
    \draw[-{Latex}, thick] (-2, -1) --++ (0.5, 1.5);
    \node[rotate = 68] () at (-1.45, -0.4) {\contour{white}{\large$ \vv $}};
    \draw[-{Latex}, thick] (-2, -1) --++ (-1.5*1.3, -0.4*1.3);
    \node[] () at (-3, -1.3) {};
    \node[rotate=15] () at (-2.6, -1.6) {\contour{white}{\large$ \nabla_\vx C(\vx) $}};
    \draw[] (-2 -1.5 * 2/3, -1 -0.4 * 2/3) arc (185:77:1.0349);
    \node[] () at (-2.35, -0.65) {\contour{white}{\large$ \theta $}};
\end{tikzpicture}

%% file: grad_orth_concave.tex
\begin{tikzpicture}
    \clip(-3, 2.5) rectangle (3, -3);
    \fill[opacity=0.2] (-4, 3.5) rectangle (4, 0);
    \draw[thick, blue] (-4, 3.5) rectangle (4, 0);
    \fill[white] (0,0) circle (1.5);
    \draw[thick, blue] (-1.5, 0) arc (180:0:1.5);
    \node () at (0, 2) {\textbf{obstacle}};

    \draw[red, thick] (-4, -0.75) -- (-1.5, -0.75);
    \draw[red, thick] (4, -0.75) -- (1.5, -0.75);
    \draw[red, thick] (-1.5, -0.75) arc (-90:0:0.75);
    \draw[red, thick] (-0.75, 0) arc (180:0:0.75);
    \draw[red, thick] (0.75, 0) arc (180:270:0.75);

    \draw[ForestGreen, thick] (-4, -1.5) -- (-1.5, -1.5);
    \draw[ForestGreen, thick] (4, -1.5) -- (1.5, -1.5);
    \draw[ForestGreen, thick] (0, 0) arc (0:-90:1.5);
    \draw[ForestGreen, thick] (0, 0) arc (180:270:1.5);

    \draw[RedOrange, thick] (-4, -2.25) -- (-1.5, -2.25);
    \draw[RedOrange, thick] (4, -2.25) -- (1.5, -2.25);
    \draw[RedOrange, thick] (-1.5, -2.25) arc (-90:-48:2.25);
    \draw[RedOrange, thick] (1.5, -2.25) arc (270:228:2.25);

    \node[] () at (2.3, 0) {\contour{white}{\textcolor{blue}{$ C(\vx) = 0 $}}};
    \node[] () at (2.3, -0.75) {\contour{white}{\textcolor{red}{$ C(\vx) = 1 $}}};
    \node[] () at (2.3, -1.5) {\contour{white}{\textcolor{ForestGreen}{$ C(\vx) = 2 $}}};
    \node[] () at (2.3, -2.25) {\contour{white}{\textcolor{RedOrange}{$ C(\vx) = 3 $}}};

    \draw[ultra thick, Fuchsia, dotted] (0, 0) -- (0, -2.8);
\end{tikzpicture}

%% file: potential_function.tex
\begin{tikzpicture}
    \clip (-5, 4.5) rectangle (7.2, -5);
    \node (main) at (0,0) {\includegraphics{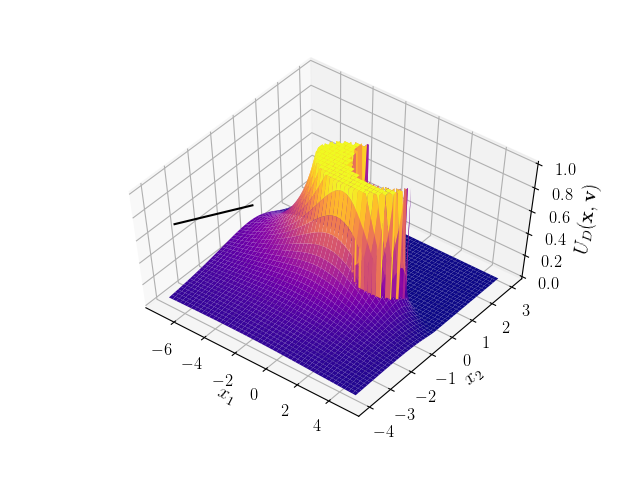}};
    \draw[thick, -{Latex}] (-3.7, 0.4) --++ (1.1*2, 1.1*0.48);
    \node () at (-2.7, 1.1) {\contour{white}{\Large $ \vv $}};
\end{tikzpicture}

%% file: isopotential.tex
\begin{tikzpicture}
    \clip (-7.5, 5.2) rectangle (6.6, -6);
    \node(main) at (0,0) {\includegraphics{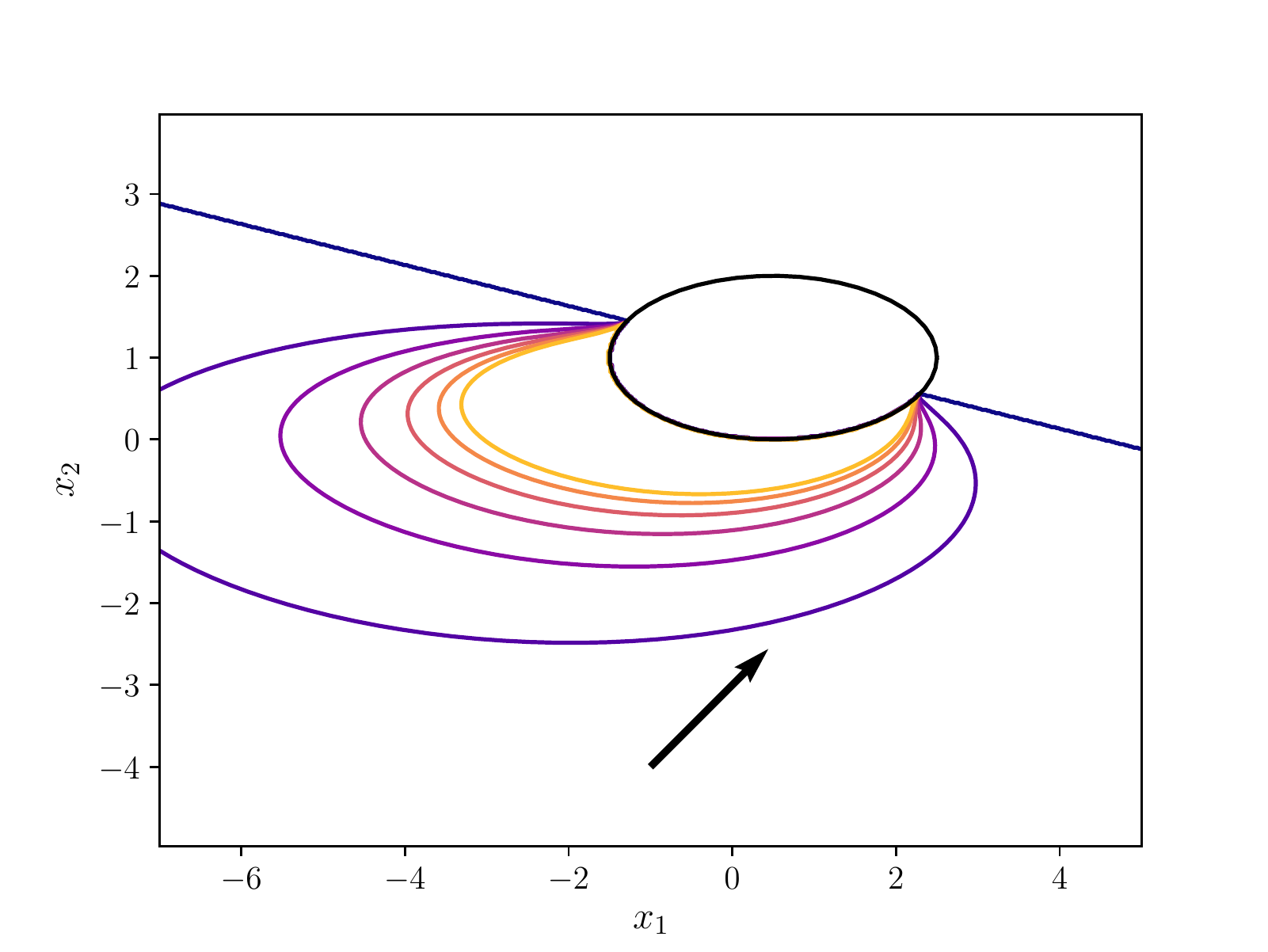}};
    \fill[opacity = 0.2] (1.8, 1.5) ellipse (2.1 and 1.08);
    \node () at (1.8, 1.5) {\Large\textbf{obstacle}};
    \node () at (1.2, -3.5) {\Large $ \vv $};
\end{tikzpicture}

%% file: one_obst_err.tex
\begin{tikzpicture}
    \node() at (0,0) {\includegraphics{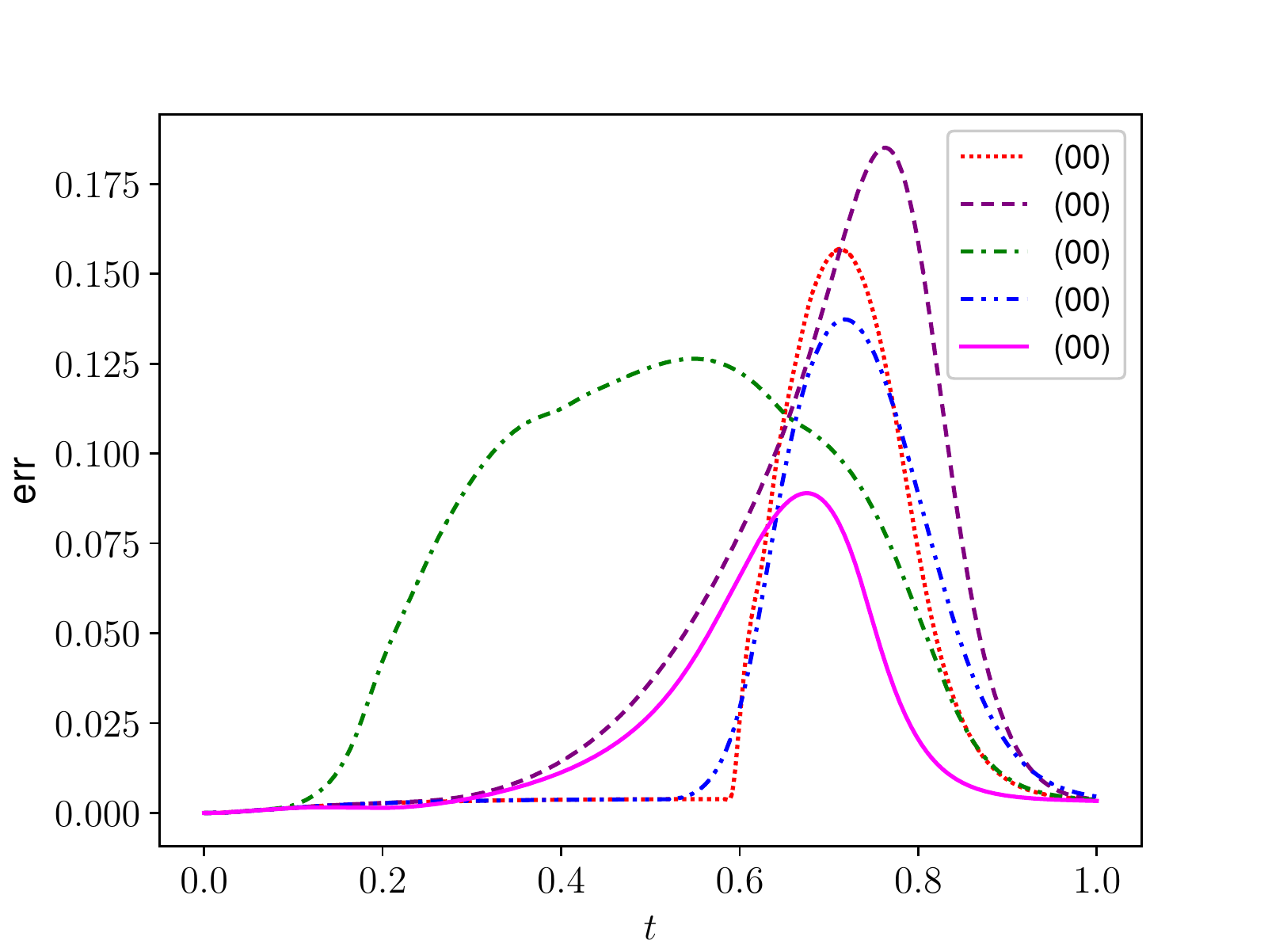}};
    \fill[white] (6.2, 4.3) rectangle++ (-1, -3);
    \node[] () at (5.6,4.1) {\large \eqref{eq:static_potential_khatib}};
    \node[] () at (5.6,3.5) {\large \eqref{eq:dynamic_potential_park}};
    \node[] () at (5.6,2.8) {\large \eqref{eq:perturb_steering_angle}};
    \node[] () at (5.6,2.2) {\large \eqref{eq:static_potential_volume}};
    \node[] () at (5.6,1.6) {\large \eqref{eq:dynamic_potential_volume}};
\end{tikzpicture}

%% file: one_obst_acc.tex
\begin{tikzpicture}
    \node() at (0,0) {\includegraphics{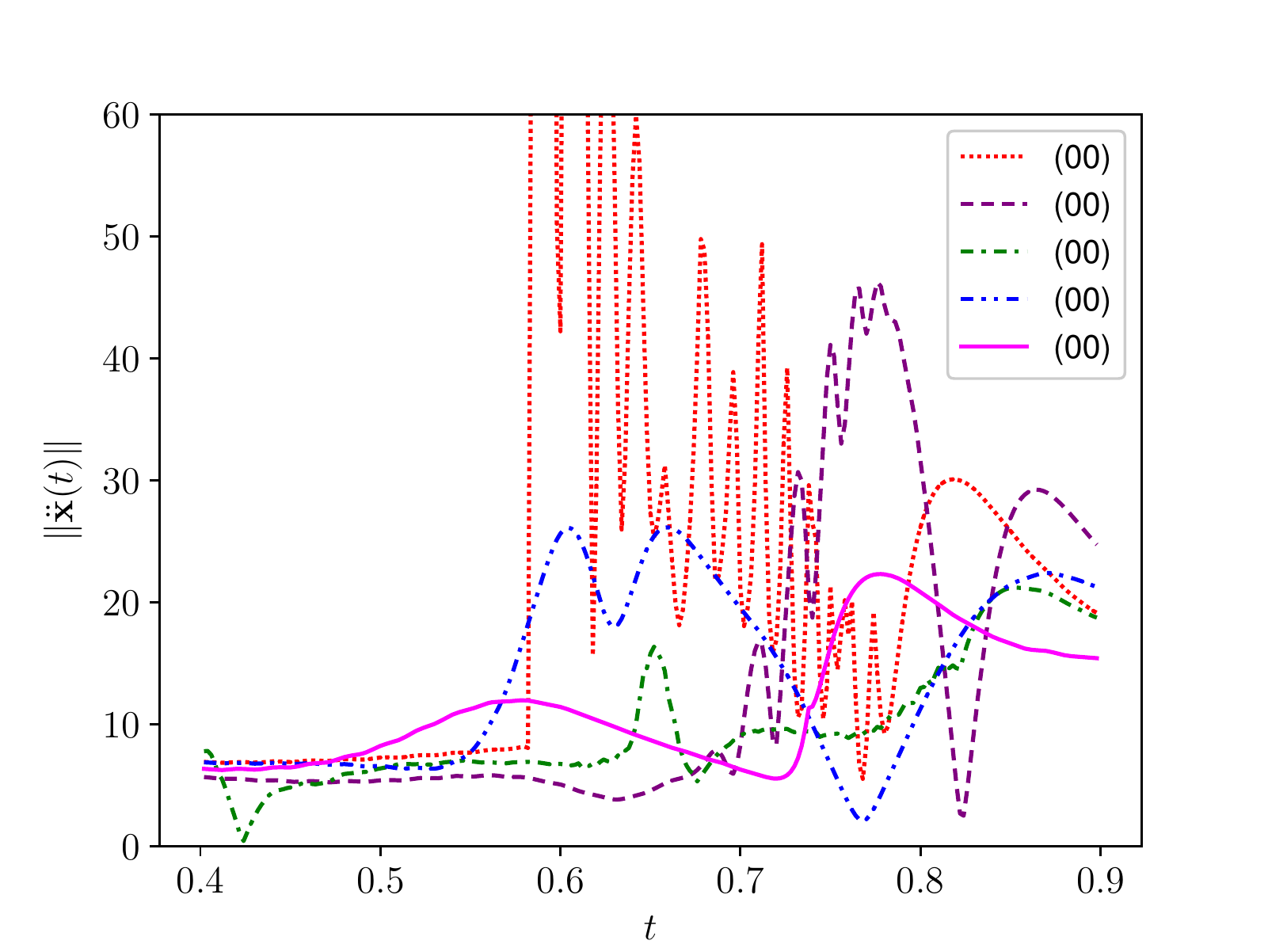}};
    \fill[white] (6.2, 4.3) rectangle++ (-1, -3);
    \node[] () at (5.6,4.1) {\large \eqref{eq:static_potential_khatib}};
    \node[] () at (5.6,3.5) {\large \eqref{eq:dynamic_potential_park}};
    \node[] () at (5.6,2.8) {\large \eqref{eq:perturb_steering_angle}};
    \node[] () at (5.6,2.2) {\large \eqref{eq:static_potential_volume}};
    \node[] () at (5.6,1.6) {\large \eqref{eq:dynamic_potential_volume}};
\end{tikzpicture}

%% file: two_obst_err.tex
\begin{tikzpicture}
    \node() at (0,0) {\includegraphics{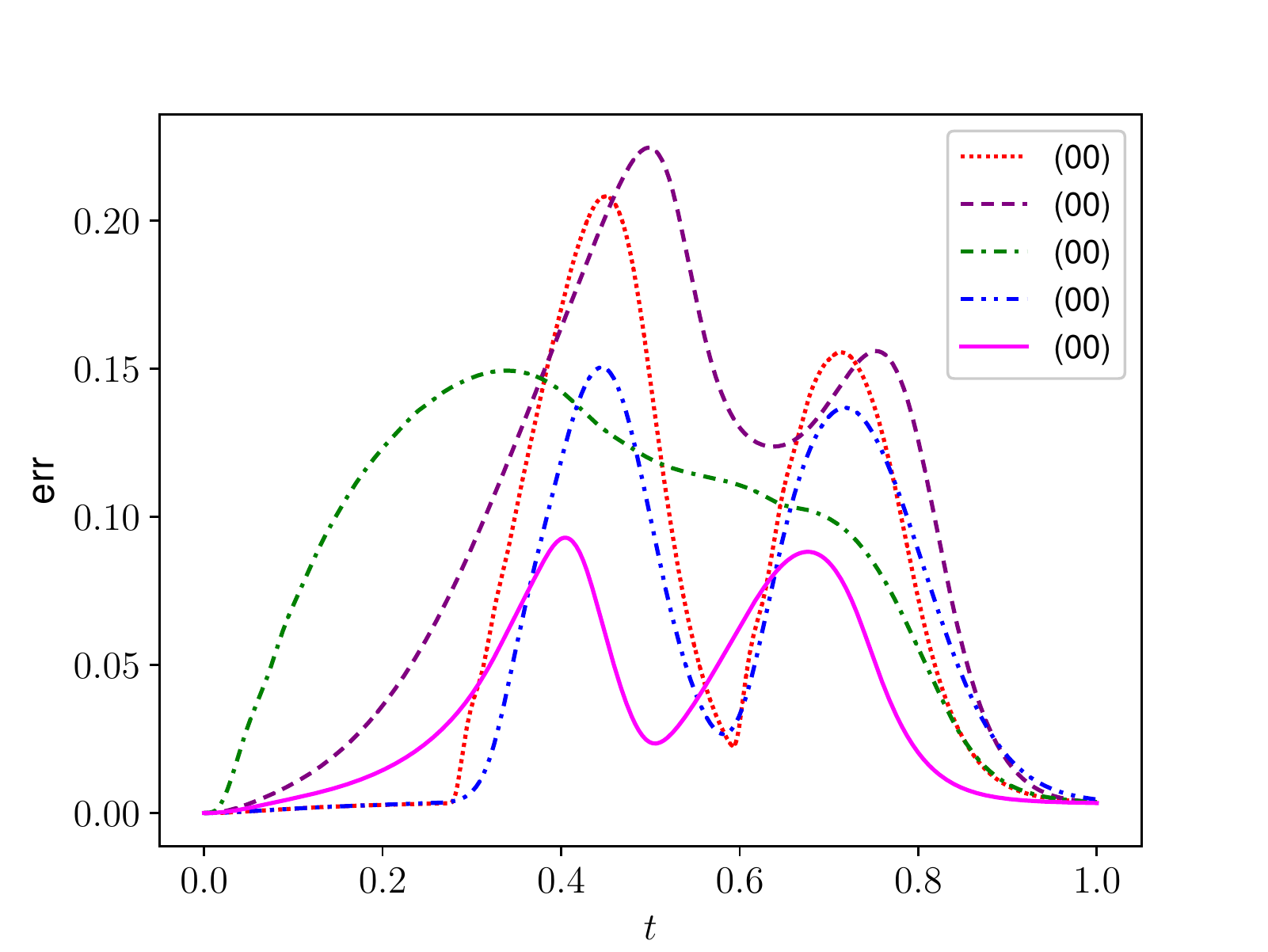}};
    \fill[white] (6.2, 4.3) rectangle++ (-1, -3);
    \node[] () at (5.6,4.1) {\large \eqref{eq:static_potential_khatib}};
    \node[] () at (5.6,3.5) {\large \eqref{eq:dynamic_potential_park}};
    \node[] () at (5.6,2.8) {\large \eqref{eq:perturb_steering_angle}};
    \node[] () at (5.6,2.2) {\large \eqref{eq:static_potential_volume}};
    \node[] () at (5.6,1.6) {\large \eqref{eq:dynamic_potential_volume}};
\end{tikzpicture}

%% file: two_obst_acc.tex
\begin{tikzpicture}
    \node() at (0,0) {\includegraphics{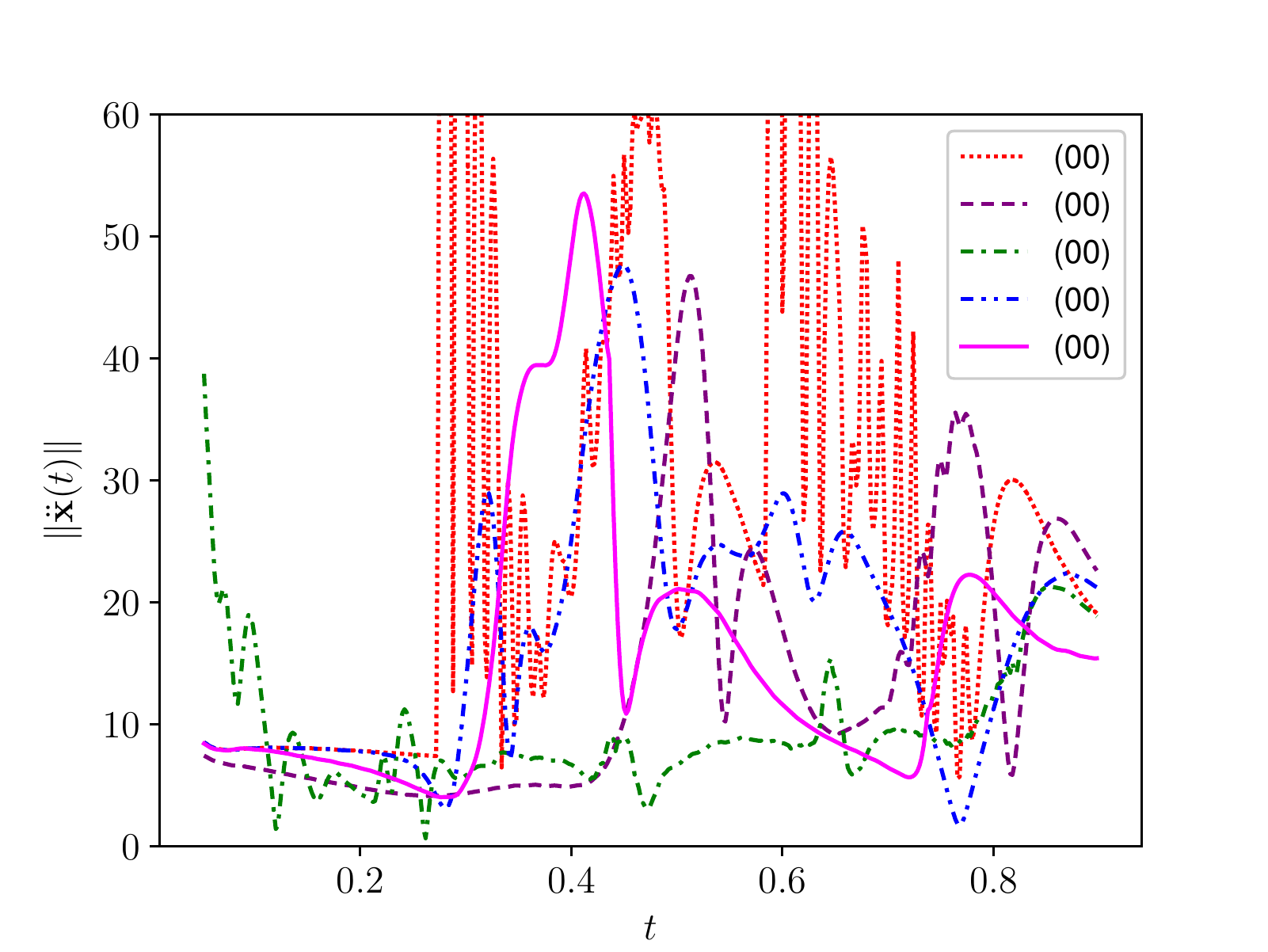}};
    \fill[white] (6.2, 4.3) rectangle++ (-1, -3);
    \node[] () at (5.6,4.1) {\large \eqref{eq:static_potential_khatib}};
    \node[] () at (5.6,3.5) {\large \eqref{eq:dynamic_potential_park}};
    \node[] () at (5.6,2.8) {\large \eqref{eq:perturb_steering_angle}};
    \node[] () at (5.6,2.2) {\large \eqref{eq:static_potential_volume}};
    \node[] () at (5.6,1.6) {\large \eqref{eq:dynamic_potential_volume}};
\end{tikzpicture}